\documentclass{article}

 \usepackage[eandd, final]{neurips_2026}

\usepackage[utf8]{inputenc} 
\usepackage[T1]{fontenc}    
\usepackage{hyperref}       
\usepackage{url}            
\usepackage{booktabs}       
\usepackage{amsfonts}       
\usepackage{nicefrac}       
\usepackage{microtype}      
\usepackage[dvipsnames,svgnames,x11names,hyperref]{xcolor}         
\usepackage[pdftex]{graphicx}
\usepackage{adjustbox} 
\usepackage{threeparttable}
\usepackage{amsmath} 
\usepackage{multirow}
\usepackage{tabularx}
\usepackage{wrapfig}
\usepackage{subcaption}

\hypersetup{
    colorlinks=true,
    citecolor=NavyBlue,
    linkcolor=NavyBlue,
    urlcolor=NavyBlue,
}
\title{GlucoFM-Bench: Benchmarking Time-Series Foundation Models for Blood Glucose Forecasting}

\author{%
  Baiying Lu \\
  Department of Computer Science\\
  Dartmouth College\\
  Hanover, NH 03755 \\
  \texttt{baiying.lu.gr@dartmouth.edu} \\
\And
  Zhaohui Liang \\
  Department of Computer Science\\
  Emory University\\
  Atlanta, GA 30322 \\
  \texttt{zhaohui.liang@emory.edu} \\
\And
  Ryan Pontius \\
  Quantitative Biomedical Sciences\\
  Dartmouth College\\
  Hanover, NH 03755 \\
  \texttt{ryan.m.pontius.hse@dartmouth.edu} \\
\And
  Shengpu Tang \\
  Department of Computer Science\\
  Emory University\\
  Atlanta, GA 30322 \\
  \texttt{shengpu.tang@emory.edu} \\
\And
  Temiloluwa Prioleau \\
  Department of Computer Science\\
  Emory University\\
  Atlanta, GA 30322 \\
  \texttt{tpriole@emory.edu} \\
}

\begin{document}

\maketitle

\begin{abstract}
    Blood glucose forecasting models are foundational for modern diabetes management systems, as reliable short-term predictions can enable proactive interventions, support automated insulin delivery, and reduce the risk of hypo- and hyperglycemic events. From a modeling perspective, glucose forecasting poses unique challenges due to heterogeneous physiological dynamics across different diabetes populations. Traditional machine learning and deep learning models have been extensively evaluated for glucose prediction. However, there is significantly less research on evaluating and benchmarking recent time-series foundation models (TSFMs) for blood glucose forecasting. To bridge this gap, we present GlucoFM-Bench, a comprehensive benchmark evaluating state-of-the-art TSFMs alongside supervised deep learning models for blood glucose forecasting. We assess eight representative architectures, including pre-trained TSFMs, time-series large language models, and task-specific deep learning models, across 15 publicly available diabetes-relevant datasets comprising 1,117 individuals with type 1 diabetes, type 2 diabetes, prediabetes, and no diabetes. Models are evaluated under zero-shot, few-shot, and full-shot protocols, with systematic variation in context length and prediction horizon. Across datasets, pre-trained TSFMs, especially Chronos-2 and TimesFM, show strong zero-shot and few-shot transfer, with the best zero-shot model performing within 5\% of the best full-shot supervised model. Yet, when task-specific data is abundant, a lightweight LSTM remains strongest, outperforming TSFMs by 4--21\% under full-shot training. Stratified analyses reveal persistent challenges in T1D cohorts and hypo-/hyperglycemic ranges, highlighting the need for evaluation beyond aggregate error metrics. Together, GlucoFM-Bench provides a standardized and reproducible foundation for evaluating, comparing, and improving foundation models for blood glucose forecasting. The dataset and code are publicly available at \url{https://huggingface.co/datasets/glucofmbench/GlucoFM-Bench} and \url{https://github.com/Augmented-Health-Lab/GlucoseML_benchmark}, respectively.

\end{abstract}

\section{Introduction}

Diabetes is one of the most prevalent chronic diseases with significant implications and costs in the U.S. and globally \cite{ong2023global,guzman2025number,wang2021trends_US}. People living with diabetes require continuous and lifelong management of blood glucose (BG) levels to maintain a healthy standard of living and reduce the risk of micro- and macro-vascular complications \cite{papatheodorou2018complications}. In recent years, continuous glucose monitors (CGMs) have transformed the standard of diabetes care by enabling more timely and data-driven management \cite{american20247,dovc2020evolution}. As a result, modern automated insulin delivery (AID) relies heavily on accurate blood glucose forecasting to support management decisions that seek to prevent adverse glycemic events (i.e., hypo- and hyperglycemia). However, glucose forecasting remains challenging because blood glucose trends are shaped by daily-living factors as well as complex metabolic regulation \cite{Kim2023, Phillips2023, Kovatchev2016}. In addition, heterogeneous physiological dynamics across populations with different diabetes types (i.e., type 1 diabetes (T1D), type 2 diabetes (T2D), prediabetes (PreD), and no diabetes (ND)) add further complexity to the task of glucose forecasting \cite{Afsaneh2022, Khan2019, Spartano2025}.

Most state-of-the-art (SOTA) blood glucose forecasting methods rely on supervised learning, including traditional statistical models and deep learning (DL) architectures such as RNNs, LSTMs, CNNs, and Transformers~\cite{zhu2020deep, Martinsson2019, Deng2021, Zhu2025, Karagoz2025, lu2026}. However, these models are often trained and evaluated on individual datasets or narrow cohorts, leading to limited generalizability and strong dataset dependence~\cite{Afsaneh2022, Martinsson2019, Deng2021, lu2026, Ghimire2024, Shao2024}. This limitation motivates interest in transferable forecasting approaches that can better adapt across datasets, cohorts, and deployment settings. Time-series foundation models (TSFMs), including pre-trained time-series models and LLM-based forecasting frameworks, offer a promising direction by learning transferable representations from large-scale, multi-domain time-series data~\cite{Li2025, Liang2024, Woo2024, Jin2023timellm}. These models span diverse architectural paradigms, including decoder-only, encoder-only, encoder--decoder, and language-model-based adaptation frameworks, and have demonstrated strong zero-shot and few-shot forecasting potential on general-purpose benchmarks~\cite{ Woo2024, Ansari2024chronos, Liu2024timer, Das2023timesfm, Liu2024calf}. However, existing TSFM evaluations have primarily focused on domains such as energy, finance, traffic, and weather forecasting, with limited evaluation on critical health-relevant tasks like glucose forecasting. Only a few studies have explored glucose-specific foundation models or LLM-based adaptation for diabetes-related tasks~\cite{Lutsker2026, Luo2025, Li2025LLM, Lara-Abelenda2025}. To date, there is no existing benchmark evaluation of general-purpose TSFMs for blood glucose forecasting (see related work in Appendix~\ref{sec:related_work}.)

To bridge this gap, we present GlucoFM-Bench, a comprehensive benchmark of SOTA TSFMs for continuous blood glucose forecasting. GlucoFM-Bench comprises benchmark evaluation on a total of 15 publicly available CGM datasets, including 12 open-access datasets and 3 controlled-access datasets. In this work, we evaluate two major classes of general-purpose TSFMs, namely pre-trained time-series foundation models and LLM-based time-series forecasting models, as well as two SOTA supervised deep learning architectures specifically designed for glucose forecasting to contextualize performance. Models are assessed under zero-shot, few-shot fine-tuning, and full-shot fine-tuning protocols to evaluate out-of-domain generalization, data-efficient adaptation, and task-specific performance, respectively. To capture glycemic-relevant heterogeneity, we further analyze performance across glycemic ranges and diabetes cohorts, and evaluate predictions using both standard accuracy metrics (RMSE and MAE) and clinical risk metrics based on Clarke and Surveillance Error Grids. In addition to model evaluation, we curate and standardize twelve open-access CGM datasets through unified preprocessing and benchmark-ready train-test partitioning, enabling reproducible evaluation and establishing a standardized downstream task for future healthcare foundation model research.

Our contributions are summarized as follows:
\begin{itemize}
    \item We introduce GlucoFM-Bench, the first comprehensive benchmark evaluating eight SOTA TSFMs and task-specific supervised deep learning models for continuous blood glucose prediction across 15 publicly available CGM datasets.
    \item We employ zero-shot, few-shot, and full-shot evaluations alongside stratified analyses to assess model transferability, adaptation efficiency, and performance across heterogeneous diabetes types and glycemic conditions.
    \item We curate and standardize a unified, ready-to-use open-access CGM benchmark dataset, enabling consistent and reproducible evaluation of time-series foundation models and establishing a standardized downstream task for future healthcare foundation model research.
\end{itemize}

\begin{figure}[h]
  \centering
  \includegraphics[width=\linewidth]{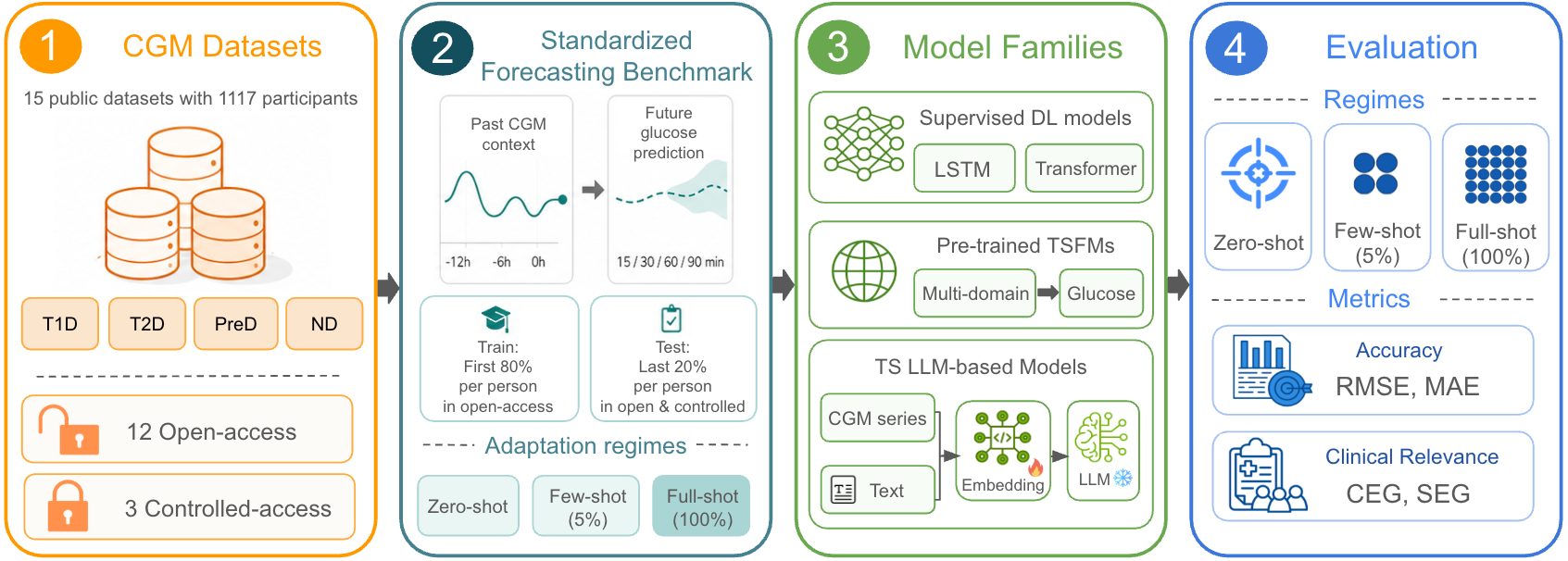}
  \caption{Overview of GlucoFM-Bench, including datasets, models, and evaluation protocols leveraged for glucose forecasting.}
  \label{fig:overview}
\end{figure}

\section{GlucoFM-Bench}
Figure \ref{fig:overview} presents an overview of the GlucoFM-Bench pipeline, including CGM datasets used for analysis, benchmarking approach, models compared, and evaluation methods and metrics. 


\subsection{Datasets}

To support reproducibility and provide an accessible downstream benchmark for blood glucose forecasting with TSFMs, we prioritize recent open-access CGM datasets (2016–2026), while maximizing diversity in participant populations, diabetes status, recording conditions, and dataset scale. Based on these selection criteria, we include 12 open-access datasets for the main model training, fine-tuning, and evaluation. We additionally incorporate three widely used controlled-access datasets, OhioT1DM~\cite{Marling2020}, DiaTrend~\cite{Prioleau2023}, and T1DEXI~\cite{T1DEXI}, as independent external test sets to assess cross-dataset generalization, given their broad adoption in prior blood glucose forecasting studies~\cite{Martinsson2019, Neumann2025, Liu2025}. Across all datasets, CGM time-series measurements serve as the forecasting target, with glucose values ranging from 40 to 400 mg/dL.

The 12 open-access datasets span five countries (U.S., U.K., China, Spain, and Chile), cover a broad age range (18--97 years), and include diverse metabolic phenotypes, ranging from T1D and T2D to PreD and ND populations. This cohort heterogeneity leads to substantial differences in glycemic profiles across datasets. Table~\ref{tab:dataset_overview} summarizes these differences using daily time-in-range ratio (TIR; 70--180 mg/dL), daily mean blood glucose, daily glycemic variability (GV), and 24-hour circadian periodicity. The definition and detailed calculation of these metrics are provided in Appendix~\ref{sec:appendix_calculation}. Several open-access datasets, including Hall2018~\cite{Hall2018}, Colas2019~\cite{Cols2019}, and BIG IDEAS Lab~\cite{Bent2021}, include T2D, PreD, and ND participants and show relatively stable glycemic profiles, with average daily TIR above 0.95 and low mean glucose levels. In contrast, T1D-dominated datasets, such as D1NAMO~\cite{Dubosson2018}, ShanghaiT1DM~\cite{Zhao2023}, and HUPA-UCM~\cite{Hidalgo2024}, have substantially lower average TIR below 0.65 and higher glycemic variability, indicating more challenging forecasting conditions. Together, these differences demonstrate the broad range of glycemic patterns represented in GlucoFM-Bench. Circadian periodicity also varies across datasets: most datasets show low-to-moderate daily regularity, with values ranging from 0.26 to 0.53, while the higher values observed in D1NAMO and Colas2019 are likely influenced by their short recording durations (e.g. 2-3 days per subject).

The 3 controlled-access datasets complement the open-access collections with larger T1D cohorts, comprising over 11 million CGM measurements in total. Compared with the open-access datasets, these cohorts show lower average TIR ratios and higher mean glucose levels, providing a challenging external evaluation setting. Together, the open- and controlled-access datasets enable comprehensive benchmarking of TSFMs across diverse populations, dataset scales, and glycemic management regimes. Additional participant metadata, including gender ratio, age distribution, race/ethnicity, and cohort summary, are provided in Appendix Section~\ref{sec:meta_cohort_summary}.

\begin{table*}[ht]
  \caption{Overview of 15 public continuous glucose monitoring (CGM) datasets used in this study, including 12 open-access \& 3 controlled-access CGM datasets.}
  \label{tab:dataset_overview}
  \centering
  \resizebox{\textwidth}{!}{
  \begin{threeparttable}
 \begin{tabular}{lccccccccc}
  \toprule
  DatasetName & \# of CGM  & \# of  &
  Participant  & Sampling  & Missing &
  Avg. Daily TIR* ratio & Avg. Daily BG* & Avg. GV* & Periodicity \\
  & record & Participant & diabetes type & frequency & ratio &
  Mean (STD) & Mean (STD) & Mean (STD) & Mean (STD)\\
\midrule
\textbf{OpenAccess} & & & & & & & & \\

\quad Hall2018 \cite{Hall2018}
  & 105{,}426 & 57 & T2D, PreD, ND
  & 5 min & 9.56\%
  & 95.03\% (5.22\%) & 102.54 (11.67) & 16.01 (3.85) & 0.27 (0.14) \\

\quad D1NAMO \cite{Dubosson2018}
  & 8{,}221 & 9 & T1D
  & 5 min & 11.32\%
  & 63.51\% (17.24\%) & 156.17 (28.87) & 32.54 (7.53) & 0.67 (0.24)\\

\quad Colas2019 \cite{Cols2019}
  & 114{,}253 & 208 & T2D, ND
  & 5 min & 0\%
  & 95.59\% (7.84\%) & 102.28 (12.21) & 15.89 (5.34) & 0.92 (0.08)\\

\quad BIG IDEAS Lab \cite{Bent2021}
  & 36{,}898 & 16 & PreD, ND
  & 5 min & 3.30\%
  & 97.68\% (1.91\%) & 114.65 (10.34) & 16.00 (2.96) & 0.30 (0.14)\\

\quad ShanghaiT1DM \cite{Zhao2023}
  & 15{,}695 & 12 & T1D
  & 15 min & 0\%
  & 54.42\% (11.74\%) & 165.69 (28.36) & 31.43 (7.70) & 0.48 (0.13)\\

\quad ShanghaiT2DM \cite{Zhao2023}
  & 112{,}462 & 100 & T2D
  & 15 min & 0.24\%
  & 77.02\% (17.45\%) & 141.18 (30.36) & 24.43 (5.31) & 0.53 (0.16)\\

\quad UCHTT1DM \cite{Langarica2024}
  & 29{,}174 & 20 & T1D, ND
  & 5 min & 6.54\%
  & 82.98\% (14.68\%) & 116.15 (33.45) & 21.11 (9.91) & 0.38 (0.23)\\

\quad HUPA-UCM \cite{Hidalgo2024}
  & 309{,}392 & 25 & T1D
  & 5 min & 0\%
  & 60.91\% (14.45\%) & 154.44 (25.91) & 35.08 (6.65) & 0.30 (0.12)\\

\quad CGMacros \cite{Das2025}
  & 629{,}825 & 45 & T2D, PreD, ND
  & 1 min & 1.45\%
  & 85.68\% (19.61\%) & 140.07 (27.77) & 17.76 (4.75) & 0.27 (0.13)\\

\quad T1DM-UOM \cite{Alsuhaymi2025}
  & 356{,}146 & 17 & T1D
  & 5/15 min & 5.14\%
  & 72.90\% (13.87\%) & 150.07 (19.85) & 30.88 (4.97) & 0.28 (0.06)\\

\quad Bris-T1D \cite{James2025}
  & 848{,}574 & 20 & T1D
  & 5 min & 3.55\%
  & 69.03\% (11.80\%) & 157.58 (18.68) & 30.94 (4.25) & 0.26 (0.16)\\

\quad AZT1D \cite{Khamesian2025}
  & 306{,}712 & 25 & T1D
  & 5 min & 0.51\%
  & 77.80\% (10.75\%) & 147.09 (16.53) & 27.25 (3.49) & 0.31 (0.07)\\

\midrule
\textbf{ControlledAccess} & & & & & & & & \\

\quad OhioT1DM \cite{Marling2020}
  & 166{,}533 & 12 & T1D
  & 5 min & 11.83\%
  & 63.40\% (9.58\%) & 159.15 (17.08) & 31.24 (3.72) & 0.26 (0.07)\\

\quad DiaTrend \cite{Prioleau2023}
  & 7{,}680{,}740 & 54 & T1D
  & 5 min & 6.71\%
  & 53.69\% (14.61\%) & 182.69 (27.17) & 30.65 (4.30) & 0.28 (0.05)\\

\quad T1DEXI \cite{T1DEXI}
  & 3{,}785{,}253 & 497 & T1D
  & 5 min & 4.15\%
  & 73.23\% (17.04\%) & 145.06 (29.19) & 29.18 (5.05) & 0.26 (0.10)\\

\bottomrule
\end{tabular}
\begin{tablenotes}[flushleft]
    \small
    \item[*] TIR: Time in range (70-180 mg/dL, glucose management target range); BG: Blood glucose; GV: Glycemic variability; STD: Standard deviation.
  \end{tablenotes}
\end{threeparttable}
}
\end{table*}

\paragraph{Ethical Considerations}
The use of publicly available, de-identified data was deemed not human subjects research, thus institutional review board (IRB) review is not required. The evaluated models are intended for research benchmarking only.

\subsection{Standardized Forecasting Benchmark}
\label{sec:standardized_forecasting_bench}

\subsubsection{Preprocessing}
As shown in Table~\ref{tab:dataset_overview}, the original datasets exhibit varying sampling intervals. Also, there are 0\%-11.83\% of missing values across different datasets, which are common in real-world CGM data. To standardize the forecasting task while preserving data quality, we first harmonized the sampling frequency across datasets by upsampling the ShanghaiT1DM and ShanghaiT2DM datasets and downsampling the CGMacros dataset to a unified resolution of one sample per five minutes. To handle missing values, we applied linear interpolation to gaps shorter than one hour, a common practice for CGM data~\cite{Zhu2025, Karagoz2025}. Gaps longer than one hour were treated as sequence boundaries to avoid introducing spurious temporal continuity.

\subsubsection{Standardized Forecasting Setup} 
\paragraph{Training–Test Split.}
For the open-access datasets, we chronologically split each participant's CGM time series, using the first 80\% for training or fine-tuning and the remaining 20\% for evaluation, following common practice in glucose forecasting studies~\cite{Martinsson2019, Zhu2025, Ghimire2024, Karagoz2025}. For the controlled-access datasets, we use only the final 20\% of each participant's records as external test sets to align with prior protocols and enable comparison with existing results. This per-participant chronological split maximizes data utilization under imbalanced diabetes-type distributions while avoiding the randomness and strong inter-individual variability introduced by participant-level splits.

\begin{wrapfigure}{r}{0.5\linewidth}
  \centering
  \vspace{-8pt}
  \includegraphics[width=\linewidth]{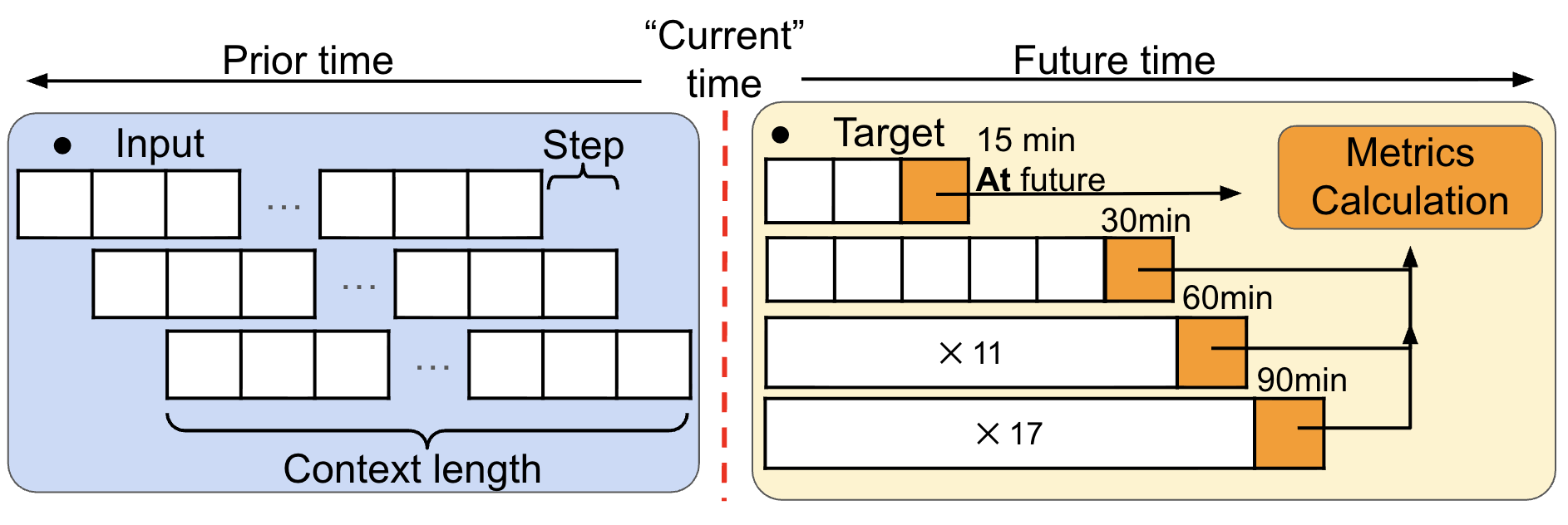}
  \caption{Sliding window setup for forecasting glucose values at 15, 30, 60, and 90 minutes ahead.}
  \label{fig:sliding_window}
  \vspace{-8pt}
\end{wrapfigure}

\paragraph{Window Sampling for Training and Evaluation.} 
We use a sliding-window approach to construct model inputs, with each windowed segment serving as the prediction context. To prevent temporal leakage, sliding windows are constructed separately within each partition. Forecasting targets are CGM values at 15, 30, 60, and 90 minutes ahead, following common settings in prior glucose forecasting studies~\cite{Martinsson2019, Deng2021, Zhu2025, Ghimire2024, Karagoz2025}. For each horizon, models forecast up to the target horizon, and performance is computed using the final-step prediction, as illustrated in Figure~\ref{fig:sliding_window}. To assess context-length effects, we vary the input window size in zero-shot evaluation across 1, 4, 8, 12, 16, and 24 hours. For few-shot and full-shot fine-tuning, we use 12 hours as the default context length because it balances forecasting performance with participant-level data utilization; longer contexts can substantially reduce the number of eligible participant sequences.


Given the training corpus size of approximately 2.8 million CGM records, we use a deterministic subsampling strategy for model fine-tuning. In the full-shot setting, sliding windows are generated with a 1-hour stride (12 samples), balancing training density and computational feasibility. In the few-shot setting, we increase the stride to 20 hours (240 samples), yielding approximately 5\% of the full-shot training windows while preserving uniform temporal coverage across participants. This strategy reduces computational cost without introducing random sampling variability. During evaluation, we use a stride of one sample to generate dense per-record forecasts and ensure comprehensive performance assessment.

\subsection{Models}

To provide a comprehensive evaluation of blood glucose forecasting methods, GlucoFM-Bench includes representative models spanning multiple forecasting paradigms, covering both established and recently proposed approaches with competitive reported performance and publicly available implementations. The following sections describe the selected models and their architectural characteristics.

\paragraph{Persistence Forecasting Baseline}
We include a persistence baseline, a common reference method in glucose forecasting that uses the most recent CGM value as the prediction for a future time point. Despite its simplicity, it provides an informative reference for evaluating model performance~\cite{Freiburghaus2020,prioleau2025glucose}.

\paragraph{Supervised Deep Learning Models}
Among supervised deep learning methods for glucose forecasting, LSTM- and Transformer-based models are two widely used architectures with strong reported performance~\cite{Ghimire2024}. We therefore include two task-specific point-forecasting models: the LSTM-based model from Martinsson et al.~\cite{Martinsson2019} and the Transformer-based model (GPFormer) from Zhu et al.~\cite{Zhu2025}. 

\paragraph{Pre-trained Time Series Foundation Models}
In this study, we evaluate four state-of-the-art pre-trained TSFMs: Chronos-2~\cite{FatirAnsari2025chronos2}, Moirai 2.0~\cite{Liu2025moirai2}, Timer~\cite{Liu2024timer}, and TimesFM 2.5~\cite{Das2023timesfm}. We select these models due to their general-purpose design, strong reported performance in zero-shot and transfer learning settings~\cite{Li2025, Shi2024timemoe}, and coverage of diverse forecasting architectures. These models are pre-trained on large-scale, multi-domain time-series corpora spanning domains such as energy, retail demand, web traffic, and common forecasting benchmarks, but have not been specifically exposed to glucose time series. Architecturally, they cover encoder-only and decoder-only Transformer variants, including sequence-to-sequence generation, autoregressive forecasting, GPT-style generative pretraining, and patch-based forecasting designs. Chronos-2, Moirai 2.0, and TimesFM 2.5 also support probabilistic forecasting, enabling prediction intervals for uncertainty estimation.

\paragraph{Large Language Models for Time Series Forecasting}
We also evaluate two representative LLM-based time-series forecasting methods: TimeLLM~\cite{Jin2023timellm} and CALF~\cite{Liu2024calf}. Unlike TSFMs pre-trained directly on time-series data, these methods adapt language models pre-trained on text to forecasting by converting numerical time series into token-like or cross-modal representations. We include TimeLLM and CALF as representative and competitive frameworks, covering two distinct paradigms for adapting language models to time-series forecasting: textual-prompt-based reprogramming and cross-modal fine-tuning. TimeLLM maps patched time-series inputs into the embedding space of a frozen LLM and uses textual prompts for task conditioning, with the prompts used in our experiments provided in Appendix~\ref{timellm_prompts}. CALF aligns numerical sequences and structured textual inputs through a cross-modal fine-tuning framework. In all experiments, both methods use GPT-2~\cite{gpt2} as the LLM backbone, which is supported by both TimeLLM and CALF and enables a controlled comparison under a shared language-model backbone. 

Overall, our model selection is intended to enable a representative and reproducible comparison across major forecasting paradigms, rather than an exhaustive evaluation of all available methods.

\subsection{Evaluation}
\subsubsection{Evaluation Protocols}

To assess model generalization, adaptability, and data efficiency, GlucoFM-Bench uses a three-tiered evaluation protocol comprising: \textit{zero-shot}, \textit{few-shot}, and \textit{full-shot}. For fine-tuning, we use model-specific configurations following the original implementations.
\begin{itemize}
    \item \textbf{Zero-shot.} Pre-trained TSFMs are applied directly to blood glucose forecasting without fine-tuning, evaluating their ability to transfer temporal representations to unseen glucose domains. This setting is limited to pre-trained TSFMs, as supervised deep learning models and LLM-based forecasting models require task-specific training and do not provide directly usable pre-trained forecasting checkpoints.
    
    \item \textbf{Few-shot.} Models are fine-tuned using 5\% of the training data, simulating low-resource conditions and assessing sample efficiency with limited domain-specific data.
    
    \item \textbf{Full-shot.} Models are fine-tuned on the complete training set, measuring performance under fully supervised conditions with abundant task-specific data. 
\end{itemize}

\subsubsection{Evaluation metrics}

\paragraph{Predictive Performance}
We evaluate forecasting accuracy using Root Mean Square Error (RMSE) and Mean Absolute Error (MAE), two most widely used metrics for quantifying the discrepancy between predicted and actual glucose values at specific prediction horizons~\cite{Martinsson2019, Deng2021, Zhu2025, Ghimire2024}. Metrics are computed separately at 15-, 30-, 60-, and 90-minute horizons to assess short- and medium-term predictive performance. We additionally compute RMSE stratified by ground-truth CGM values into hypoglycemia (<70 mg/dL), in-range glucose (70--180 mg/dL), and hyperglycemia (>180 mg/dL) to characterize performance across clinically distinct glucose ranges~\cite{Battelino2019}.
For each horizon, evaluation compares only the model prediction at the target horizon with the corresponding ground-truth CGM value, rather than averaging errors over intermediate forecast steps. We report RMSE in the main paper and provide MAE in Appendix tables. Formulas for these metrics are provided in Appendix Section~\ref{sec:evaluation_metrics_eq}.

\paragraph{Clinical Relevance} We assess the clinical risk of forecasting errors using the Clarke Error Grid (CEG) and Surveillance Error Grid (SEG)~\cite{Clarke2005, Klonoff2014}, which map glucose prediction errors to clinically meaningful risk zones. We summarize both grids by the no-risk zone ratio, i.e., the proportion of predictions in the no-clinical-risk zone. We report SEG no-risk ratios in the main results, with CEG results and grid explanations provided in Appendix result tables. The CEG, SEG, and detailed explanation can be found in Appendix Figure~\ref{fig:ceg_seg}.

\section{Experiment}

\subsection{Zero-shot}
We first evaluate the zero-shot performance of pre-trained TSFMs under the primary setting of a 12-hour context window and a 30-minute prediction horizon. Appendix Figure~\ref{fig:sampling_horizon} examines the effect of context length on zero-shot performance and supports our choice of a 12-hour context window, which balances forecasting accuracy with participant-level data utilization.  Zero-shot evaluation is restricted to pre-trained TSFMs because other model categories require task-specific training. Table~\ref{tab:zeroshot_rmse} reports RMSE and SEG no-risk ratios for the persistence baseline and pre-trained TSFMs. Additional zero-shot results across other forecasting horizons are reported in Appendix Table~\ref{tab:zeroshot_combined_horizon}. Detailed persistence baseline analyses, including MAE, CEG/SEG no-risk ratios, and forecasting skill scores, are provided separately in Appendix Table~\ref{tab:persistence_30_details}. For TSFMs that support probabilistic forecasting (Chronos-2, Moirai2.0, TimesFM2.5), we additionally report 80\% prediction interval (PI80) coverage across zero-, few-, and full-shot protocols in Appendix Tables~\ref{tab:zeroshot_pi80}--\ref{tab:fullshot_pi80}.


\begin{table}[ht]
  \caption{Zero-shot RMSE (mg/dL) and SEG no-risk ratio (\%) for pre-trained TSFMs under the 12-hour context, 30-minute horizon setting. Values are participant-level mean (STD).}
  \label{tab:zeroshot_rmse}
  \centering
  \resizebox{\textwidth}{!}{
  \begin{threeparttable}
  \begin{tabular}{lccccc|cccc}
    \toprule
    & \multicolumn{5}{c|}{RMSE (mg/dL)}
    & \multicolumn{4}{c}{SEG No-risk Zone Ratio (\%)} \\
    \cmidrule(lr){2-6}
    \cmidrule(lr){7-10}
    Dataset & Baseline & Chronos-2 & Moirai2.0 & Timer & TimesFM2.5  & Chronos-2 & Moirai2.0 & Timer & TimesFM2.5\\
    \midrule
    \textbf{OpenAccess} & & & & & \\
    \quad Hall2018  & 13.65 (4.52)   & 13.24 (6.96)      & \underline{13.16 (6.06)}      & 14.05 (7.11)      & \textbf{12.96 (5.83)}  & 90.53 (10.93) & \textbf{91.97 (10.26)} & 89.07 (15.01) & \underline{91.61 (10.49)}\\
    \quad D1NAMO   & 28.18 (11.92)   & 24.46 (17.72)      & \underline{21.06 (13.59)}      & 30.56 (21.02)      & \textbf{20.65 (13.49)} & 75.32 (19.93) & \textbf{95.24 (8.13)} & 73.16 (13.87) & \underline{88.10 (15.85)}\\
    \quad Colas2019$^{1}$  & 13.26 (7.14)    & --      & --      & --      & --  & --      & --      & --      & -- \\
    \quad BIG IDEAS Lab  & 16.91 (4.03)    & \underline{15.17 (3.04)}      & 15.18 (3.24)      & 16.24 (4.78)      & \textbf{14.88 (3.00)} & 94.04 (4.90) & \underline{94.85 (4.78)} & \textbf{95.30 (5.59)} & 94.73 (5.04)\\
    \quad ShanghaiT1DM  & 17.18 (6.57)    & 12.82 (3.41)      & \underline{12.50 (3.18)}      & 19.10 (6.16)      & \textbf{12.11 (3.45)} & 92.86 (3.74) & \underline{95.37 (4.29)} & 88.26 (6.52) & \textbf{95.99 (5.57)}\\
    \quad ShanghaiT2DM  & 15.03 (4.14)   & \underline{11.94 (3.35)}      & 11.96 (3.45)      & 15.60 (4.94)      & \textbf{11.47 (3.31)} & 95.01 (4.04) & \underline{95.26 (5.24)} & 91.44 (8.42) & \textbf{95.91 (4.82)}\\
    \quad UCHTT1DM   & 18.35 (8.40)   & 16.30 (11.28)      & \underline{14.67 (9.10)}      & 16.04 (10.22)      & \textbf{14.41 (9.35)} & 77.15 (21.44) & \textbf{80.08 (19.61)} & 75.18 (24.75) & \underline{79.81 (19.90)}\\
    \quad HUPA-UCM  & 22.27 (6.15)    & \underline{18.07 (5.49)}      & 18.21 (5.45)      & 24.59 (6.58)      & \textbf{17.67 (5.19)} & 87.79 (5.60) & \underline{88.03 (6.97)} & 81.13 (8.24) & \textbf{88.88 (6.93)}\\
    \quad CGMacros  & 19.44 (4.88)    & 16.85 (4.16)      & \underline{16.81 (4.36)}      & 17.42 (4.88)      & \textbf{16.28 (4.18)} & \underline{92.30 (4.33)} & 91.65 (6.76) & 91.98 (5.97) & \textbf{92.43 (5.55)}\\
    \quad T1DM-UOM  & 24.95 (5.17)    & \underline{22.23 (4.92)}      & 22.55 (4.91)      & 25.83 (5.33)      & \textbf{21.89 (4.83)} & 85.38 (4.69) & \underline{85.45 (4.92)} & 81.54 (4.71) & \textbf{86.21 (3.83)}\\
    \quad Bris-T1D  & 22.24 (6.47)    & \underline{19.42 (5.62)}      & 19.64 (5.81)      & 22.81 (6.17)      & \textbf{19.00 (5.58)} & \underline{89.65 (5.53)} & \underline{89.65 (5.71)} & 87.14 (6.05) & \textbf{90.23 (5.23)}\\
    \quad AZT1D  & 23.36 (3.39)    & \underline{20.77 (3.34)}      & 21.23 (3.34)      & 23.83 (3.93)      & \textbf{20.53 (3.29)} & \underline{86.93 (4.23)} & 86.44 (4.52) & 84.41 (4.74) & \textbf{87.45 (4.30)}\\
    \midrule
    \textbf{ControlledAccess} & & & & \\
    \quad OhioT1DM  & 23.42 (2.97)    & \underline{19.72 (2.81)}      & 20.24 (2.46)      & 24.58 (2.89)      & \textbf{19.63 (2.84)} & 88.46 (4.48) & \underline{89.11 (4.02)} & 83.82 (4.63) & \textbf{89.53 (3.41)}\\
    \quad DiaTrend  & 30.46 (5.41)    & \underline{25.29 (4.73)}      & 26.02 (4.96)      & 30.22 (5.90)      & \textbf{24.95 (4.82)} & \underline{87.04 (4.07)}  & 86.50 (4.31) & 82.53 (4.73) & \textbf{87.22 (4.15)}\\
    \quad T1DEXI   & 24.31 (11.92)   & \underline{20.85 (5.03)}      & 21.44 (5.07)      & 24.48 (5.88)      & \textbf{20.70 (4.88)} & \underline{86.11 (5.92)} & 85.63 (6.48) & 82.86 (6.72) & \textbf{86.60 (6.20)}\\
    \midrule
    \textbf{OpenAccess}  & 18.31 (7.19)    & 15.62 (6.61)      & \underline{15.48 (6.18)}      & 18.37 (7.69)      & \textbf{15.01 (6.04)} & 90.56 (9.28) & \underline{91.41 (8.80)} & 87.90 (11.57) & \textbf{91.72 (8.88)}\\
    \textbf{Overall}  & 22.25 (6.87)    & \underline{19.16 (6.34)}      & 19.48 (6.38)      & 22.55 (7.45)      & \textbf{18.75 (5.91)} & 87.86 (7.55) & \underline{87.88 (7.80)} & 84.73 (9.08) & \textbf{88.59 (7.60)}\\
    \bottomrule
  \end{tabular}
  \begin{tablenotes}[flushleft]
    \small
    \item[1] Colas2019 does not have sufficient test sequence length to support zero-shot evaluation with a context length of 12 hours. Results on Colas2019 with smaller context length are included in Fig~\ref{fig:sampling_horizon}.
    \item[*] Bold values indicate the best performance (lowest RMSE for the left block; highest SEG no-risk ratio for the right block), and underlined values indicate the second-best performance.

  \end{tablenotes}
  \end{threeparttable}
  }
\end{table}

From Table \ref{tab:zeroshot_rmse}, we can observe that among the evaluated pre-trained TSFMs, TimesFM2.5~\cite{Das2023timesfm} achieves the strongest overall zero-shot performance on average across all evaluated datasets, with the lowest RMSE of 18.75 mg/dL and the highest SEG no-risk ratio of 88.59\%. Its advantage is consistent across most datasets, including all three controlled-access external test sets, and it substantially improves over the persistence baseline with an overall RMSE of 22.25 mg/dL. Chronos-2 and Moirai2.0 also demonstrate strong zero-shot transfer, often achieving the second-best RMSE or SEG no-risk ratio across datasets. In contrast, Timer~\cite{Liu2024timer} performs less consistently and does not improve over the persistence baseline overall, suggesting that zero-shot transfer varies substantially across TSFM architectures.

When compared with the few-shot and full-shot results in Tables~\ref{tab:fewshot_rmse} and~\ref{tab:fullshot_rmse}, TimesFM remains competitive despite the absence of task-specific adaptation; its overall zero-shot RMSE is only 4.3\% higher than that of the best-performing full-shot LSTM model. These results suggest that general-purpose TSFMs can provide robust zero-shot transfer for blood glucose forecasting, while also establishing a reference point for evaluating the additional benefits of few-shot and full-shot fine-tuning. 


\subsection{Few-shot}

Table~\ref{tab:fewshot_rmse} reports few-shot RMSE and SEG no-risk ratios for supervised deep learning models, pre-trained TSFMs, and LLM-based forecasting models, using 5\% of the open-access CGM training data with a 12-hour context window and 30-minute prediction horizon. Additional results for other metrics and forecasting horizons are provided in Appendix Table~\ref{tab:fewshot_horizon_stack}.

\begin{table*}[ht]
  \caption{Few-shot participant-level RMSE (mg/dL) for each dataset and aggregate RMSE and SEG no-risk ratios (\%) under the 12-hour context, 30-minute horizon setting. Values are participant-level mean (STD); lower RMSE and higher SEG are better. Complete few-shot SEG no-risk ratio is included in Appendix Table~\ref{tab:seg_fewshot_per_dataset}.}
  \label{tab:fewshot_rmse}
  \resizebox{\textwidth}{!}{
  \centering
  \begin{threeparttable}
  \begin{tabular}{lcc|cccc|cc}
    \toprule
    & \multicolumn{2}{c|}{Supervised DL Model}
    & \multicolumn{4}{c|}{Pre-trained TSFM}
    & \multicolumn{2}{c}{TS LLM-based Model} \\
    \cmidrule(lr){2-3}
    \cmidrule(lr){4-7}
    \cmidrule(lr){8-9}
    Dataset
      & LSTM & GPFormer
      & Chronos-2 & Moirai2.0 & Timer & TimesFM2.5
      & TimeLLM & CALF \\
    \midrule
    \textbf{OpenAccess} & & & & & & & & \\
    \quad Hall2018      
    & 14.35 (5.88) & 15.23 (5.45) & \textbf{13.21 (6.66)} & 14.05 (6.14) & 14.50 (6.03) & \underline{13.40 (5.85)} & 15.66 (6.48) & 14.39 (5.22) \\
    
    \quad D1NAMO      
    & 20.52 (12.77) & 24.53 (11.67) & 23.37 (16.01) & \underline{19.90 (10.97)} & 21.74 (12.80) & \textbf{19.87 (12.23)} & 34.42 (21.21) & 23.07 (13.08) \\
    
    \quad Colas2019$^{1}$      
    & -- & -- & -- & -- & -- & -- & -- & -- \\
    
    \quad BIG IDEAS Lab      
    & 16.25 (2.74) & 16.39 (2.12) & \textbf{14.78 (3.06)} & 15.64 (3.04) & 16.46 (2.88) & \underline{15.22 (2.89)} & 17.71 (4.40) & 16.13 (2.99) \\
    
    \quad ShanghaiT1DM      
    & \underline{12.75 (3.28)} & 18.19 (5.18) & 13.38 (3.66) & 12.95 (3.54) & 14.36 (3.86) & \textbf{12.10 (3.03)} & 23.60 (8.33) & 14.64 (4.06) \\
    
    \quad ShanghaiT2DM      
    & \underline{11.70 (3.17)} & 14.76 (4.24) & 12.22 (3.58) & 11.84 (3.42) & 12.50 (3.44) & \textbf{11.40 (3.16)} & 18.71 (6.40) & 13.38 (4.04) \\
    
    \quad UCHTT1DM      
    & 17.26 (8.28) & 18.92 (8.50) & 16.42 (11.82) & \underline{15.84 (9.01)} & 15.84 (8.89) & \textbf{14.53 (8.46)} & 18.62 (11.83) & 18.81 (10.51) \\
    
    \quad HUPA-UCM      
    & \underline{18.38 (5.54)} & 21.91 (4.76) & 18.55 (5.42) & 18.58 (5.41) & 19.89 (5.87) & \textbf{17.86 (5.13)} & 29.25 (8.04) & 19.82 (5.41) \\
    
    \quad CGMacros      
    & 17.64 (4.59) & 17.88 (4.97) & \textbf{16.55 (4.10)} & 17.93 (4.49) & 18.32 (4.87) & \underline{16.73 (4.31)} & 19.37 (5.49) & 17.65 (4.34) \\
    
    \quad T1DM-UOM      
    & \textbf{21.76 (4.40)} & 23.02 (4.71) & \underline{22.17 (4.83)} & 23.33 (4.65) & 23.84 (4.84) & 22.34 (4.70) & 30.13 (6.13) & 23.53 (4.59) \\
    
    \quad Bris-T1D      
    & \textbf{18.84 (5.40)} & 21.51 (4.54) & 19.34 (5.48) & 20.26 (5.97) & 20.71 (5.94) & \underline{19.27 (5.71)} & 26.85 (7.26) & 20.32 (5.65) \\
    
    \quad AZT1D      
    & \underline{20.79 (3.73)} & 21.11 (3.47) & \textbf{20.65 (3.29)} & 22.35 (3.73) & 22.55 (3.69) & 20.86 (3.35) & 27.23 (4.13) & 21.95 (3.53) \\
    
    \midrule
    \textbf{ControlledAccess} & & & & & & & & \\
    \quad OhioT1DM      & \underline{19.76 (2.62)}      & 22.31 (2.80)      & \textbf{19.50 (2.81)}      & 21.30 (3.00)      & 21.83 (2.89)      & 20.59 (2.77)      & 29.49 (3.98)      & 22.47 (3.13) \\
    \quad DiaTrend      & \underline{25.77 (4.90)}      & 28.16 (6.10)      & \textbf{24.96 (4.67)}      & 27.38 (5.15)      & 27.72 (5.14)      & 25.93 (4.77)      & 35.42 (7.04)      & 28.11 (5.32) \\
    \quad T1DEXI      & \underline{20.90 (4.83)}      & 22.28 (5.34)      & \textbf{20.50 (4.91)}      & 22.28 (5.29)      & 22.68 (5.41)      & 21.24 (4.96)      & 28.73 (7.00)      & 22.91 (5.36) \\
    \midrule
    \textbf{OpenAccess RMSE}      & 15.76 (5.93)      & 17.73 (5.83)      & \underline{15.63 (6.48)}      & 15.99 (6.26)      & 16.60 (6.36)      & \textbf{15.20 (5.96)}    & 21.25 (8.75)  & 16.77 (6.14)       \\
    \textbf{OpenAccess SEG(\%)}      &  89.44 (10.59)     &  87.28 (13.39)   &  \underline{90.85 (9.14)}     &  90.68 (9.93)     &  89.89 (10.44)     &  \textbf{91.67 (8.88)}   &  84.52 (10.94) &  89.26 (8.84)   \\
    \textbf{Overall RMSE}      & \underline{19.24 (6.02)}      & 20.94 (6.23)      & \textbf{18.95 (6.16)}      & 20.23 (6.63)      & 20.71 (6.66)      & 19.27 (6.27)   & 26.34 (8.77)   & 20.91 (6.60)       \\
    \textbf{Overall SEG(\%)}      &   87.68 (7.96)      & 86.33 (9.41)    &   \textbf{88.30 (7.33)}   &  87.10 (8.34)     &   86.46 (8.33)    &   \underline{88.07 (7.85)} &  80.64 (9.35)  &   86.14 (7.60)     \\
    \bottomrule
  \end{tabular}
  \begin{tablenotes}[flushleft]
    \footnotesize
    \item[1] The Colas2019 dataset does not have sufficient test sequence length to support forecasting with 12-h context length. 
    \item[*] Bold values indicate the best performance (lowest RMSE), and underlined values indicate the second-best performance.
    
  \end{tablenotes}
  \end{threeparttable}
  }
\end{table*}

Pre-trained TSFMs emerge as the strongest model category under limited-data adaptation. Chronos-2 achieves the best overall RMSE of 18.95 mg/dL and the highest overall SEG no-risk ratio of 88.30\%, while TimesFM achieves the best OpenAccess RMSE of 15.20 mg/dL and OpenAccess SEG no-risk ratio of 91.67\%. Chronos-2 also achieves the lowest RMSE on all three controlled-access external test sets, suggesting strong transfer beyond the open-access training datasets. The supervised LSTM baseline remains highly competitive, obtaining the second-best overall RMSE of 19.24 mg/dL. Among LLM-based methods, CALF substantially outperforms TimeLLM, but both remain less competitive than the best TSFMs and LSTM baseline under limited-data adaptation. Additional prompt ablation and personalized-prompt case studies for TimeLLM are reported in Appendix~\ref{timellm_prompts} and \ref{sec:timellm_ablation_results}; these variants did not consistently improve TimeLLM forecasting performance. Overall, these results indicate that pre-trained TSFMs can adapt effectively with limited glucose-specific data.

\subsection{Full-shot}

Table~\ref{tab:fullshot_rmse} summarizes full-shot results under the same 12-hour context window and 30-minute prediction horizon. Additional full-shot results across other metrics and forecasting horizons are provided in Appendix Table~\ref{tab:fullshot_horizon_stack}. Statistical tests are provided in Appendix~\ref{sec:statistical_test_model}.

\begin{table*}[h]
  \caption{Full-shot participant-level RMSE (mg/dL) for each dataset and aggregate RMSE and SEG no-risk ratios (\%) under the 12-hour context, 30-minute horizon setting. Values are participant-level mean (STD); lower RMSE and higher SEG are better. Complete full-shot SEG no-risk ratio is included in Appendix Table~\ref{tab:seg_fullshot_per_dataset}.}
  \label{tab:fullshot_rmse}
  \centering
  \resizebox{\textwidth}{!}{
  \begin{threeparttable}
  \begin{tabular}{lcc|cccc|cc}
    \toprule
    & \multicolumn{2}{c|}{Supervised DL Model}
    & \multicolumn{4}{c|}{Pre-trained TSFM}
    & \multicolumn{2}{c}{TS LLM-based Model} \\
    \cmidrule(lr){2-3}
    \cmidrule(lr){4-7}
    \cmidrule(lr){8-9}
    Dataset
      & LSTM & GPFormer
      & Chronos-2 & Moirai2.0 & Timer & TimesFM2.5
      & TimeLLM & CALF \\
    \midrule
    \textbf{OpenAccess} & & & & & & & & \\
    \quad Hall2018      
    & \textbf{12.94 (5.72)} & 13.41 (6.06) & \underline{13.04 (6.27)} & 14.79 (5.87) & 13.91 (5.60) & 13.62 (5.82) & 14.64 (5.36) & 14.40 (4.81) \\
    
    \quad D1NAMO      
    & \textbf{18.65 (12.38)} & 21.48 (12.35) & 20.73 (13.57) & \underline{18.86 (11.45)} & 19.36 (12.30) & 19.82 (11.57) & 29.12 (16.96) & 23.27 (16.37) \\
    
    \quad Colas2019$^{1}$     
    & -- & -- & -- & -- & -- & -- & -- & -- \\
    
    \quad BIG IDEAS Lab      
    & \textbf{14.65 (2.75)} & 15.67 (2.67) & \textbf{14.65 (2.97)} & 15.50 (3.12) & 15.51 (2.63) & \underline{15.26 (2.92)} & 16.69 (3.48) & 16.02 (3.10) \\
    
    \quad ShanghaiT1DM      
    & \textbf{11.00 (2.88)} & 13.86 (3.54) & 13.00 (3.63) & \underline{11.73 (2.91)} & 11.81 (3.14) & 12.17 (3.27) & 19.34 (6.34) & 13.39 (3.47) \\
    
    \quad ShanghaiT2DM      
    & \textbf{10.57 (2.94)} & 12.29 (3.30) & 11.80 (3.55) & \underline{11.16 (3.30)} & 11.29 (3.25) & 11.25 (3.16) & 15.55 (5.10) & 12.13 (3.43) \\
    
    \quad UCHTT1DM      
    & \textbf{14.26 (8.28)} & \underline{15.31 (9.15)} & 15.94 (10.73) & 16.40 (9.07) & 15.93 (7.27) & 15.34 (8.47) & 17.93 (10.68) & 18.48 (10.04) \\
    
    \quad HUPA-UCM      
    & \textbf{16.71 (4.82)} & 18.65 (4.86) & 18.35 (5.20) & 18.12 (5.30) & 17.89 (5.18) & \underline{17.87 (5.01)} & 23.99 (6.21) & 18.54 (5.63) \\
    
    \quad CGMacros      
    & \underline{16.37 (4.21)} & 17.38 (4.22) & \textbf{16.16 (4.14)} & 17.96 (4.60) & 17.45 (4.15) & 16.99 (4.41) & 18.03 (4.67) & 17.52 (4.37) \\
    
    \quad T1DM-UOM      
    & \textbf{20.81 (4.26)} & \underline{21.48 (4.39)} & 21.78 (4.49) & 23.23 (4.54) & 22.27 (4.85) & 22.38 (4.48) & 25.95 (5.19) & 22.96 (4.46) \\
    
    \quad Bris-T1D      
    & \textbf{17.83 (5.28)} & \underline{18.71 (5.27)} & 19.05 (5.44) & 20.32 (6.11) & 19.45 (5.86) & 19.53 (5.60) & 22.69 (6.11) & 19.54 (5.58) \\
    
    \quad AZT1D      
    & \textbf{19.43 (3.18)} & \underline{20.34 (3.42)} & 20.47 (3.25) & 22.71 (3.98) & 21.54 (3.96) & 21.30 (3.51) & 23.36 (3.51) & 21.69 (3.79) \\
    \midrule
    \textbf{ControlledAccess} & & & & & & & & \\
    \quad OhioT1DM      & \textbf{18.78 (2.43)}      & 19.94 (2.93)      & \underline{19.58 (2.41)}      & 21.73 (3.13)      & 20.11 (2.79)      & 20.48 (2.63)      & 24.56 (3.18)      & 21.65 (3.33) \\
    \quad DiaTrend      & \textbf{24.20 (4.55)}      & \underline{24.83 (4.69)}      & 25.00 (4.67)      & 27.68 (5.13)      & 26.10 (4.86)      & 26.27 (4.73)      & 30.47 (5.90)      & 27.47 (5.24) \\
    \quad T1DEXI      & \textbf{19.70 (4.58)}      & 20.53 (4.70)      & \underline{20.44 (4.85)}      & 22.46 (5.30)      & 21.37 (5.05)      & 21.49 (4.96)      & 24.75 (5.94)      & 22.25 (5.21) \\
    \midrule
    \textbf{OpenAccess RMSE}      & \textbf{14.37 (5.70)}      & 15.66 (5.81)      & \underline{15.26 (6.16)}      & 15.85 (6.38)      & 15.47 (5.97)      & 15.32 (6.00)   & 18.50 (7.08)   & 16.18 (6.25)       \\
    \textbf{OpenAccess SEG(\%)}      &  90.31 (9.42)     &  89.19 (11.28)   &  \textbf{91.52 (8.40)}     &  90.66 (9.95)     &  90.95 (9.38)     &  \underline{91.07 (10.24)}   &  87.62 (9.60) &  89.88 (8.57)   \\
    \textbf{Overall RMSE}      & \textbf{17.97 (5.82)}      & 18.97 (5.80)      & \underline{18.79 (6.04)}      & 20.30 (6.76)      & 19.44 (6.29)      & 19.47 (6.32)  & 22.75 (7.28)    & 20.28 (6.55)      \\
    \textbf{Overall SEG(\%)}      &  \underline{88.32 (7.31)}   & 87.38 (8.37)    &   \textbf{88.62 (7.06)}   &  87.04 (8.39)     &   87.74 (7.72)   &  87.65 (8.40) &  84.35 (8.22)  &  86.93 (7.35)    \\
    \bottomrule
  \end{tabular}
  \begin{tablenotes}[flushleft]
    \footnotesize
    \item[1] The Colas2019 dataset does not have sufficient test sequence length to support forecasting with 12-h context length. 
    \item[*] Bold values indicate the best performance (lowest RMSE), and underlined values indicate the second-best performance.
  \end{tablenotes}
  \end{threeparttable}
  }
\end{table*}

In the full-shot setting, the supervised LSTM model achieves the strongest predictive performance, with the lowest overall RMSE of 17.97 mg/dL. It outperforms other models on most open-access datasets and all three controlled-access external test sets, indicating that task-specific architectures remain highly effective when sufficient glucose-specific training data are available. Chronos-2 remains competitive as the closest-performing TSFM, with an overall RMSE of 18.79 mg/dL, only 0.82 mg/dL higher than LSTM. These results suggest that full-shot LSTM establishes the predictive-performance upper bound in this benchmark.

\begin{wrapfigure}{r}{0.52\textwidth}
  \centering
  \includegraphics[width=0.52\textwidth]{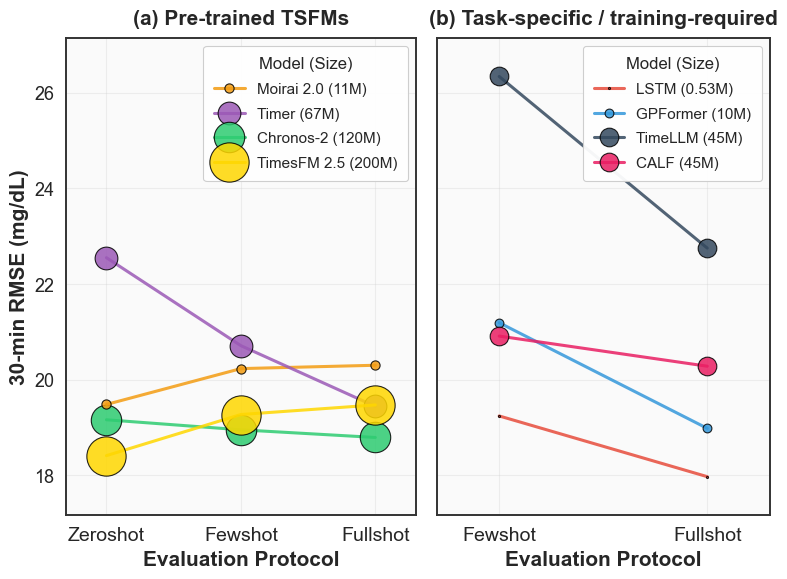}
  \caption{Overall 30-minute RMSE across evaluation protocols, with marker size indicating model parameter count.}
  \label{fig:size_regime}
\end{wrapfigure}

Comparing few-shot and full-shot regimes, most models benefit from additional training data, including supervised deep learning models, Chronos-2, Timer, TimeLLM, and CALF, as summarized in Figure~\ref{fig:size_regime}.
However, these gains are not uniform across architectures or model scales.
Under the shared fine-tuning budget, TimesFM and Moirai2.0 do not show consistent gains from full-shot fine-tuning: their overall RMSE changes from 18.75 to 19.47 mg/dL and from 19.48 to 20.30 mg/dL, respectively.
Their learning curves (Appendix Figure~\ref{fig:fullshot_curves}) show that training loss decreases steadily throughout fine-tuning, whereas validation loss decreases only within the first epoch and then fluctuates without a clear downward trend.
This indicates mild overfitting that sets in early: for these two models, additional full-shot training mainly fits the training data and yields limited task-specific adaptation rather than improved generalization.
Figure~\ref{fig:size_regime} further shows that larger model size does not necessarily translate into better full-shot glucose forecasting performance, suggesting that adaptation behavior depends on model design in addition to scale. This degradation is consistent with prior studies' findings~\cite{Li2025}. 
Appendix Tables~\ref{tab:zeroshot_pi80}--\ref{tab:fullshot_pi80} also show that, under the same protocol, fine-tuning does not uniformly improve probabilistic calibration: Chronos-2 maintains stable PI80 coverage, whereas Moirai2.0 and TimesFM2.5 show less stable coverage after fine-tuning.

\subsection{Stratified Performance Analysis}

\begin{figure}[!ht]
  \centering
  \includegraphics[width=\linewidth]{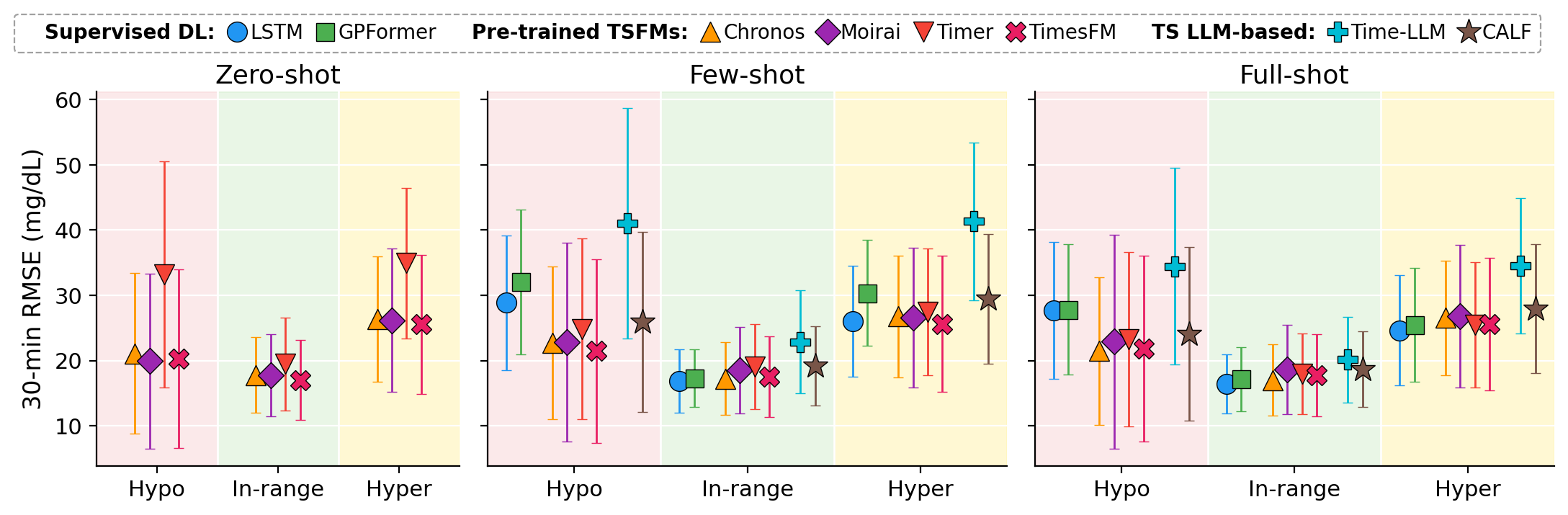}
  \caption{Glycemic-stratified 30-minute RMSE under zero-, few-, and full-shot evaluation, with points and error bars showing per-participant mean and standard deviation across glycemic ranges. Detailed values are reported in Appendix Table~\ref{tab:glycemic_stratified_rmse}.}
  \label{fig:stratified_glycemic}
\end{figure}

Aggregate RMSE can obscure performance differences across clinically distinct glucose ranges. 
We therefore stratify prediction points by ground-truth CGM values into hypoglycemia (<70 mg/dL), in-range glucose (70--180 mg/dL), and hyperglycemia (>180 mg/dL). 
Figure~\ref{fig:stratified_glycemic} summarizes 30-minute RMSE across these ranges under zero-shot, few-shot, and full-shot protocols.

Across evaluation protocols, RMSE is consistently lowest in the in-range condition and substantially higher in hypo- and hyperglycemic ranges, indicating greater forecasting challenges in these clinically important ranges.
Notably, pre-trained TSFMs perform the best in the hypoglycemic range across three protocols, whereas supervised deep learning and TS LLM-based methods generally show higher errors. 
This result suggests that pre-trained TSFMs may retain advantages in hypoglycemia ranges, even when no or limited glucose-specific training data are available.

One possible explanation is the relative sparsity of hypoglycemic records. 
As shown in Appendix Table~\ref{tab:test_glucose_distribution}, hypoglycemic samples account for only 2.4\% of all evaluation records in the test split, compared with 61.7\% in-range and 35.9\% hyperglycemic samples. This distribution is consistent with prior studies showing that time below range is typically much smaller than time in range or time above range~\cite{Battelino2019, Danne2017}. Pre-trained TSFMs may therefore benefit from transferable representations learned from diverse time-series corpora, improving robustness in less frequent glucose ranges. 
In contrast, supervised DL models and TS LLM-based models may tend to fit more frequently observed glucose patterns and therefore perform worse on rare hypoglycemic events.


We also perform a cohort-level stratified analysis by diabetes type (ND, PreD, T2D, and T1D) to assess robustness across heterogeneous populations, as shown in Appendix Figures~\ref{fig:rmse-by-phenotype}. Overall, T1D cohorts show higher forecasting errors, consistent with higher glycemic variability, while relative model rankings remain broadly similar across cohorts. Beyond diabetes type, we further examine how forecasting performance varies with temporal glucose dynamics and individual glycemic variability (Appendix Tables~\ref{tab:time_of_day} and~\ref{tab:cv_strata}). Errors are lowest overnight, increase substantially during post-meal periods, and are higher among participants with greater glycemic variability, highlighting the influence of both temporal and individual glucose dynamics on forecasting difficulty.

\subsection{Limitations}


Although there are several important findings from this study, there are also a few limitations. First, GlucoFM-Bench focuses on univariate CGM forecasting without covariates such as carbohydrate intake or insulin delivery. Multivariate glucose forecasting is important but requires standard, high-quality, consistently recorded covariates, which remain rare across public CGM datasets. In addition, because only Chronos-2 and Moirai2.0 currently support multivariate inputs among the evaluated TSFMs, a fair cross-model comparison is infeasible. We therefore stick to univariate CGM forecasting benchmarking as a necessary foundation for future standardized multivariate benchmarking.

Second, the TSFM landscape is evolving rapidly, and GlucoFM-Bench does not exhaustively cover all recently proposed foundation-model architectures or model-specific configurations. We evaluate representative TSFMs and LLM-based forecasting frameworks spanning major modeling paradigms, while leaving newer architectures and additional tuning strategies to future benchmark updates.

Although glucose forecasting has direct application pathways in diabetes technologies such as predictive CGM alerts and automated insulin delivery systems, GlucoFM-Bench is a retrospective research benchmark, not a clinical decision system; clinical deployment of benchmarked models would require prospective validation and regulatory review.

\section{Conclusion}
We introduced GlucoFM-Bench, a standardized benchmark for evaluating time-series foundation models in continuous glucose forecasting across 15 diabetes-relevant CGM datasets. By comparing pre-trained TSFMs, LLM-based time-series forecasting methods, and task-specific deep learning models under zero-shot, few-shot, and full-shot protocols, GlucoFM-Bench enables systematic evaluation of model generalization, adaptation, and clinical relevance across heterogeneous glucose populations. Our results show that pre-trained TSFMs, especially Chronos-2 and TimesFM, achieve strong zero-shot and few-shot transfer, supporting their potential for low-resource glucose forecasting. In contrast, under full-shot training, a lightweight task-specific LSTM achieves the strongest predictive performance, highlighting the continued value of domain-specific supervised learning when sufficient data are available. Stratified analyses further reveal persistent forecasting challenges in clinically critical hypo- and hyperglycemic ranges and T1D cohorts. Overall, GlucoFM-Bench provides a reproducible and extensible foundation for advancing foundation-model-based glucose forecasting and supporting more reliable, clinically aware, and equitable evaluation in future diabetes management research.

\begin{ack}

\textbf{Acknowledgment.} This publication is based on research using the T1DEXI dataset from Jaeb Center for Health Research Foundation, that has been made available through Vivli, Inc. Vivli has not contributed to or approved, and is not in any way responsible for, the contents of this publication.

\textbf{Funding.} The authors acknowledge internal funding from Dartmouth College and Emory University which supports presentation of this work.

\textbf{Competing interests.} The authors declare no competing interests.

\end{ack}

\clearpage 

\bibliographystyle{unsrt}
\bibliography{sections/6_reference}

\clearpage 
\appendix
\title{Appendix}

\section{Related work}
\label{sec:related_work}

\subsection{Time Series Foundation Model} 
Recent advances in TSFMs have enabled transferable temporal representations via large-scale pretraining on heterogeneous time series data, supporting generalization across diverse forecasting tasks. These models exhibit strong zero-shot forecasting performance, often matching or surpassing fully supervised models on unseen datasets without task-specific fine-tuning \cite{Li2025}. These models typically follow one of three paradigms: (1) decoder-only architectures (e.g., Timer~\cite{Liu2024timer}, TimesFM~\cite{Das2023timesfm}, Moirai 2.0~\cite{Liu2025moirai2}, Time-MoE~\cite{Shi2024timemoe}) using autoregressive prediction; (2) encoder-only models (e.g., Moirai 1.0~\cite{Woo2024}, YingLong~\cite{Wang2025yinglong}) leveraging masked token recovery; and (3) encoder-decoder frameworks (e.g., Chronos~\cite{Ansari2024chronos}) that frame forecasting as sequence-to-sequence translation. Together, they demonstrate the architectural diversity under a shared pretraining paradigm.

In parallel, the rapid advancement of large language models (LLMs) has motivated a new class of architectures that adapt LLMs for time-series forecasting. LLM-based forecasting has progressed from structural reuse of frozen transformers (e.g., GPT4TS~\cite{Zhou2023gpt4ts}) to semantic reprogramming and prompt-based conditioning (e.g., LLMTime~\cite{Gruver2023LLMtime}, Time-LLM~\cite{Jin2023timellm}). Recent models like CALF~\cite{Liu2024calf} and UniTime~\cite{Liu2023UniTime} further enhance generalization through cross-modal alignment and domain-instructed prompts, marking a shift from passive adaptation to active temporal reasoning with LLMs.

\subsection{Blood Glucose Forecasting Models} 
In the domain of blood glucose prediction, the majority of prior work has focused on traditional statistical and machine learning approaches (e.g., ARIMA, linear regression) \cite{Singh2025, Krishnamoorthy2024, Shuvo2023}, as well as deep learning models such as CNNs, LSTMs, and GRUs \cite{Martinsson2019, Deng2021}. These models typically utilize CGM data, either alone or in combination with auxiliary signals such as insulin dosage and meal intake. More recently, the success of Transformer-based architectures in general time series forecasting has motivated their application to glucose prediction. Notable examples include the Glucose Transformer proposed by Lee et al.~\cite{lee2023} and GPFormer by Zhu et al.~\cite{Zhu2025}.

With the growing interest in foundation models and large language models (LLMs), several studies have begun exploring their applications in diabetes management and glucose forecasting. Lutsker et al.~\cite{Lutsker2026} introduced GluFormer, a CGM-specific foundation model designed for HbA1c prediction, treatment outcome forecasting, and long-term diabetes risk assessment. Luo et al.~\cite{Luo2025} proposed a transformer-decoder–based sensor foundation model pre-trained on CGM data to forecast 2-hour-ahead glucose levels. Li et al.~\cite{Li2025LLM} and Lara-Abelenda et al.~\cite{Lara-Abelenda2025} developed multivariate and univariate Time-LLM–based frameworks respectively, targeting on glucose forecasting. While these efforts demonstrate the promise of domain-specific foundation models and adaptation frameworks for glucose prediction, their focus is primarily on glucose-specific model development or clinical outcome prediction, rather than the systematic evaluation of general-purpose TSFMs under standardized forecasting protocols. Consequently, the performance and robustness of general-purpose TSFMs on glucose forecasting remain largely unexplored, representing the critical gap this work aims to address.

\clearpage
\section{Participants Metadata and Cohort Summary }
\label{sec:meta_cohort_summary}
\begin{table*}[ht]
\centering
\caption{Demographic characteristics of the datasets. NR indicates not reported.}
\label{tab:dataset_demographics}
\small
\setlength{\tabcolsep}{4pt}
\renewcommand{\arraystretch}{1.15}
\begin{tabularx}{\textwidth}{l c c c X c c}
\toprule
\multirow{2}{*}{Dataset} & Gender & \multicolumn{2}{c}{Age (yrs.)} & \multirow{2}{*}{Race/Ethnicity} & \multicolumn{2}{c}{Hemoglobin A1C (\%)} \\
\cmidrule(lr){3-4} \cmidrule(lr){6-7}
& M / F / Unknown & Mean (SD) & Range & & Mean (SD) & Range \\
\midrule
Hall\_2018        & 25 / 32 / 0   & 48.9 (13.8) & 25--76 & NR & 5.4 (0.4) & 4.6--6.7 \\
D1NAMO            & 6 / 3 / 0      & NR          & 20--79 & NR & NR        & NR \\
Colas\_2019       & 103 / 105 / 0 & 59.6 (10.1) & 29--88 & NR & 5.8 (0.3) & 5.1--6.5 \\
BIGIDEAs          & 7 / 9 / 0      & NR          & 35--65 & 68.75\% White, 25\% Black/African Am. & 5.7 (0.3) & 5.3--6.4 \\
ShanghaiT1DM      & 5 / 7 / 0      & 57.8 (11.1) & 37--73 & NR & 10.3 (3.4) & 7.1--17.3 \\
ShanghaiT2DM      & 44 / 56 / 0   & 60.2 (13.7) & 22--97 & NR & 9.1 (2.5) & 4.3--15.4 \\
UCHTT1DM          & 10 / 10 / 0   & 27.2 (4.0)  & NR     & NR & NR & 4.4--8.2 \\
HUPA-UCM          & 12 / 13 / 0   & 39.2 (12.1) & 18--62 & NR & 7.4 (0.8) & 6.0--9.7 \\
CGMacros\_Dexcom  & 16 / 29 / 0   & 48.1 (12.7) & 18--69 & 75.5\% Hispanic, 15.5\% White, 8.8\% Black/African Am. & 6.1 (0.9) & 4.6--8.5 \\
T1D-UOM           & 7 / 10 / 0    & 43.5 (15.3) & 23--70 & NR & NR & NR \\
Bris-T1D\_Open    & 3 / 15 / 2    & 22.3 (2.0)  & 18--26 & 85\% White & NR & NR \\
AZT1D             & 12 / 13 / 0   & 59.2 (15.1) & 27--80 & NR & 6.6 (0.7) & 5.0--8.2 \\
\midrule
OhioT1DM          & 7 / 5 / 0      & NR          & 20--80 & NR & NR & NR \\
DiaTrend          & 17 / 37 / 0   & 32.1 (15.0) & 19--74 & 88.9\% White & 7.8 (0.8) & 6.3--10.0 \\
T1DEXI            & 134 / 363 / 0 & 36.7 (14.0) & 18--70 & 91.3\% White & 6.6 (0.8) & 4.8--10.0 \\
\bottomrule
\end{tabularx}
\end{table*}

\begin{table}[h]
\centering
\caption{Summary statistics of blood glucose characteristics across diabetes cohorts, including only participants with sufficient data for evaluation (mean $\pm$ STD).}
\label{tab:cohort_management_summary}
\resizebox{0.8\textwidth}{!}{
\begin{tabular}{lcccc}
\toprule
\textbf{Cohort (Num.)} & \textbf{Avg BG (mg/dL)} & \textbf{Avg Daily TIR (\%)} & \textbf{Avg Daily GV}  & \textbf{Periodicity}\\
\midrule
No Diabetes (67)
& 104.34 $\pm$ 16.41 
& 94.90\% $\pm$ 6.04\% 
& 14.82 $\pm$ 3.15 
& 0.27 $\pm$ 0.14\\

Prediabetes (38)
& 122.05 $\pm$ 17.53 
& 94.15\% $\pm$ 7.70\%
& 17.15 $\pm$ 3.40 
& 0.28 $\pm$ 0.13\\

T2D (118)
& 143.04 $\pm$ 31.37 
& 76.56\% $\pm$ 18.88\%
& 23.94 $\pm$ 5.36 
& 0.49 $\pm$ 0.17\\

T1D (672)
& 151.06 $\pm$ 26.39 
& 71.14\% $\pm$ 16.08\% 
& 29.70 $\pm$ 5.24 
& 0.27 $\pm$ 0.10\\
\bottomrule
\end{tabular}
}
\end{table}

\begin{table}[ht]
  \caption{Ground-truth glucose-range distribution of the \textbf{test split} of each dataset. Counts and percentages are computed over held-out CGM values only (no training data).}
  \label{tab:test_glucose_distribution}
  \centering
  \resizebox{\textwidth}{!}{
  \begin{threeparttable}
  \begin{tabular}{lrccc}
    \toprule
    Dataset & \textbf{N} & \textbf{$<$70 mg/dL (hypo)} & \textbf{70--180 mg/dL (target)} & \textbf{$>$180 mg/dL (hyper)} \\
    \midrule
    \textbf{OpenAccess} & & & & \\
    \quad Hall2018       & 19{,}577    & 951 (4.86\%)       & 18{,}379 (93.88\%)  & 247 (1.26\%)        \\
    \quad D1NAMO         & 1{,}617     & 88 (5.44\%)        & 967 (59.80\%)       & 562 (34.76\%)       \\
    \quad Colas2019      & 12{,}064    & 233 (1.93\%)       & 11{,}722 (97.17\%)  & 109 (0.90\%)        \\
    \quad BIG IDEAS Lab  & 7{,}394     & 50 (0.68\%)        & 7{,}188 (97.21\%)   & 156 (2.11\%)        \\
    \quad ShanghaiT1DM   & 9{,}417     & 853 (9.06\%)       & 5{,}340 (56.71\%)   & 3{,}224 (34.24\%)   \\
    \quad ShanghaiT2DM   & 67{,}486    & 2{,}213 (3.28\%)   & 55{,}611 (82.40\%)  & 9{,}662 (14.32\%)   \\
    \quad UCHTT1DM       & 5{,}923     & 593 (10.01\%)      & 4{,}894 (82.63\%)   & 436 (7.36\%)        \\
    \quad HUPA-UCM       & 61{,}889    & 4{,}714 (7.62\%)   & 44{,}242 (71.49\%)  & 12{,}933 (20.90\%)  \\
    \quad CGMacros       & 25{,}441    & 35 (0.14\%)        & 21{,}775 (85.59\%)  & 3{,}631 (14.27\%)   \\
    \quad T1DM-UOM       & 89{,}844    & 2{,}161 (2.41\%)   & 67{,}815 (75.48\%)  & 19{,}868 (22.11\%)  \\
    \quad Bris-T1D       & 219{,}923   & 5{,}277 (2.40\%)   & 147{,}212 (66.94\%) & 67{,}434 (30.66\%)  \\
    \quad AZT1D          & 60{,}288    & 986 (1.64\%)       & 47{,}697 (79.12\%)  & 11{,}605 (19.25\%)  \\
    \midrule
    \textbf{ControlledAccess} & & & & \\
    \quad OhioT1DM       & 31{,}994    & 876 (2.74\%)       & 19{,}860 (62.07\%)  & 11{,}258 (35.19\%)  \\
    \quad DiaTrend       & 1{,}220{,}222 & 19{,}943 (1.63\%) & 589{,}095 (48.28\%) & 611{,}184 (50.09\%) \\
    \quad T1DEXI         & 750{,}174   & 22{,}203 (2.96\%)  & 552{,}746 (73.68\%) & 175{,}225 (23.36\%) \\
    \midrule
    \textbf{Overall Open}  & 580{,}863   & 18{,}154 (3.13\%)  & 432{,}842 (74.52\%)   & 129{,}867 (22.36\%)  \\
    \textbf{Overall}       & 2{,}583{,}253 & 61{,}176 (2.37\%) & 1{,}594{,}543 (61.73\%) & 927{,}534 (35.91\%) \\
    \bottomrule
  \end{tabular}
  \begin{tablenotes}[flushleft]
    \footnotesize
    \item[*] Hypoglycemia (\textless{}70 mg/dL), target range (70--180 mg/dL), and hyperglycemia (\textgreater{}180 mg/dL) bins follow standard clinical CGM reporting (Battelino et al., 2019).
    \item[*] \textbf{Overall Open} aggregates the 12 OpenAccess datasets; \textbf{Overall} aggregates all 15 datasets (OpenAccess + ControlledAccess).
  \end{tablenotes}
  \end{threeparttable}
  }
\end{table}

\clearpage
\section{Implementation Details}
\label{sec:implementation_details}
All experiments were conducted using GPU-accelerated environments. All of the evaluation experiments were implemented on Google Colab NVIDIA A100 GPUs and a local workstation with an NVIDIA RTX 4080 GPU.

\subsection{Glucose Management Metrics Calculation}
\label{sec:appendix_calculation}
In Table~\ref{tab:dataset_overview}, we summarize dataset-level glucose management and data quality metrics computed using the definitions below.

\paragraph{Data completeness.}
We quantify missingness in CGM recordings using the missing ratio:
\begin{equation}
\text{Missing Ratio}
=
1 - \frac{N_{\text{records}} \times \Delta t}{T_{\text{end}} - T_{\text{start}}},
\end{equation}
where $N_{\text{records}}$ is the total number of CGM measurements, $\Delta t$ denotes the sampling interval (e.g., 5 minutes), and $T_{\text{start}}$ and $T_{\text{end}}$ are the start and end timestamps of the recording period, respectively.

\paragraph{Time in range (TIR)}
Average daily time in range (70–180 mg/dL) is computed across individuals as:
\begin{equation}
\text{Avg. Daily TIR}
=
\frac{1}{N} \sum_{i=1}^{N}
\left(
\frac{1}{D_i} \sum_{d=1}^{D_i}
\frac{n_{i,d}^{[70,180]}}{n_{i,d}}
\right),
\end{equation}
where $N$ is the number of individuals, $D_i$ denotes the number of recorded days for individual $i$, $n_{i,d}^{[70,180]}$ is the number of glucose measurements within 70–180 mg/dL on day $d$ for individual $i$, and $n_{i,d}$ is the total number of glucose measurements on that day.

\paragraph{Average daily glucose level.}
Average daily blood glucose (BG) is calculated as:
\begin{equation}
\text{Avg. Daily BG}
=
\frac{1}{N} \sum_{i=1}^{N}
\left(
\frac{1}{D_i} \sum_{d=1}^{D_i}
\frac{1}{n_{i,d}} \sum_{k=1}^{n_{i,d}} G_{i,d,k}
\right),
\end{equation}
where $G_{i,d,k}$ denotes the $k$-th glucose measurement on day $d$ for individual $i$.

\paragraph{Glycemic variability.}
Daily glycemic variability (GV) is quantified using the coefficient of variation and averaged across valid days:
\begin{equation}
\text{Avg. Daily GV}
=
\frac{1}{D} \sum_{d=1}^{D}
\left(
\frac{\sigma_d}{\mu_d} \times 100
\right),
\end{equation}
where $D$ is the number of valid days,
\begin{equation}
\mu_d = \frac{1}{n_d} \sum_{k=1}^{n_d} G_{d,k},
\qquad
\sigma_d =
\sqrt{
\frac{1}{n_d - 1}
\sum_{k=1}^{n_d}
(G_{d,k} - \mu_d)^2
},
\end{equation}
and $G_{d,k}$ denotes the $k$-th glucose measurement on day $d$.
The $n_d - 1$ denominator corresponds to the sample standard deviation, consistent with the default implementation in \texttt{pandas}.

\paragraph{Circadian periodicity.}
To characterize longer-term periodic patterns in glucose dynamics, we compute circadian periodicity based on STL decomposition:
\begin{equation}
\text{Circadian periodicity}
=
\max\!\left(
0,\;
1 - \frac{\mathrm{Var}(R)}{\mathrm{Var}(S + R)}
\right),
\end{equation}
where $S$ and $R$ denote the seasonal and remainder (residual) components obtained from STL decomposition with a fixed period (24 hours), and $\mathrm{Var}(\cdot)$ denotes the variance operator. The $\max(0,\cdot)$ term ensures the resulting value is non-negative.

\subsection{Model Summary}
Table~\ref{tab:model_summary} summarizes the eight evaluated models included in GlucoFM-Bench. For fine-tuning, we use model-appropriate training configurations following official recommended implementations or prior glucose forecasting studies when available. To ensure comparability, all models are evaluated under the same participant-level chronological splits, context lengths, prediction horizons, and metrics.

\begin{table}[h]
\centering
\caption{Summary of forecasting models evaluated in GlucoFM-Bench.}
\label{tab:model_summary}
\small
\resizebox{\textwidth}{!}{
\begin{tabular}{llllll}
\toprule
\textbf{Model} & \textbf{Category} & \textbf{Backbone} & \textbf{Input} & \textbf{Forecast Type} & \textbf{Setting} \\
\midrule
Persistence & Baseline & Last-value & CGM & Point & Zero-shot \\
LSTM & Supervised DL & LSTM & CGM & Point & Few-/Full-shot \\
GPFormer & Supervised DL & Transformer & CGM & Point & Few-/Full-shot \\
Chronos-2 & Pre-trained TSFM & Enc.-only. TF & CGM & Prob. / Patch & Zero-/Few-/Full-shot \\
Moirai 2.0 & Pre-trained TSFM & Dec.-only TF & CGM & Prob. / Patch & Zero-/Few-/Full-shot \\
Timer & Pre-trained TSFM & Dec.-only TF & CGM & Point / Patch & Zero-/Few-/Full-shot \\
TimesFM 2.5 & Pre-trained TSFM & Dec.-only TF & CGM & Prob. / Patch & Zero-/Few-/Full-shot \\
TimeLLM & TS LLM-based & GPT-2 reprogram. & CGM + prompt & Point / Patch & Few-/Full-shot \\
CALF & TS LLM-based & GPT-2 cross-modal & CGM + text & Point / Patch & Few-/Full-shot \\
\bottomrule
\end{tabular}
}
\parbox{\textwidth}{\footnotesize{\textit{Note:} TF denotes Transformer; Prob. denotes probabilistic forecasting.}}
\end{table}

\begin{table*}[t]
\centering
\caption{Training hyperparameters for each model family in full-shot and few-shot regimes.
All methods use a 144-step (12\,h) context window and evaluate at horizons of 15/30/60/90\,min.
``Train stride'' is the sliding-window stride used to build the training set
(few-shot uses a much larger stride to subsample the training pool).
``ES'' = early stopping (patience in epochs).}
\label{tab:hparams}
\resizebox{\textwidth}{!}{%
\begin{tabular}{l l c c c c c c}
\toprule
\textbf{Model family} & \textbf{Regime} & \textbf{Batch} & \textbf{LR} & \textbf{Optimizer} & \textbf{Epochs} & \textbf{Train stride} & \textbf{Early stopping} \\
\midrule
\multirow{2}{*}{LSTM (Martinsson 2019)}
  & Full-shot & 1024 & $1\!\times\!10^{-3}$ & Adam (NLL) & 1000\,(cap) & 12  & ES, patience\,=\,50 \\
  & Few-shot  & 1024 & $1\!\times\!10^{-3}$ & Adam (NLL) & 1000\,(cap) & 240 & ES, patience\,=\,50 \\
\midrule
\multirow{2}{*}{GPFormer}
  & Full-shot & 16 & $1\!\times\!10^{-3}$ & AdamW (wd $10^{-2}$) & 30 & 12  & ES, patience\,=\,10 \\
  & Few-shot  & 16 & $1\!\times\!10^{-3}$ & AdamW (wd $10^{-2}$) & 30 & 240 & ES, patience\,=\,10 \\
\midrule
\multirow{2}{*}{TimesFM }
  & Full-shot & 16 & $1\!\times\!10^{-5}$ & AdamW & 30 & 12  & ES, patience\,=\,10 \\
  & Few-shot  & 16 & $1\!\times\!10^{-5}$ & AdamW & 30 & 240 & ES, patience\,=\,10 \\
\midrule
\multirow{2}{*}{Timer}
  & Full-shot & 16 & $1\!\times\!10^{-5}$ & AdamW & 30 & 12  & ES, patience\,=\,10 \\
  & Few-shot  & 16 & $1\!\times\!10^{-5}$ & AdamW & 30 & 240 & ES, patience\,=\,10 \\
\midrule
\multirow{2}{*}{Moirai }
  & Full-shot & 16 & $1\!\times\!10^{-5}$ & AdamW & 30 & 12  & ES, patience\,=\,10 \\
  & Few-shot  & 16 & $1\!\times\!10^{-5}$ & AdamW & 30 & 240 & ES, patience\,=\,10 \\
\midrule
\multirow{2}{*}{Chronos-2 }
  & Full-shot & 32 & $1\!\times\!10^{-5}$ & AdamW (LoRA) & 16{,}000 steps & 12  & none (fixed steps) \\
  & Few-shot  & 32 & $1\!\times\!10^{-5}$ & AdamW (LoRA) & 800 steps     & 240 & none (fixed steps) \\
\midrule
\multirow{2}{*}{CALF (GPT-2)}
  & Full-shot & 8 & $1\!\times\!10^{-4}$ & Adam & 30 & 12  & ES, patience\,=\,10 \\
  & Few-shot  & 8 & $1\!\times\!10^{-4}$ & Adam & 30 & 240 & ES, patience\,=\,10 \\
\midrule
\multirow{2}{*}{Time-LLM (GPT-2)}
  & Full-shot & 16 & $5\!\times\!10^{-4}$ & Adam (OneCycleLR) & 30 & 12  & ES, patience\,=\,10 \\
  & Few-shot  & 16 & $5\!\times\!10^{-4}$ & Adam (OneCycleLR) & 30 & 240 & ES, patience\,=\,10 \\
\bottomrule
\end{tabular}}
\end{table*}

\subsection{Evaluation Metrics}
\label{sec:evaluation_metrics_eq}
\subsubsection{Mean Absolute Error}
Let \( \hat{y}_i^{(h)} \) and \( y_i^{(h)} \) denote the predicted and ground-truth CGM values for sample \( i \) at horizon \( h \in \{15,30,60,90\} \) minutes. RMSE at horizon \(h\) is defined as:
\begin{equation}
\text{RMSE}^{(h)} = \sqrt{ \frac{1}{N_h} \sum_{i=1}^{N_h} \left( \hat{y}_i^{(h)} - y_i^{(h)} \right)^2 },
\end{equation}

\begin{equation}
\text{MAE}^{(h)} = \frac{1}{N_h} \sum_{i=1}^{N} \left| \hat{y}_i^{(h)} - y_i^{(h)} \right|
\end{equation}
where \(N_h\) is the number of evaluated prediction points at horizon \(h\).

\subsubsection{Clarke Error Grid and Surveillance Error Grid}

The Clarke Error Grid (CEG) ~\cite{Clarke2005} (Appendix Figure~\ref{fig:ceg_seg} left) partitions the (reference, prediction) plane into five discrete zones with hard boundaries — Zone A (clinically accurate, ±20\% of reference), B (benign error), C (over-correction), D (failure to detect hypo/hyperglycemia), and E (erroneous treatment, opposite-direction errors). The Surveillance Error Grid (SEG)~\cite{Klonoff2014} (Appendix Figure~\ref{fig:ceg_seg} right) replaces the discrete zones with a continuous risk surface derived from the consensus of 206 diabetes clinicians. Each (reference, predicted) pair is assigned a real-valued risk score in [0, 4], visualized here as a green→yellow→orange→red gradient (None → Mild → Moderate → High → Extreme). Unlike CEG, the SEG surface is asymmetric — for the same magnitude of error, predictions on the hypoglycemia side (upper-left of the diagonal) carry higher risk than equally large errors on the hyperglycemia side, reflecting the clinical reality that under-detected lows can drive immediate insulin overdose.

\textbf{\begin{figure}[ht!]
  \centering
  \includegraphics[width=\linewidth]{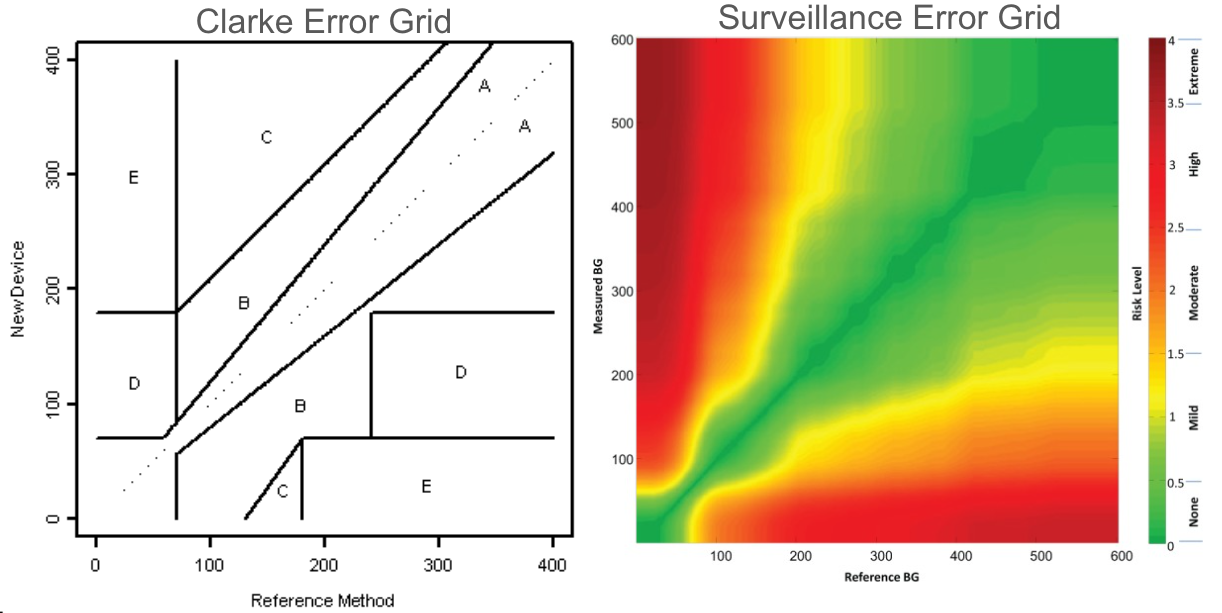}
  \caption{Comparison of two clinical error grids for blood-glucose prediction.}
  \label{fig:ceg_seg}
\end{figure}}


\subsection{TimeLLM Prompts}
\label{timellm_prompts}

We include two population-level prompts applied uniformly to all participants, namely a basic version and an enhanced version, as well as several personalized prompt examples used in a case study of personalized glucose forecasting. These personalized prompts incorporate participant demographic information and summaries of glycemic management patterns.

\subsubsection{Basic public prompt}

Continuous glucose monitoring (CGM) data provide high-frequency measurements of blood glucose levels and are widely used in diabetes management and research. This dataset consists of multivariate time-series glucose data collected from multiple individuals, with measurements sampled every 5 minutes. Each data point represents an interstitial glucose value recorded by a wearable sensor. The time series lengths vary across individuals due to different monitoring durations. The goal of this task is long-term time-series forecasting, where future glucose values are predicted based on historical observations. The dataset is split chronologically into training, validation, and test sets for each individual to prevent information leakage across time.

\subsubsection{Enhanced public prompt}
Blood glucose data are recorded every 5 minutes via continuous glucose monitors (CGM) worn by individuals with varying metabolic conditions, including type 1 diabetes, type 2 diabetes, prediabetes, and no diabetes. Most of the subjects are type 1 diabetes. Each data point represents an interstitial glucose measurement in mg/dL, ranging from 40 to 400 mg/dL. The clinical target range is 70–180 mg/dL; values below 70 mg/dL indicate hypoglycemia and values above 180 mg/dL indicate hyperglycemia, both of which carry significant health risks. Blood glucose exhibits short-term fluctuations over minutes to hours driven by physiological processes, alongside daily periodic patterns. The dataset spans multiple individuals across 15 studies from five countries. The input context window is 12 hours (144 time steps) and the forecasting target is 15, 30, 60, 90 minutes ahead. In blood glucose forecasting, the more recent CGM value is more valuable than values in earlier time.

\subsubsection{Personalized prompt examples}
\paragraph{HUPA0006P} Blood glucose data are recorded every 5 minutes via a continuous glucose monitor (CGM) worn by this individual. Each data point represents an interstitial glucose measurement in mg/dL. The clinical target range is 70–180 mg/dL; values below 70 mg/dL indicate hypoglycemia and values above 180 mg/dL indicate hyperglycemia, both of which carry significant health risks. The input context window is 12 hours (144 time steps) and the forecasting target is 15, 30, 60, 90 minutes ahead. In blood glucose forecasting, the more recent CGM value is more valuable than values in earlier time.

This subject is a 22-year-old male with type 1 diabetes (T1D) diagnosed 14 years ago (duration since age 8), weight 71 kg, height 170 cm (BMI 24.6). His most recent HbA1c is 7.8\%, reflecting suboptimal long-term glycemic control.

His CGM profile shows high glucose variability: mean glucose 156 mg/dL, standard deviation 77 mg/dL, and coefficient of variation (CV) 49.5\% — well above the clinical threshold of 36\% that defines high variability. Glucose ranges from 42 to 382 mg/dL. Time-in-range (TIR, 70–180 mg/dL) is 58.8\%, time-below-range (TBR, <70 mg/dL) is 9.0\%, and time-above-range (TAR, >180 mg/dL) is 32.2\%. He has experienced 11 distinct hypoglycemic episodes, indicating a significant and recurrent risk of hypoglycemia that must be anticipated in short-horizon forecasts. Glucose tends to be lowest around 05:00 and peaks around 21:00, suggesting evening meal-driven postprandial excursions and a risk of nocturnal hypoglycemia in the early morning hours. Given his high variability and frequent hypoglycemia, accurate prediction of downward glucose trends is especially critical for this subject.

\paragraph{HUPA0009P} Blood glucose data are recorded every 5 minutes via a continuous glucose monitor (CGM) worn by this individual. Each data point represents an interstitial glucose measurement in mg/dL. The clinical target range is 70–180 mg/dL; values below 70 mg/dL indicate hypoglycemia and values above 180 mg/dL indicate hyperglycemia, both of which carry significant health risks. The input context window is 12 hours (144 time steps) and the forecasting target is 15, 30, 60, 90 minutes ahead. In blood glucose forecasting, the more recent CGM value is more valuable than values in earlier time.

This subject is a 41-year-old female with type 1 diabetes (T1D) diagnosed 30 years ago (duration since age 11), weight 64 kg, height 165 cm (BMI 23.5). Her most recent HbA1c is 7.6\%, reflecting moderately suboptimal long-term glycemic control.

Her CGM profile shows persistently elevated glucose with relatively low variability: mean glucose 199 mg/dL, standard deviation 42 mg/dL, and coefficient of variation (CV) 21.3\% — below the 36\% high-variability threshold, indicating a stable but chronically elevated trajectory. Glucose ranges from 107 to 307 mg/dL. Time-in-range (TIR, 70–180 mg/dL) is only 36.7\%, time-above-range (TAR, >180 mg/dL) is 63.3\%, and time-below-range (TBR) is 0.0\% — she has experienced no hypoglycemic episodes. Her dominant challenge is sustained hyperglycemia rather than hypoglycemia. Glucose tends to peak around 07:00 (likely fasting or dawn phenomenon) and reach its daily minimum around 11:00, an unusual pattern that may reflect a pronounced dawn effect or delayed breakfast insulin response. For this subject, forecasting accuracy in the hyperglycemic range (>180 mg/dL) is particularly important, while hypoglycemia risk is negligible.

\paragraph{HUPA0010P} Blood glucose data are recorded every 5 minutes via a continuous glucose monitor (CGM) worn by this individual. Each data point represents an interstitial glucose measurement in mg/dL. The clinical target range is 70–180 mg/dL; values below 70 mg/dL indicate hypoglycemia and values above 180 mg/dL indicate hyperglycemia, both of which carry significant health risks. The input context window is 12 hours (144 time steps) and the forecasting target is 15, 30, 60, 90 minutes ahead. In blood glucose forecasting, the more recent CGM value is more valuable than values in earlier time.

This subject is a 42-year-old female with type 1 diabetes (T1D) diagnosed 15 years ago (duration since age 27), weight 51 kg, height 164 cm (BMI 19.0). Her most recent HbA1c is 6.0\%, reflecting excellent long-term glycemic control — the lowest HbA1c among the three subjects.

Her CGM profile reflects generally well-controlled but highly variable glucose: mean glucose 135 mg/dL, standard deviation 58 mg/dL, and coefficient of variation (CV) 43.0\%, indicating high intra-day variability despite the low HbA1c. Glucose ranges from 41 to 416 mg/dL — the widest range of the three subjects, suggesting large excursions in both directions. Time-in-range (TIR, 70–180 mg/dL) is 73.2\%, time-below-range (TBR, <70 mg/dL) is 9.5\%, and time-above-range (TAR, >180 mg/dL) is 17.2\%. She has experienced 13 hypoglycemic episodes — the most frequent of the three subjects — making hypoglycemia detection and early warning the most clinically critical forecasting objective for her. Her low body weight (BMI 19.0) may contribute to insulin sensitivity and increased hypoglycemia susceptibility. Glucose tends to be lowest around 05:00 and peaks around 14:00, suggesting postprandial hyperglycemia after lunch and a risk of early-morning hypoglycemia. Given her tight mean control but frequent severe excursions, the model should pay particular attention to rapid downward glucose trends in the context window.



\clearpage
\section{More Results}

\subsection{Context Length Exploration}
\begin{figure*}[h]
  \centering
  \includegraphics[width=\linewidth]{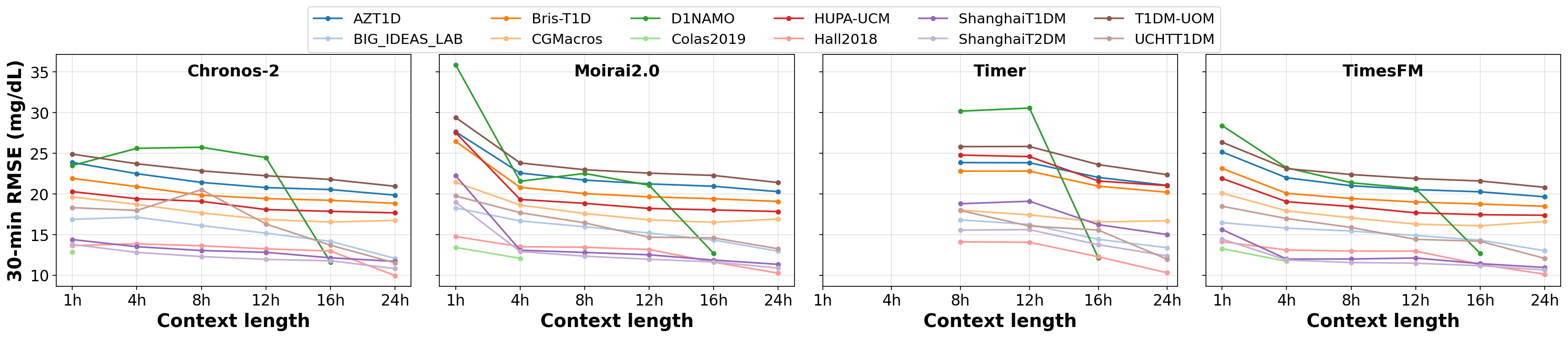}
  \caption{Zero-shot evaluation of pre-trained TSFMs on different context lengths. }
  \label{fig:sampling_horizon}
\end{figure*}

\subsection{Persistence Forecasting Baseline Results}
\begin{table}[!ht]
  \caption{Persistence forecasting baseline results (30-min prediction horizon) on RMSE, MAE, CEG zone-A ratio, SEG no-risk ratio, and RMSE skill scores relative to the persistence baseline (\%)}
  \label{tab:persistence_30_details}
  \centering
  \resizebox{\textwidth}{!}{
  \begin{tabular}{lcccccc}
    \toprule
    Dataset & RMSE (mg/dL) & MAE (mg/dL) & CEG zone A ratio (\%) & SEG no-risk ratio (\%) 
    & \multicolumn{2}{c}{Skill Score (\%)} \\
    \cmidrule(lr){6-7}
    & & & & & TimesFM Zeroshot & LSTM Fullshot \\
    \midrule
    \textbf{OpenAccess} & & & & & & \\
    \quad Hall2018       & 13.65 (4.52)   & 9.22 (2.90)    & 89.06 (6.21)    & 90.51 (7.61)  & 5.05  & 5.20 \\
    \quad D1NAMO         & 28.18 (11.92)  & 20.44 (7.50)   & 76.27 (14.68)   & 77.78 (14.26) & 26.72 & 33.82 \\
    \quad Colas2019      & 13.26 (7.14)   & 10.16 (5.44)   & 88.34 (12.11)   & 92.34 (11.45) & --    & -- \\
    \quad BIG IDEAS Lab  & 16.91 (4.03)   & 10.98 (2.55)   & 86.65 (5.32)    & 91.73 (6.04)  & 12.00 & 13.36 \\
    \quad ShanghaiT1DM   & 17.18 (6.57)   & 12.72 (4.63)   & 90.03 (7.13)    & 88.33 (6.59)  & 29.51 & 35.97 \\
    \quad ShanghaiT2DM   & 15.03 (4.14)   & 10.51 (3.07)   & 92.07 (5.18)    & 92.02 (5.18)  & 23.69 & 29.67 \\
    \quad UCHTT1DM       & 18.35 (8.40)   & 12.89 (6.47)   & 83.02 (11.10)   & 81.11 (12.69) & 21.47 & 22.29 \\
    \quad HUPA-UCM       & 22.27 (6.15)   & 15.91 (4.83)   & 82.85 (7.37)    & 83.48 (7.02)  & 20.66 & 24.97 \\
    \quad CGMacros       & 19.44 (4.88)   & 13.08 (3.46)   & 88.28 (4.38)    & 90.02 (4.95)  & 16.26 & 15.79 \\
    \quad T1DM-UOM       & 24.95 (5.17)   & 17.86 (3.28)   & 80.10 (5.21)    & 82.29 (4.54)  & 12.26 & 16.59 \\
    \quad Bris-T1D       & 22.24 (6.47)   & 14.09 (6.11)   & 85.23 (7.68)    & 87.00 (6.61)  & 14.57 & 19.83 \\
    \quad AZT1D          & 23.36 (3.39)   & 17.14 (2.92)   & 81.65 (5.75)    & 83.48 (4.47)  & 12.11 & 16.82 \\
    \midrule
    \textbf{ControlledAccess} & & & & & & \\
    \quad OhioT1DM       & 23.42 (2.97)   & 17.04 (2.17)   & 83.39 (5.94)    & 84.60 (5.13)  & 16.18 & 19.81 \\
    \quad DiaTrend       & 30.46 (5.41)   & 21.86 (4.03)   & 81.11 (5.96)    & 83.67 (4.94)  & 18.09 & 20.55 \\
    \quad T1DEXI         & 24.31 (11.92)  & 17.64 (4.21)   & 80.25 (7.48)    & 82.34 (6.69)  & 14.85 & 18.96 \\
    \midrule
    \textbf{Overall Open} & 18.31 (7.19)   & 13.98 (5.04)   & 87.76 (9.64)    & 89.85 (9.57)  & 18.02 & 21.52 \\
    \textbf{Overall}      & 22.25 (6.87)   & 16.93 (5.19)   & 84.04 (9.30)    & 86.15 (8.96)  & 17.58 & 19.24 \\
    \bottomrule
  \end{tabular}
  }
\end{table}

\begin{table}[ht]
  \caption{Persistence baseline across prediction horizons. Each cell is Mean (STD) across all participants (open-access and controlled-access combined). RMSE and MAE are in mg/dL; CEG/SEG are the no-risk zone ratios in \%.}
  \label{tab:persistence_overall_horizons}
  \centering
  \begin{tabular}{lcccc}
    \toprule
    Horizon & RMSE (mg/dL) & MAE (mg/dL) & CEG Zone A ratio (\%) & SEG no-risk ratio (\%) \\
    \midrule
    15 min & 12.72 (4.90)  & 9.02 (3.62)   & 93.68 (6.62)  & 93.75 (6.18)  \\
    30 min & 22.25 (6.87)  & 16.93 (5.19)  & 84.04 (9.30)  & 86.15 (8.96)  \\
    60 min & 31.76 (13.22) & 23.47 (9.88)  & 69.25 (14.04) & 75.08 (13.63) \\
    90 min & 39.21 (17.38) & 29.51 (13.14) & 60.55 (16.59) & 68.05 (16.65) \\
    \bottomrule
  \end{tabular}
\end{table}

\clearpage
\subsection{All Metrics on Zero-, Few-, and Full-shot with Different Prediction Horizons}

\begin{table*}[!ht]
\caption{Zero-shot performance of pre-trained TSFMs across prediction horizons using a 12-hour context window. Values are reported as mean (STD) across all evaluated datasets for RMSE, MAE, CEG Zone A ratio, and SEG no-risk ratio.}
\label{tab:zeroshot_combined_horizon}
\centering
\resizebox{0.8\textwidth}{!}{
\begin{tabular}{lcccc}
\toprule
Prediction Horizon & Chronos-2 & Moirai2.0 & Timer & TimesFM \\
\midrule

\multicolumn{5}{l}{RMSE (mg/dL)} \\
15 min &  10.22 (3.56)  &  10.45 (3.57)  &  12.84 (4.14)  &   9.70 (3.41) \\
30 min &  19.16 (6.34)  &  19.48 (6.38)  &  22.55 (7.45)  &  18.75 (5.91) \\
60 min &  32.04 (11.13) &  33.83 (10.88) &  35.17 (12.64) &  31.79 (10.35) \\
90 min &  40.16 (14.95) &  42.86 (14.70) &  42.28 (16.31) &  39.95 (13.92) \\

\midrule
\multicolumn{5}{l}{MAE (mg/dL)} \\
15 min &   7.11 (2.59)  &   7.32 (2.57)  &   9.16 (3.11)  &   6.73 (2.46) \\
30 min &  13.51 (4.70)  &  14.13 (4.59)  &  16.61 (5.79)  &  13.24 (4.42) \\
60 min &  23.25 (8.47)  &  24.84 (8.35)  &  26.63 (10.09) &  23.20 (8.05) \\
90 min &  29.83 (11.63) &  32.14 (11.44) &  32.59 (13.10) &  29.79 (10.91) \\

\midrule
\multicolumn{5}{l}{CEG Zone A ratio (\%)} \\
15 min &  96.77 (3.44)  &  96.74 (5.02)  &  95.05 (5.37)  &  97.05 (4.91) \\
30 min &  87.11 (7.60)  &  86.84 (8.45)  &  82.18 (9.95)  &  87.63 (8.22) \\
60 min &  70.80 (12.19) &  69.11 (12.91) &  64.63 (14.27) &  70.81 (13.15) \\
90 min &  61.05 (14.17) &  59.05 (15.43) &  56.16 (16.42) &  60.87 (15.31) \\

\midrule
\multicolumn{5}{l}{SEG no-risk ratio (\%)} \\
15 min &  95.92 (3.50)  &  95.95 (5.45)  &  94.26 (5.87)  &  96.36 (4.70) \\
30 min &  87.86 (7.55)  &  87.88 (7.80)  &  84.73 (9.08)  &  88.59 (7.60) \\
60 min &  75.44 (11.17) &  74.21 (12.25) &  71.53 (13.17) &  75.59 (12.56) \\
90 min &  67.55 (13.49) &  65.57 (14.92) &  64.21 (15.63) &  67.50 (14.70) \\

\bottomrule
\end{tabular}
}
\end{table*}

\begin{table*}[!ht]
\caption{Few-shot performance across prediction horizons using a 12-hour context window. Values are reported as mean (STD) across all evaluated datasets for RMSE, MAE, CEG Zone A ratio, and SEG no-risk ratio.}
\label{tab:fewshot_horizon_stack}
\centering
\resizebox{\textwidth}{!}{
\begin{tabular}{lcc|cccc|cc}
\toprule
& \multicolumn{2}{c|}{Supervised DL Models}
& \multicolumn{4}{c|}{Pre-trained TSFMs}
& \multicolumn{2}{c}{TS LLM-based Models} \\
\cmidrule(lr){2-3}
\cmidrule(lr){4-7}
\cmidrule(lr){8-9}
Prediction Horizon 
& LSTM & GPFormer
& Chronos-2 & Moirai2.0 & Timer & TimesFM
& TimeLLM & CALF \\
\midrule

\multicolumn{9}{l}{RMSE} \\
15 min & 9.87 (3.42) & 11.46 (3.91) & 10.28 (3.42) & 10.37 (4.05) & 10.82 (4.03) & 9.93 (3.87) & 19.03 (6.10) & 12.68 (4.19) \\
30 min & 19.24 (6.02) & 20.94 (6.23) & 18.95 (6.16) & 20.23 (6.63) & 20.71 (6.66) & 19.27 (6.28) & 26.34 (8.77) & 20.91 (6.60) \\
60 min & 30.28 (9.70) & 35.49 (12.33) & 31.39 (10.85) & 33.68 (11.70) & 34.74 (11.84) & 32.20 (11.07) & 35.89 (12.94) & 32.58 (10.88) \\
90 min & 37.35 (12.74) & 38.71 (12.70) & 39.33 (14.64) & 42.52 (15.51) & 42.44 (14.93) & 40.13 (14.50) & 40.91 (15.13) & 39.51 (14.14) \\

\midrule
\multicolumn{9}{l}{MAE} \\
15 min & 6.92 (2.41) & 8.41 (2.79) & 7.24 (2.52) & 7.43 (2.91) & 7.79 (2.91) & 7.17 (2.80) & 14.05 (4.73) & 9.00 (2.94) \\
30 min & 13.86 (4.37) & 15.22 (4.64) & 13.44 (4.60) & 14.49 (4.88) & 14.93 (4.95) & 13.90 (4.68) & 19.55 (6.83) & 14.95 (4.85) \\
60 min & 22.53 (7.43) & 27.04 (9.43) & 22.86 (8.30) & 24.89 (9.05) & 25.74 (9.15) & 24.11 (8.65) & 27.10 (10.44) & 23.95 (8.36) \\
90 min & 28.51 (9.96) & 29.58 (9.92) & 29.31 (11.42) & 32.01 (12.03) & 32.04 (11.60) & 30.59 (11.41) & 31.33 (12.28) & 29.67 (11.12) \\

\midrule
\multicolumn{9}{l}{CEG Zone A ratio (\%)} \\
15 min & 97.21 (2.58)  & 95.98 (3.52)  & 96.98 (3.25)  & 96.60 (4.82)  & 96.21 (4.09)  & 96.90 (4.66)  & 86.15 (7.77)  & 94.89 (3.84) \\
30 min & 86.45 (7.60)  & 84.31 (8.48)  & 87.62 (7.27)  & 85.72 (8.93)  & 85.02 (8.86)  & 86.84 (8.51)  & 76.69 (10.50) & 84.88 (8.00) \\
60 min & 70.78 (11.70) & 62.08 (12.44) & 71.51 (11.98) & 68.01 (13.31) & 66.60 (13.31) & 68.88 (13.09) & 63.29 (13.92) & 69.44 (11.85) \\
90 min & 61.09 (13.31) & 52.63 (15.11) & 61.67 (14.10) & 57.76 (14.87) & 57.29 (14.17) & 58.83 (14.88) & 57.79 (14.65) & 60.70 (13.70) \\

\midrule
\multicolumn{9}{l}{SEG no-risk ratio (\%)} \\
15 min & 96.13 (3.77)  & 94.92 (3.94)  & 96.08 (3.28)  & 95.77 (5.13)  & 95.35 (5.33)  & 96.09 (4.95)  & 87.62 (7.01)  & 93.88 (4.48) \\
30 min & 87.68 (7.96)  & 86.33 (9.41)  & 88.30 (7.33)  & 87.10 (8.34)  & 86.46 (8.33)  & 88.07 (7.85)  & 80.64 (9.35)  & 86.14 (7.60) \\
60 min & 77.32 (10.86) & 70.93 (12.34) & 76.07 (10.99) & 73.39 (12.29) & 72.02 (12.61) & 74.24 (12.38) & 70.23 (12.82) & 74.59 (11.04) \\
90 min & 69.58 (12.99) & 63.45 (14.88) & 68.17 (13.48) & 64.79 (14.88) & 64.51 (14.14) & 66.18 (14.74) & 65.47 (14.31) & 67.75 (13.10) \\

\bottomrule
\end{tabular}
}
\end{table*}

\begin{table*}[!ht]
\caption{Full-shot performance across prediction horizons using a 12-hour context window. Values are reported as mean (STD) across all evaluated datasets for RMSE, MAE, CEG Zone A ratio, and SEG no-risk ratio.}
\label{tab:fullshot_horizon_stack}
\centering
\resizebox{\textwidth}{!}{
\begin{tabular}{lcc|cccc|cc}
\toprule
& \multicolumn{2}{c|}{Supervised DL Models}
& \multicolumn{4}{c|}{Pre-trained TSFMs}
& \multicolumn{2}{c}{TS LLM-based Models} \\
\cmidrule(lr){2-3}
\cmidrule(lr){4-7}
\cmidrule(lr){8-9}
Prediction Horizon 
& LSTM & GPFormer
& Chronos-2 & Moirai2.0 & Timer & TimesFM
& TimeLLM & CALF \\
\midrule

\multicolumn{9}{l}{RMSE} \\
15 min & 9.61 (3.44) & 10.23 (3.76) & 10.19 (3.42) & 10.18 (4.14) & 10.18 (3.97) & 9.98 (4.00) & 14.78 (4.68) & 11.72 (4.14) \\
30 min & 17.97 (5.82) & 18.97 (5.80) & 18.79 (6.04) & 20.30 (6.76) & 19.44 (6.35) & 19.47 (6.32) & 22.75 (7.28) & 20.28 (6.55) \\
60 min & 30.24 (9.51) & 31.04 (10.02) & 30.96 (10.55) & 33.78 (11.73) & 32.71 (11.21) & 32.28 (11.18) & 33.48 (11.56) & 32.10 (10.83) \\
90 min & 37.43 (12.51) & 39.46 (13.27) & 38.54 (14.16) & 42.23 (15.42) & 40.93 (14.08) & 40.19 (14.49) & 39.77 (14.51) & 39.21 (14.22) \\

\midrule
\multicolumn{9}{l}{MAE} \\
15 min & 6.69 (2.42) & 7.33 (2.65) & 7.14 (2.47) & 7.18 (2.93) & 7.30 (2.84) & 7.18 (2.89) & 10.79 (3.51) & 8.12 (2.81) \\
30 min & 12.81 (4.25) & 13.70 (4.44) & 13.34 (4.49) & 14.43 (5.02) & 13.89 (4.68) & 14.03 (4.76) & 16.63 (5.62) & 14.32 (4.73) \\
60 min & 22.54 (7.31) & 23.45 (7.83) & 22.71 (8.11) & 24.84 (8.97) & 24.16 (8.51) & 24.22 (8.69) & 25.13 (9.24) & 23.41 (8.22) \\
90 min & 28.63 (9.77) & 30.95 (10.39) & 28.92 (11.09) & 31.74 (12.03) & 30.76 (10.92) & 30.67 (11.43) & 30.50 (11.78) & 29.26 (11.04) \\

\midrule
\multicolumn{9}{l}{CEG Zone A ratio (\%)} \\
15 min & 97.28 (2.55)  & 96.97 (2.86)  & 97.12 (2.84)  & 96.69 (3.75)  & 96.72 (4.10)  & 96.86 (3.86)  & 91.53 (5.54)  & 94.89 (3.84) \\
30 min & 87.21 (7.32)  & 86.24 (7.75)  & 88.15 (7.05)  & 85.50 (8.99)  & 86.54 (8.69)  & 86.54 (8.56)  & 81.77 (8.94)  & 84.88 (8.00) \\
60 min & 70.56 (11.74) & 70.67 (11.49) & 71.81 (11.93) & 68.14 (13.04) & 68.94 (12.61) & 68.64 (12.92) & 67.01 (12.75) & 69.44 (11.85) \\
90 min & 60.56 (13.17) & 60.55 (12.91) & 62.00 (14.34) & 58.57 (15.13) & 59.15 (14.19) & 58.77 (14.85) & 58.77 (14.41) & 60.70 (13.70) \\

\midrule
\multicolumn{9}{l}{SEG no-risk ratio (\%)} \\
15 min & 96.26 (3.43)  & 95.93 (3.68)  & 96.27 (3.08)  & 95.86 (5.46)  & 95.87 (5.18)  & 95.98 (5.36)  & 91.64 (5.61)  & 93.88 (4.48) \\
30 min & 88.32 (7.31)  & 87.38 (8.37)  & 88.62 (7.06)  & 87.04 (8.39)  & 87.74 (7.72)  & 87.65 (8.40)  & 84.35 (8.22)  & 86.14 (7.60) \\
60 min & 77.21 (10.99) & 76.95 (10.56) & 76.43 (11.03) & 73.37 (12.19) & 74.17 (12.08) & 73.80 (12.54) & 73.34 (11.57) & 74.59 (11.04) \\
90 min & 69.47 (12.98) & 68.38 (13.02) & 68.73 (13.74) & 65.35 (14.74) & 66.23 (14.06) & 65.80 (14.75) & 66.60 (13.76) & 67.75 (13.10) \\
\bottomrule
\end{tabular}
}
\end{table*}

\begin{figure}[htbp]
\centering
\includegraphics[width=0.9\linewidth]{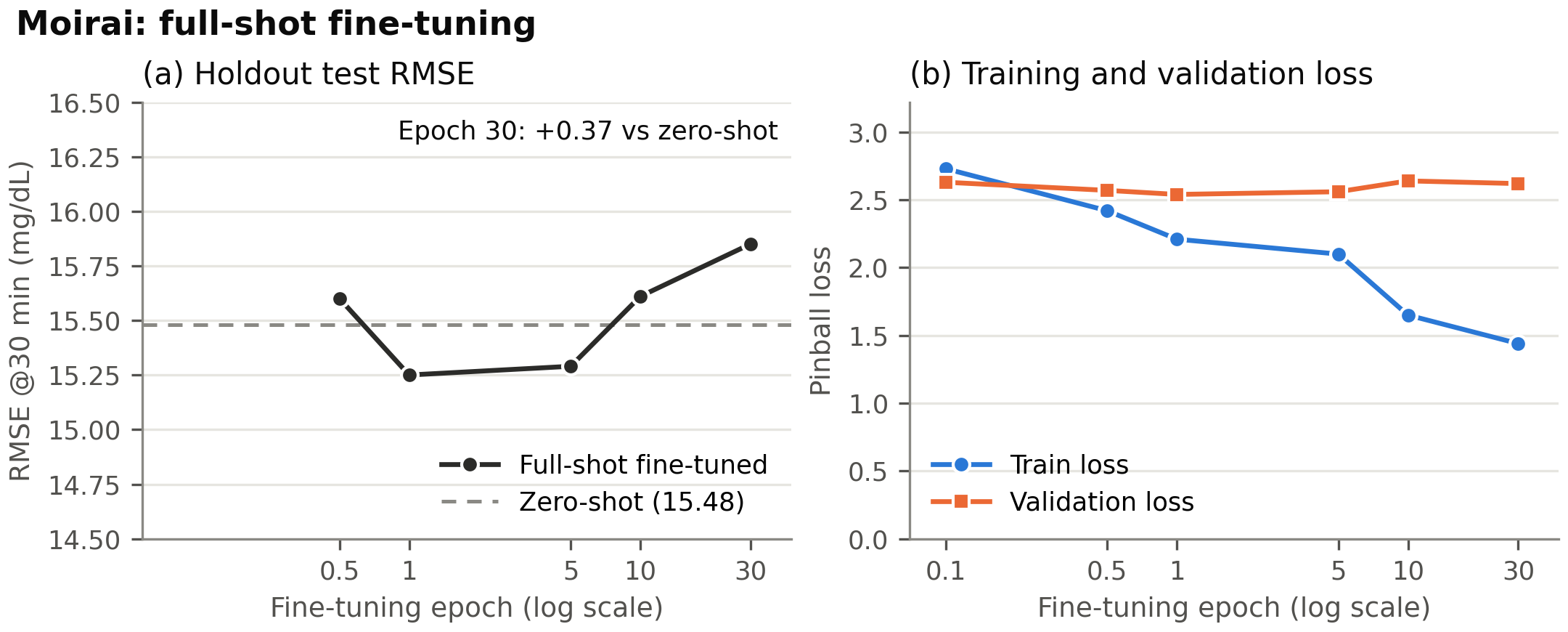}
\par\vspace{0.8em}
\includegraphics[width=0.9\linewidth]{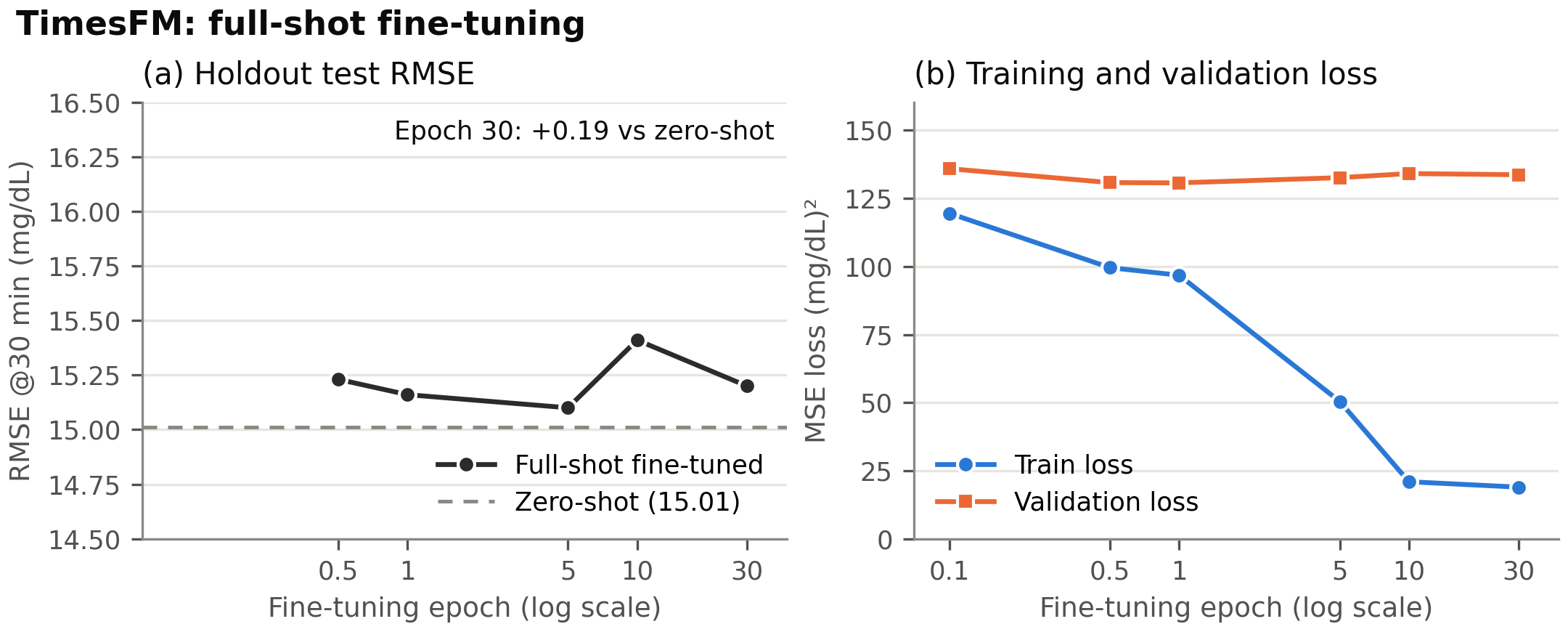}
\caption{Full-shot fine-tuning dynamics of Moirai-2 (top) and TimesFM 2.5 (bottom).
In each row, (a) shows the holdout test RMSE at the 30-minute horizon as a
function of fine-tuning epoch, with the dashed line marking the zero-shot RMSE
of the pre-trained model, and (b) shows the training and validation loss
(pinball loss for Moirai 2.0, MSE for TimesFM 2.5; each model uses its own
fine-tuning objective). The epoch axis is on a log scale. For both models,
training loss decreases steadily, while validation loss improves only slightly
within the first epoch and then plateaus or fluctuates. This is consistent with
early, mild overfitting. After 30 epochs, test RMSE is 0.37 and 0.31~mg/dL
worse than zero-shot for Moirai 2.0 and TimesFM 2.5, respectively.}
\label{fig:fullshot_curves}
\end{figure}

\clearpage
\subsection{Per-dataset SEG No-risk Ratio}
\label{sec:full_seg_30min}

\begin{table}[ht]
  \caption{Per-dataset SEG no-risk ratio (\%) at the 30-minute horizon under the few-shot setting.}
  \label{tab:seg_fewshot_per_dataset}
  \centering
  \begin{tabular}{lcccccccc}
    \toprule
    Dataset & LSTM & GPFormer & Chronos & Moirai & Timer & TimesFM & Time-LLM & CALF \\
    \midrule
    \multicolumn{9}{l}{\textit{Open Access}} \\
    Hall2018 & 89.5 & 86.2 & 90.8 & 90.8 & 90.2 & 92.1 & 86.7 & 89.4 \\
    D1NAMO & 80.6 & 76.7 & 76.1 & 92.0 & 85.2 & 89.3 & 67.9 & 74.2 \\
    Colas2019 & -- & -- & -- & -- & -- & -- & -- & -- \\
    BIG IDEAS Lab & 93.2 & 92.4 & 94.5 & 94.7 & 94.2 & 95.6 & 92.2 & 93.0 \\
    ShanghaiT1DM & 89.8 & 84.5 & 92.9 & 94.9 & 94.3 & 95.3 & 79.6 & 90.7 \\
    ShanghaiT2DM & 93.0 & 91.0 & 95.1 & 95.7 & 94.6 & 96.0 & 87.7 & 93.5 \\
    UCHTT1DM & 74.6 & 67.4 & 77.4 & 73.2 & 75.1 & 80.4 & 76.2 & 75.1 \\
    HUPA-UCM & 84.7 & 81.9 & 87.8 & 87.5 & 86.8 & 88.5 & 76.0 & 86.0 \\
    CGMacros & 92.4 & 92.3 & 92.8 & 91.8 & 90.3 & 92.0 & 90.3 & 91.8 \\
    T1DM-UOM & 85.9 & 85.0 & 86.0 & 84.2 & 84.2 & 85.4 & 77.2 & 84.2 \\
    Bris-T1D & 90.2 & 87.9 & 90.0 & 89.1 & 88.5 & 89.9 & 83.3 & 88.9 \\
    AZT1D & 87.2 & 87.4 & 87.4 & 85.1 & 85.0 & 86.5 & 79.7 & 85.9 \\
    \midrule
    \multicolumn{9}{l}{\textit{Controlled Access}} \\
    OhioT1DM & 88.6 & 86.7 & 89.3 & 87.7 & 86.7 & 88.2 & 78.3 & 86.2 \\
    DiaTrend & 87.1 & 86.5 & 87.4 & 85.3 & 84.7 & 86.3 & 78.1 & 84.9 \\
    T1DEXI & 86.5 & 85.7 & 86.7 & 84.9 & 84.3 & 85.8 & 78.3 & 84.2 \\
    \midrule
    Overall (Open) & 89.4 & 87.3 & 90.8 & 90.7 & 89.9 & 91.7 & 84.5 & 89.3 \\
    Overall (All) & 87.7 & 86.3 & 88.3 & 87.1 & 86.5 & 88.1 & 80.6 & 86.1 \\
    \bottomrule
  \end{tabular}
\end{table}

\begin{table}[ht]
  \caption{Per-dataset SEG no-risk ratio (\%) at the 30-minute horizon under the full-shot setting.}
  \label{tab:seg_fullshot_per_dataset}
  \centering
  \begin{tabular}{lcccccccc}
    \toprule
    Dataset & LSTM & GPFormer & Chronos & Moirai & Timer & TimesFM & Time-LLM & CALF \\
    \midrule
    \multicolumn{9}{l}{\textit{Open Access}} \\
    Hall2018 & 89.2 & 90.4 & 91.0 & 90.8 & 91.2 & 91.1 & 89.0 & 89.1 \\
    D1NAMO & 82.3 & 81.6 & 81.6 & 93.2 & 86.4 & 90.8 & 72.1 & 78.3 \\
    Colas2019 & -- & -- & -- & -- & -- & -- & -- & -- \\
    BIG IDEAS Lab & 93.0 & 93.5 & 94.7 & 94.0 & 95.5 & 93.7 & 93.4 & 93.4 \\
    ShanghaiT1DM & 92.3 & 86.6 & 93.8 & 94.9 & 95.9 & 95.1 & 84.0 & 92.0 \\
    ShanghaiT2DM & 94.8 & 92.4 & 95.9 & 95.8 & 95.4 & 95.9 & 91.4 & 94.7 \\
    UCHTT1DM & 74.8 & 76.8 & 78.8 & 72.6 & 75.9 & 72.9 & 76.2 & 74.9 \\
    HUPA-UCM & 86.5 & 82.9 & 88.4 & 88.5 & 88.3 & 88.1 & 81.0 & 87.1 \\
    CGMacros & 92.3 & 91.9 & 93.3 & 91.3 & 91.5 & 92.7 & 91.6 & 91.8 \\
    T1DM-UOM & 86.4 & 85.5 & 86.5 & 84.8 & 85.6 & 85.6 & 81.7 & 85.1 \\
    Bris-T1D & 90.9 & 89.3 & 90.3 & 89.2 & 89.7 & 89.7 & 87.0 & 89.6 \\
    AZT1D & 87.8 & 87.2 & 87.6 & 84.4 & 86.2 & 86.5 & 84.3 & 86.0 \\
    \midrule
    \multicolumn{9}{l}{\textit{Controlled Access}} \\
    OhioT1DM & 89.5 & 87.8 & 89.5 & 87.6 & 88.6 & 87.9 & 83.2 & 87.6 \\
    DiaTrend & 87.8 & 87.2 & 87.4 & 85.0 & 86.2 & 86.1 & 82.5 & 85.8 \\
    T1DEXI & 87.0 & 86.2 & 86.8 & 84.8 & 85.7 & 85.5 & 82.4 & 85.0 \\
    \midrule
    Overall (Open) & 90.3 & 89.2 & 91.5 & 90.7 & 91.0 & 91.1 & 87.6 & 89.9 \\
    Overall (All) & 88.3 & 87.4 & 88.6 & 87.0 & 87.7 & 87.7 & 84.3 & 86.9 \\
    \bottomrule
  \end{tabular}
\end{table}

\clearpage
\subsection{Pre-trained vs. Randomly Initialized TSFMs}
\label{sec:randinit-tsfm}

\begin{table}[ht]
\centering
\caption{Pre-trained (PT) vs.\ randomly initialized (RI) TSFMs with identical
architectures. Values are participant-level mean RMSE (mg/dL) on the
open-access test split (12-h context, 30-min horizon); better result in each
pair in \textbf{bold}. Pre-training drives zero-shot performance, while the gap
largely closes with task-specific fine-tuning.}
\label{tab:pretrain_vs_random}
\small
\setlength{\tabcolsep}{4pt}
\begin{tabular}{l cc cc cc cc}
\toprule
& \multicolumn{2}{c}{Chronos-2} & \multicolumn{2}{c}{Moirai-2}
& \multicolumn{2}{c}{Timer} & \multicolumn{2}{c}{TimesFM} \\
\cmidrule(lr){2-3} \cmidrule(lr){4-5} \cmidrule(lr){6-7} \cmidrule(lr){8-9}
Setting & PT & RI & PT & RI & PT & RI & PT & RI \\
\midrule
Zero-shot & \textbf{15.62} & 37.00 & \textbf{15.48} & 46.64 & \textbf{18.37} & 35.07 & \textbf{15.01} & 87.62 \\
Few-shot  & \textbf{15.63} & 34.29 & \textbf{15.99} & 16.64 & \textbf{16.60} & 17.25 & \textbf{15.20} & 16.16 \\
Full-shot & \textbf{15.26} & 17.79 & 15.85 & \textbf{15.59} & 15.47 & \textbf{15.40} & 15.32 & \textbf{14.79} \\
\bottomrule
\end{tabular}
\end{table}

\clearpage
\subsection{Result of TimeLLM Ablation Study Using Different Prompts}
\label{sec:timellm_ablation_results}
\begin{table}[ht]
\centering
\caption{TimeLLM prompt ablation and personalized prompt case study. RMSE values are reported in mg/dL; lower is better.}
\label{tab:timellm_prompt_ablation}
\small


\begin{subtable}{0.9\textwidth}
\centering
\caption{Population-level comparison of prompt variants, including an empty prompt, for 30-minute forecasting. Values are participant-level mean (STD) RMSE.}
\label{tab:timellm_public_prompt}
\begin{tabular}{llcc}
\toprule
\textbf{Dataset} & \textbf{Prompt version} & \textbf{Few-shot} & \textbf{Full-shot} \\
\midrule
OpenAccess & No prompt              & 26.53 (11.86) & 23.39 (11.50) \\
OpenAccess & Basic public prompt    & 21.25 (8.75)  & 18.50 (7.08)  \\
OpenAccess & Enhanced public prompt & 21.26 (8.73)  & 18.59 (7.12)  \\
\midrule
Overall    & Basic public prompt    & 26.34 (8.77)  & 22.75 (7.28)  \\
Overall    & Enhanced public prompt & 26.34 (8.75)  & 22.87 (7.33)  \\
\bottomrule
\end{tabular}
\end{subtable}

\vspace{0.8em}

\begin{subtable}{0.75\textwidth}
\centering
\caption{Personalized prompt case study under full-shot evaluation.}
\label{tab:timellm_personalized_prompt}
\begin{tabular}{lcc}
\toprule
\textbf{Subject} & \textbf{New public prompt} & \textbf{New personalized prompt} \\
\midrule
HUPA0006 & 32.65 & 32.57 \\
HUPA0009 & 13.34 & 13.55 \\
HUPA0010 & 19.88 & 19.90 \\
\bottomrule
\end{tabular}
\end{subtable}

\end{table}

\clearpage
\subsection{Representative CEG and SEG Figures on Full-shot evaluation}
\begin{figure}[htbp]
    \centering

    \begin{subfigure}[b]{0.49\textwidth}
        \centering
        \includegraphics[width=\textwidth]{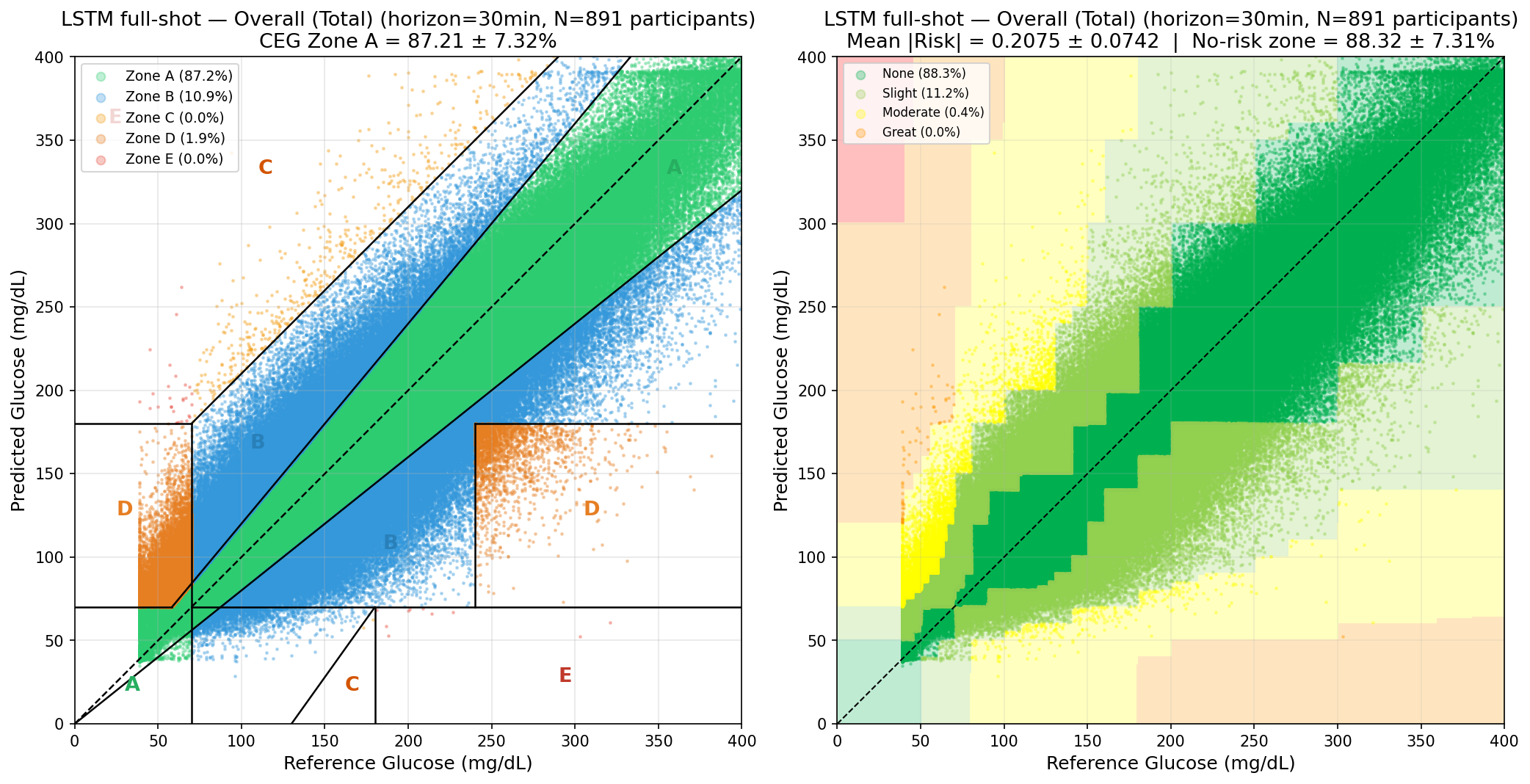}
        \caption{LSTM}
        \label{fig:ceg_seg_lstm}
    \end{subfigure}
    \hfill
    \begin{subfigure}[b]{0.49\textwidth}
        \centering
        \includegraphics[width=\textwidth]{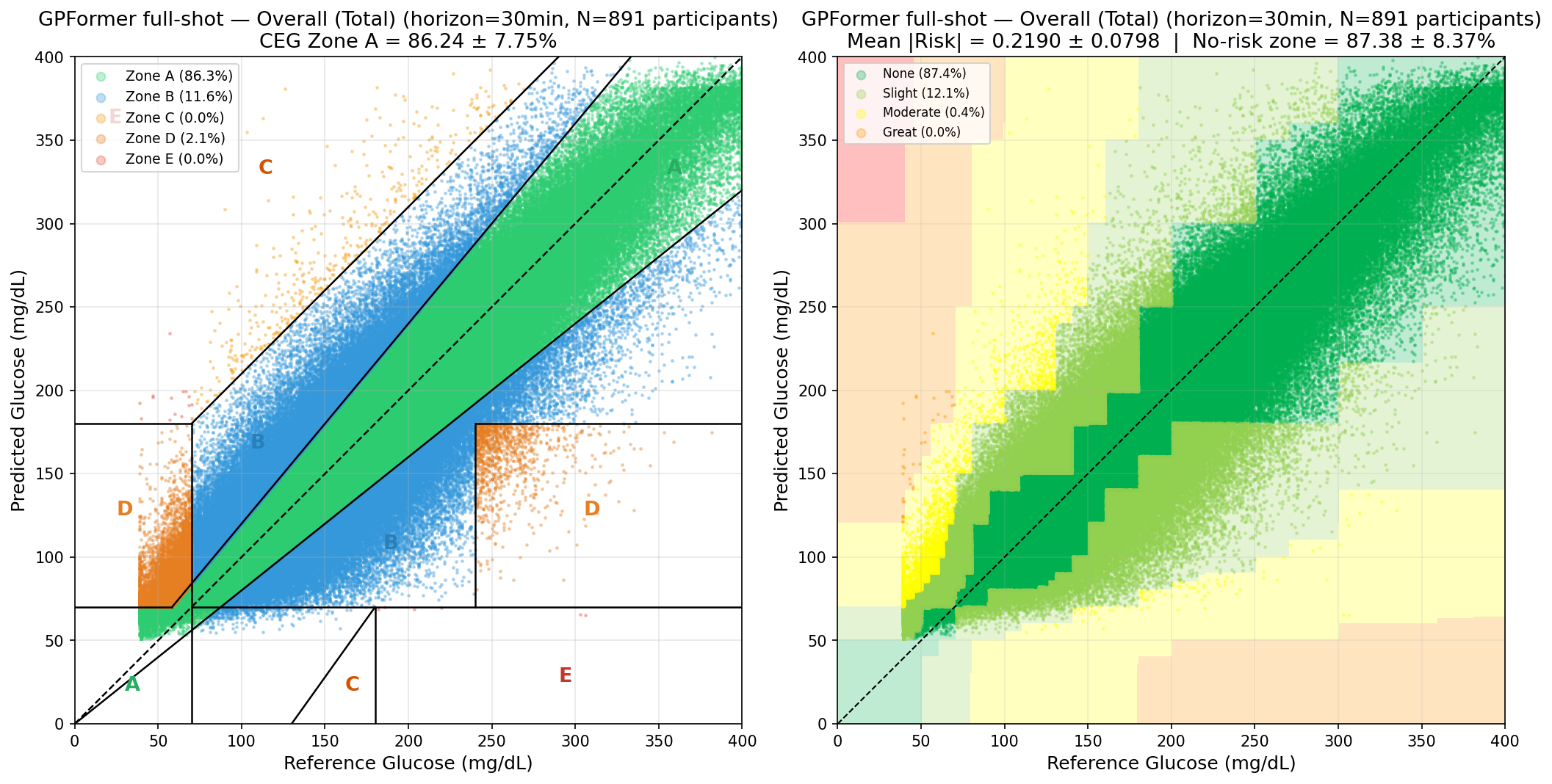}
        \caption{GPFormer}
        \label{fig:ceg_seg_gpformer}
    \end{subfigure}

    \vspace{0.8em}

    \begin{subfigure}[b]{0.49\textwidth}
        \centering
        \includegraphics[width=\textwidth]{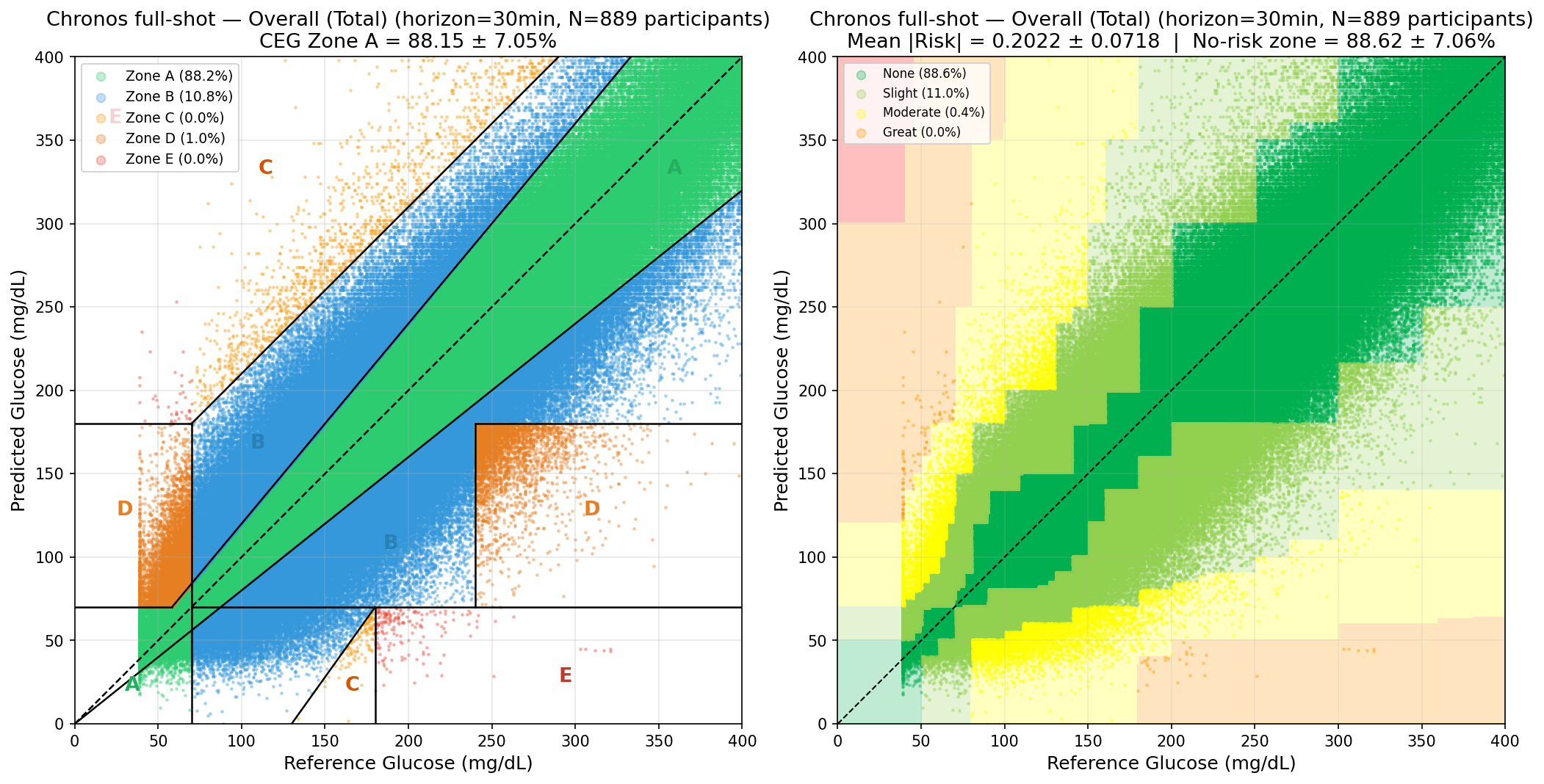}
        \caption{Chronos}
        \label{fig:ceg_seg_chronos}
    \end{subfigure}
    \hfill
    \begin{subfigure}[b]{0.49\textwidth}
        \centering
        \includegraphics[width=\textwidth]{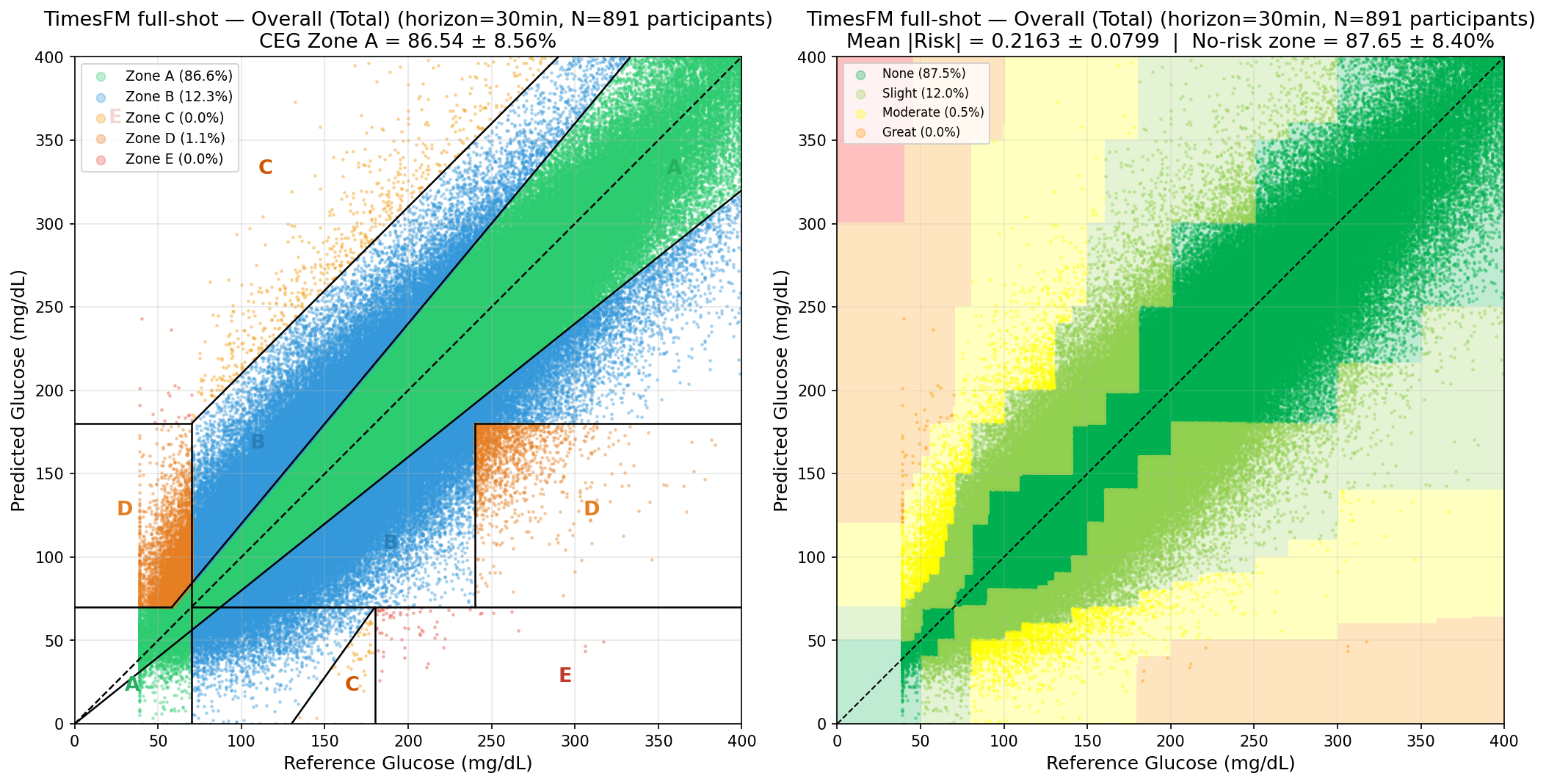}
        \caption{TimesFM}
        \label{fig:ceg_seg_timesfm}
    \end{subfigure}

    \vspace{0.8em}

    \begin{subfigure}[b]{0.49\textwidth}
        \centering
        \includegraphics[width=\textwidth]{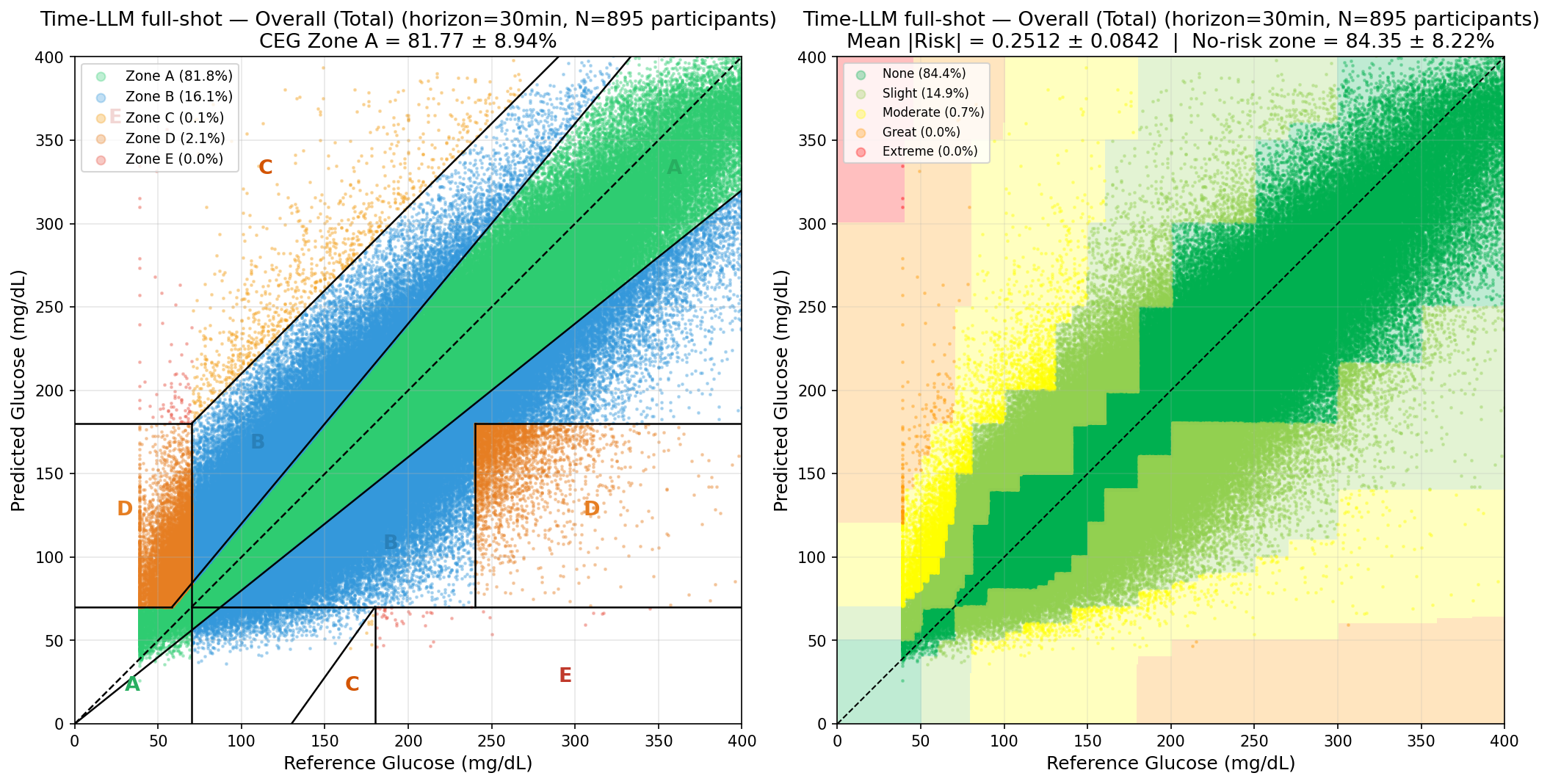}
        \caption{Time-LLM}
        \label{fig:ceg_seg_timellm}
    \end{subfigure}
    \hfill
    \begin{subfigure}[b]{0.49\textwidth}
        \centering
        \includegraphics[width=\textwidth]{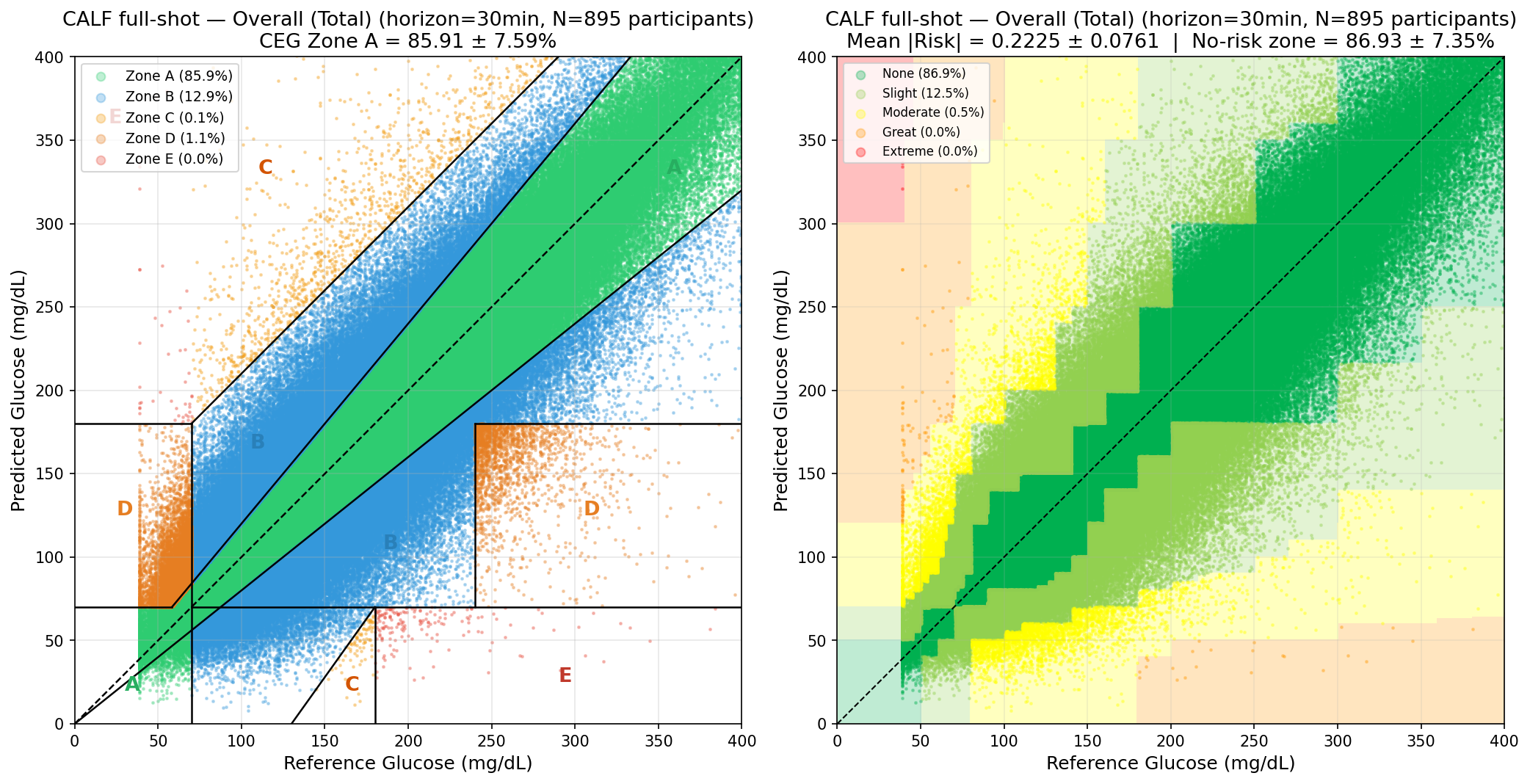}
        \caption{CALF}
        \label{fig:ceg_seg_calf}
    \end{subfigure}

    \caption{Clarke Error Grid (CEG, left) and Surveillance Error Grid (SEG, right) for 30-minute glucose predictions across all full-shot models.}
    \label{fig:ceg_seg_fullshot_30min}
\end{figure}

\clearpage
\subsection{Probabilistic Forecasting Results}
\begin{table}[ht]
  \caption{Zero-shot PI80 prediction-interval coverage of pre-trained TSFMs in Mean (STD) \% (12-h context length, 30-min prediction horizon). Bold values are closest to the nominal 80\% level; underlined values are second-closest.}
  \label{tab:zeroshot_pi80}
  \centering
  \begin{threeparttable}
  \begin{tabular}{lccc}
    \toprule
    & \multicolumn{3}{c}{PI80 Coverage (\%)} \\
    \cmidrule(lr){2-4}
    Dataset & Chronos-2 & Moirai2.0 & TimesFM \\
    \midrule
    \textbf{OpenAccess} & & & \\
    \quad Hall2018 & \textbf{75.63 (8.86)} & \underline{74.88 (16.76)} & 74.41 (19.24) \\
    \quad D1NAMO & 65.28 (15.49) & \underline{93.48 (8.13)} & \textbf{84.78 (12.23)} \\
    \quad Colas2019$^{1}$ & -- & -- & -- \\
    \quad BIG IDEAS Lab & \underline{76.70 (2.44)} & 84.69 (7.14) & \textbf{82.22 (7.43)} \\
    \quad ShanghaiT1DM & 75.53 (3.82) & \textbf{81.08 (5.57)} & \underline{81.24 (5.71)} \\
    \quad ShanghaiT2DM & 75.06 (7.77) & \underline{80.94 (9.39)} & \textbf{79.58 (10.67)} \\
    \quad UCHTT1DM & 76.49 (11.79) & \underline{76.84 (16.99)} & \textbf{78.95 (20.84)} \\
    \quad HUPA-UCM & 70.82 (4.25) & \textbf{75.75 (5.80)} & \underline{74.86 (7.55)} \\
    \quad CGMacros & 78.62 (2.86) & \underline{80.70 (6.24)} & \textbf{80.39 (5.73)} \\
    \quad T1DM-UOM & 73.50 (1.56) & \underline{77.64 (2.79)} & \textbf{78.62 (2.43)} \\
    \quad Bris-T1D & \textbf{81.67 (6.98)} & \underline{76.10 (14.78)} & 84.70 (5.72) \\
    \quad AZT1D & 74.91 (1.20) & \underline{77.48 (3.23)} & \textbf{79.84 (3.45)} \\
    \midrule
    \textbf{ControlledAccess} & & & \\
    \quad OhioT1DM & 75.10 (1.83) & \underline{79.03 (2.41)} & \textbf{80.65 (2.61)} \\
    \quad DiaTrend & 75.30 (1.38) & \underline{77.44 (1.90)} & \textbf{79.74 (1.71)} \\
    \quad T1DEXI & 74.94 (2.55) & \underline{77.87 (3.89)} & \textbf{79.13 (3.96)} \\
    \midrule
    \textbf{Overall Open}  & \underline{77.36 (7.39)} & 77.29 (10.90) & \textbf{81.14 (11.64)} \\
    \textbf{Overall}  & 75.88 (4.91) & \underline{77.54 (7.30)} & \textbf{79.90 (7.70)} \\
    \bottomrule
  \end{tabular}
  \begin{tablenotes}[flushleft]
    \footnotesize
    \item[1] Colas2019 lacks sufficient test sequence length for evaluation at a 12-h context length, so probabilistic metrics are not reported here.
    \item[*] Cells are Mean (STD) of PI80 coverage across participants (mean is patient-weighted; std is unweighted across participants).
  \end{tablenotes}
  \end{threeparttable}
\end{table}

\begin{table}[ht]
  \caption{Few-shot PI80 prediction-interval coverage of pre-trained TSFMs in Mean (STD) \% (12-h context length, 30-min prediction horizon). Bold values are closest to the nominal 80\% level; underlined values are second-closest.}
  \label{tab:fewshot_pi80}
  \centering
  \begin{threeparttable}
  \begin{tabular}{lccc}
    \toprule
    & \multicolumn{3}{c}{PI80 Coverage (\%)} \\
    \cmidrule(lr){2-4}
    Dataset & Chronos-2 & Moirai2.0 & TimesFM \\
    \midrule
    \textbf{OpenAccess} & & & \\
    \quad Hall2018 & \textbf{78.74 (8.29)} & 52.45 (19.83) & \underline{63.92 (11.63)} \\
    \quad D1NAMO & \textbf{70.83 (12.72)} & 41.30 (28.69) & \underline{54.07 (9.03)} \\
    \quad Colas2019$^{1}$ & -- & -- & -- \\
    \quad BIG IDEAS Lab & \textbf{80.48 (2.23)} & 56.30 (7.75) & \underline{71.21 (5.57)} \\
    \quad ShanghaiT1DM & \textbf{78.15 (3.63)} & 47.85 (8.16) & \underline{62.20 (7.23)} \\
    \quad ShanghaiT2DM & \textbf{78.04 (7.65)} & 48.87 (11.90) & \underline{61.13 (7.81)} \\
    \quad UCHTT1DM & \textbf{79.32 (12.58)} & 46.84 (18.05) & \underline{66.62 (15.68)} \\
    \quad HUPA-UCM & \textbf{72.64 (4.40)} & 45.95 (11.35) & \underline{56.85 (8.42)} \\
    \quad CGMacros & \textbf{81.16 (2.91)} & 54.64 (8.15) & \underline{69.36 (4.98)} \\
    \quad T1DM-UOM & \textbf{76.42 (1.68)} & 46.25 (3.18) & \underline{57.70 (2.73)} \\
    \quad Bris-T1D & \underline{61.11 (21.52)} & 61.03 (15.27) & \textbf{70.51 (11.08)} \\
    \quad AZT1D & \textbf{78.43 (1.50)} & 46.83 (4.04) & \underline{61.43 (3.06)} \\
    \midrule
    \textbf{ControlledAccess} & & & \\
    \quad OhioT1DM & \textbf{78.20 (1.81)} & 49.49 (3.73) & \underline{57.33 (2.47)} \\
    \quad DiaTrend & \textbf{78.11 (1.31)} & 46.97 (3.12) & \underline{59.46 (4.10)} \\
    \quad T1DEXI & \textbf{78.27 (2.14)} & 46.05 (5.63) & \underline{61.80 (3.95)} \\
    \midrule
    \textbf{Overall Open}  & \textbf{70.00 (9.25)} & 53.50 (13.70) & \underline{64.70 (9.48)} \\
    \textbf{Overall}  & \textbf{76.33 (5.89)} & 48.21 (9.60) & \underline{61.25 (6.68)} \\
    \bottomrule
  \end{tabular}
  \begin{tablenotes}[flushleft]
    \footnotesize
    \item[1] Colas2019 lacks sufficient test sequence length for evaluation at a 12-h context length, so probabilistic metrics are not reported here.
    \item[*] Cells are Mean (STD) of PI80 coverage across participants (mean is patient-weighted; std is unweighted across participants).
  \end{tablenotes}
  \end{threeparttable}
\end{table}

\begin{table}[ht]
  \caption{Full-shot PI80 prediction-interval coverage of pre-trained TSFMs in Mean (STD) \% (12-h context length, 30-min prediction horizon). Bold values are closest to the nominal 80\% level; underlined values are second-closest.}
  \label{tab:fullshot_pi80}
  \centering
  \begin{threeparttable}
  \begin{tabular}{lccc}
    \toprule
    & \multicolumn{3}{c}{PI80 Coverage (\%)} \\
    \cmidrule(lr){2-4}
    Dataset & Chronos-2 & Moirai2.0 & TimesFM \\
    \midrule
    \textbf{OpenAccess} & & & \\
    \quad Hall2018 & \textbf{78.22 (8.09)} & 50.55 (17.54) & \underline{53.55 (12.67)} \\
    \quad D1NAMO & \textbf{79.17 (4.83)} & \underline{73.91 (13.70)} & 18.60 (10.83) \\
    \quad Colas2019$^{1}$ & -- & -- & -- \\
    \quad BIG IDEAS Lab & \textbf{79.57 (3.64)} & 58.52 (11.95) & \underline{58.70 (8.93)} \\
    \quad ShanghaiT1DM & \textbf{81.18 (3.84)} & \underline{72.81 (8.60)} & 22.97 (9.45) \\
    \quad ShanghaiT2DM & \textbf{80.88 (8.30)} & \underline{69.33 (12.95)} & 30.06 (9.19) \\
    \quad UCHTT1DM & \textbf{78.72 (12.79)} & 50.53 (20.42) & \underline{52.63 (23.94)} \\
    \quad HUPA-UCM & \textbf{74.37 (5.77)} & \underline{59.50 (9.57)} & 26.78 (5.85) \\
    \quad CGMacros & \textbf{80.89 (3.36)} & \underline{56.20 (8.39)} & 53.16 (12.58) \\
    \quad T1DM-UOM & \textbf{77.81 (1.46)} & \underline{55.63 (5.93)} & 35.09 (8.53) \\
    \quad Bris-T1D & \underline{62.26 (22.20)} & \textbf{67.53 (14.33)} & 50.20 (19.97) \\
    \quad AZT1D & \textbf{79.32 (1.62)} & \underline{50.46 (5.24)} & 38.16 (7.28) \\
    \midrule
    \textbf{ControlledAccess} & & & \\
    \quad OhioT1DM & \textbf{80.02 (2.02)} & \underline{55.17 (4.51)} & 29.90 (3.49) \\
    \quad DiaTrend & \textbf{78.69 (1.35)} & \underline{54.13 (4.56)} & 33.77 (6.11) \\
    \quad T1DEXI & \textbf{78.45 (2.03)} & \underline{54.16 (6.70)} & 36.61 (7.54) \\
    \midrule
    \textbf{Overall Open}  & \textbf{71.32 (9.29)} & \underline{62.37 (14.96)} & 41.50 (15.99) \\
    \textbf{Overall}  & \textbf{76.98 (5.87)} & \underline{56.00 (10.90)} & 36.23 (11.41) \\
    \bottomrule
  \end{tabular}
  \begin{tablenotes}[flushleft]
    \footnotesize
    \item[1] Colas2019 lacks sufficient test sequence length for evaluation at a 12-h context length, so probabilistic metrics are not reported here.
    \item[*] Cells are Mean (STD) of PI80 coverage across participants (mean is patient-weighted; std is unweighted across participants).
  \end{tablenotes}
  \end{threeparttable}
\end{table}

\clearpage
\section{Impact of Additional T1D-Specific Training Data}
To examine whether additional T1D-specific training data improve forecasting performance, we conducted a supplementary experiment incorporating the first 80\% of each controlled-access dataset (OhioT1DM, DiaTrend, and T1DEXI) into the training data and continuing training for three representative models: LSTM, Chronos-2, and CALF. We evaluated their performance on the remaining 20\% and compared it with the original models under the same evaluation split. This experiment is separate from our primary benchmarking protocol, which uses only open-access datasets for training and reserves controlled-access datasets for external evaluation.

\begin{table}[htbp]
\centering
\caption{Effect of additional T1D-specific training data. Models are further trained on the first 80\% of each controlled dataset (Updated) and compared with the original models (Original) on the held-out 20\%. Values are RMSE (mg/dL); the better result in each pair is shown in \textbf{bold}.}
\label{tab:additional_t1d_data}
\small
\setlength{\tabcolsep}{5pt}
\begin{tabular}{l cc cc cc}
\toprule
& \multicolumn{2}{c}{LSTM} & \multicolumn{2}{c}{Chronos} & \multicolumn{2}{c}{CALF} \\
\cmidrule(lr){2-3} \cmidrule(lr){4-5} \cmidrule(lr){6-7}
Dataset & Original & Updated & Original & Updated & Original & Updated \\
\midrule
OpenAccess & \textbf{14.37} & 15.23 & 15.26 & \textbf{14.84} & \textbf{16.18} & 16.58 \\
OhioT1DM   & 18.78 & \textbf{18.66} & 19.58 & \textbf{18.96} & \textbf{21.65} & 21.90 \\
DiaTrend   & 24.20 & \textbf{23.64} & 25.00 & \textbf{24.58} & \textbf{27.47} & 27.66 \\
T1DEXI     & 19.70 & \textbf{19.37} & 20.44 & \textbf{19.83} & \textbf{22.25} & 22.57 \\
\bottomrule
\end{tabular}
\end{table}

As shown in Table~\ref{tab:additional_t1d_data}, incorporating additional T1D-specific data improved LSTM and Chronos-2 across all three controlled-access datasets, reducing RMSE by 0.12–0.62 mg/dL. In contrast, CALF showed a slight performance decrease. Performance on the open-access datasets also varied across models. These findings suggest that additional domain-specific training data can improve forecasting accuracy, but the benefits are modest and model-dependent rather than universal.

\clearpage
\section{Time Efficiency Analysis}
\label{sec:time-efficiency-analysis}
\begin{table}[ht]
\centering
\small
\setlength{\tabcolsep}{4pt}
\begin{threeparttable}
\caption{Fine-tuning time and inference latency of all models on the benchmark.}
\label{tab:efficiency}
\begin{tabular}{lcccccccc}
\toprule
 & \multicolumn{2}{c}{Task-specific}
 & \multicolumn{4}{c}{Time series foundation models}
 & \multicolumn{2}{c}{LLM-based} \\
\cmidrule(lr){2-3}\cmidrule(lr){4-7}\cmidrule(lr){8-9}
 & LSTM & GPFormer & Chronos\tnote{a} & Moirai & TimesFM & Timer & TimeLLM & CALF \\
\midrule
\multicolumn{9}{l}{\textit{Fine-tuning time (min)}} \\
Few-shot  & 1.5 & 2.4 & 2.0 & 4.4 & 22  & 2.7 & 8.3 & 8.5 \\
Full-shot & 15  & 45  & 41  & 100 & 498 & 63  & 42  & 71  \\
\addlinespace
\multicolumn{9}{l}{\textit{Inference latency (ms per forecast)}} \\
Zero-shot & --   & --   & 38 & 2.4 & 11 & 0.84 & --  & --  \\
Few-shot  & 0.06 & 1.0  & 65 & 2.4 & 11 & 0.83 & 1.7 & 1.2 \\
Full-shot & 0.06 & 0.99 & 66 & 2.4 & 11 & 0.81 & 1.7 & 1.1 \\
\bottomrule
\end{tabular}
\begin{tablenotes}\footnotesize
\item Times cover model training and inference only, excluding data loading
and preprocessing. All models were run on a single A100 GPU.
\item Inference latency is reported per forecast (one 12-h input window),
using each model's default batch size.
\item --: not applicable (no pretrained weights for zero-shot use).
\item[a] Fine-tuned with its internal LoRA framework.
\end{tablenotes}
\end{threeparttable}
\end{table}

\clearpage
\section{Stratified Result Analysis Details}
\vspace{-0.5em}
\subsection{Cohort stratified analysis}
\vspace{-0.5em}
\begin{figure*}[ht]
    \centering
    \begin{subfigure}[t]{0.7\textwidth}
        \centering
        \includegraphics[width=\linewidth]{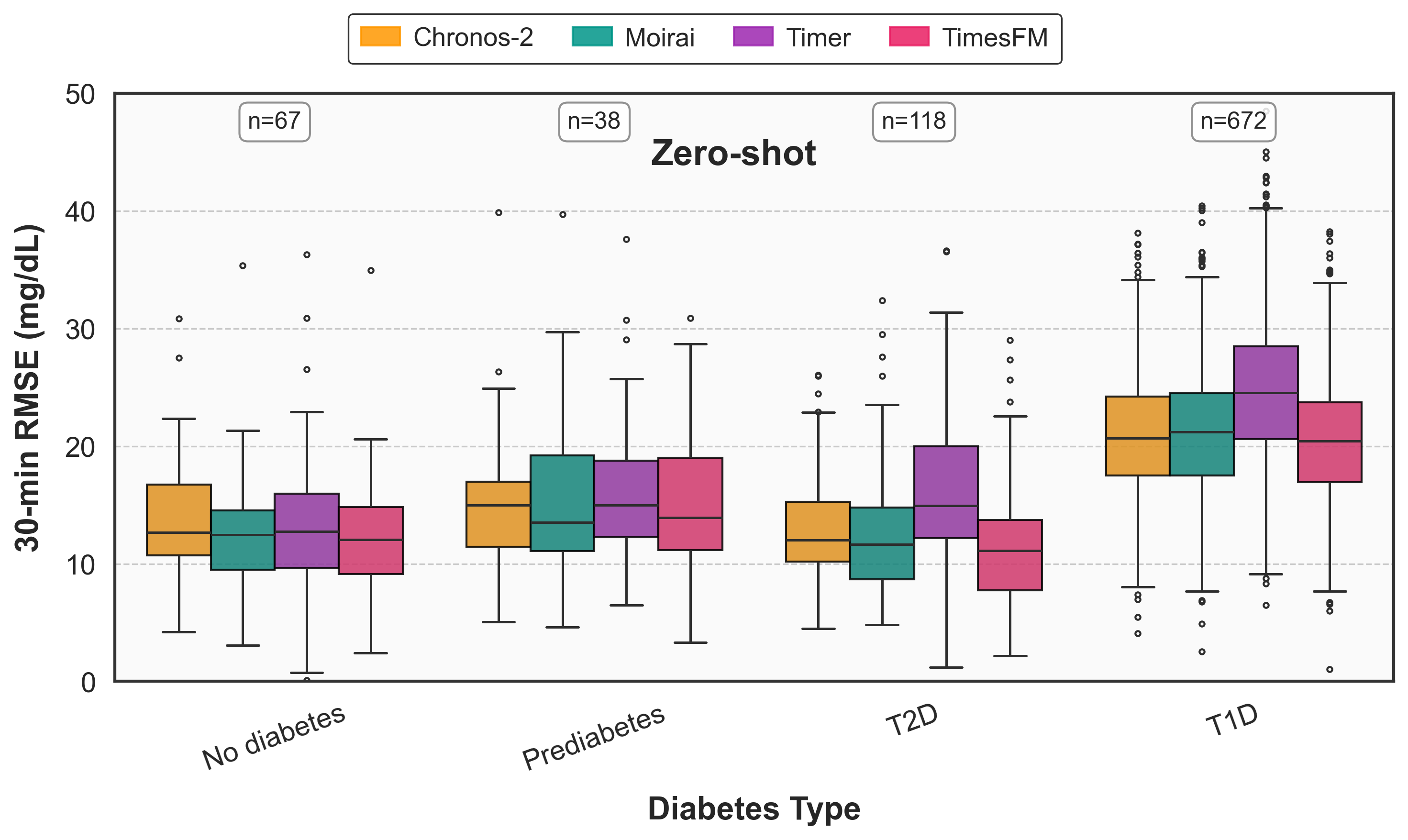}
        \caption{Zero-shot.}
        \label{fig:rmse-zero-shot}
    \end{subfigure}
    \hfill
    \begin{subfigure}[t]{\textwidth}
        \centering
        \includegraphics[width=\linewidth]{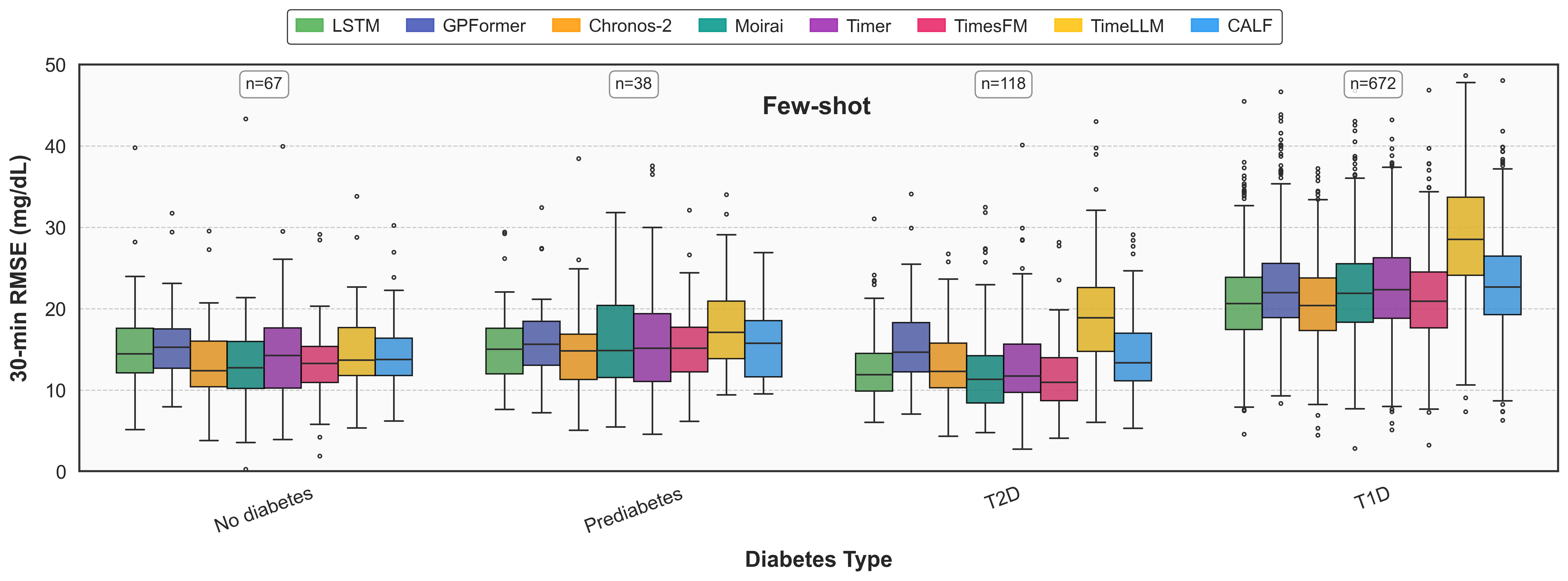}
        \caption{Few-shot.}
        \label{fig:rmse-few-shot}
    \end{subfigure}

    \begin{subfigure}[t]{\textwidth}
        \centering
        \includegraphics[width=\linewidth]{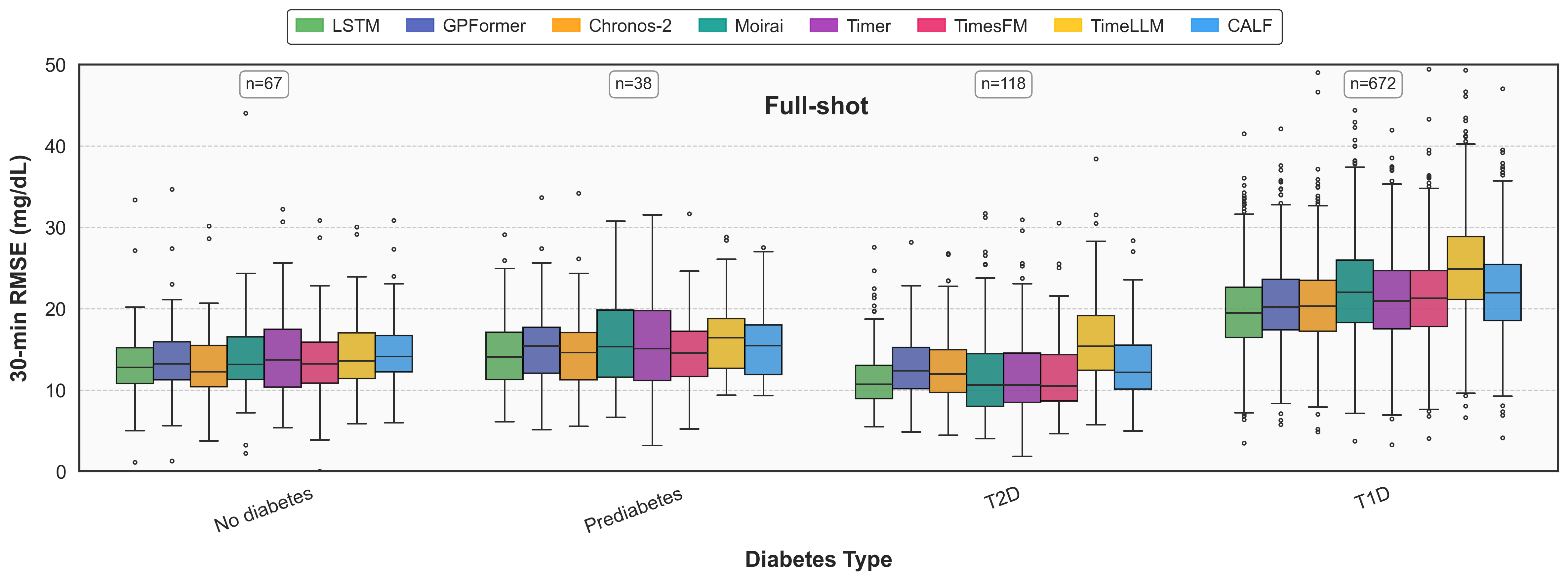}
        \caption{Full-shot.}
        \label{fig:rmse-full-shot}
    \end{subfigure}

    \caption{30-min RMSE (mg/dL) by diabetes phenotype across forecasting regimes.
    (a) Zero-shot evaluation of pre-trained TSFMs (Chronos-2, Moirai, Timer, TimesFM).
    (b) Few-shot and (c) full-shot results extend the comparison with task-specific
    baselines (LSTM, GPFormer) and TS LLM-based models (TimeLLM, CALF).
    Sample sizes per phenotype: No diabetes $n{=}67$, Prediabetes $n{=}38$,
    T2D $n{=}118$, T1D $n{=}672$.}
    \label{fig:rmse-by-phenotype}
\end{figure*}

\clearpage
\subsection{Glycemic condition stratified analysis}

\begin{table*}[ht]
\centering
\caption{Glycemic-stratified RMSE performance in mg/dL. Results are reported as mean (std), computed per participant and then averaged. Lower RMSE is better.}
\label{tab:glycemic_stratified_rmse}

\scriptsize

\begin{subtable}{\textwidth}
\centering
\caption{Zero-shot}
\begin{tabular}{lcccc}
\toprule
\textbf{Glycemic stratum} & \textbf{Chronos} & \textbf{Moirai} & \textbf{Timer} & \textbf{TimesFM} \\
\midrule
\multicolumn{5}{l}{\textbf{Horizon = 15 min}} \\
\quad Hyper ($>180$)       & 13.02 (4.98)  & 12.82 (5.87)  & 19.23 (7.21)  & \textbf{12.12 (5.66)} \\
\quad In-range ($[70,180]$) & 9.76 (4.10)   & 9.59 (4.20)   & 11.19 (4.35)  & \textbf{9.11 (4.08)} \\
\quad Hypo ($<70$)          & 9.95 (6.23)   & \textbf{8.79 (7.53)} & 14.60 (9.19) & 8.84 (7.57) \\
\addlinespace[0.2em]
\multicolumn{5}{l}{\textbf{Horizon = 30 min}} \\
\quad Hyper ($>180$)        & 26.35 (9.62)  & 26.15 (10.98) & 34.92 (11.53) & \textbf{25.56 (10.65)} \\
\quad In-range ($[70,180]$) & 17.79 (5.77)  & 17.75 (6.29)  & 19.46 (7.10)  & \textbf{17.00 (6.12)} \\
\quad Hypo ($<70$)          & 21.09 (12.31) & \textbf{19.90 (13.41)} & 33.19 (17.32) & 20.27 (13.71) \\
\addlinespace[0.2em]
\multicolumn{5}{l}{\textbf{Horizon = 60 min}} \\
\quad Hyper ($>180$)        & \textbf{48.83 (14.99)} & 49.92 (16.66) & 57.54 (15.80) & 49.06 (16.32) \\
\quad In-range ($[70,180]$) & 28.50 (10.14) & 28.74 (10.03) & 29.41 (11.68) & \textbf{27.29 (9.89)} \\
\quad Hypo ($<70$)          & 44.72 (23.37) & 43.44 (25.19) & 56.55 (25.52) & \textbf{43.15 (24.34)} \\
\addlinespace[0.2em]
\multicolumn{5}{l}{\textbf{Horizon = 90 min}} \\
\quad Hyper ($>180$)        & \textbf{63.82 (17.33)} & 65.81 (17.90) & 69.03 (17.31) & 64.04 (17.63) \\
\quad In-range ($[70,180]$) & 34.68 (13.83) & 35.62 (13.42) & 35.61 (15.26) & \textbf{33.66 (13.28)} \\
\quad Hypo ($<70$)          & 62.34 (29.21) & 61.80 (31.75) & 73.22 (32.28) & \textbf{60.54 (30.54)} \\
\bottomrule
\end{tabular}
\end{subtable}

\vspace{0.6em}

\begin{subtable}{\textwidth}
\centering
\caption{Few-shot}
\resizebox{\textwidth}{!}{
\begin{tabular}{lcccccccc}
\toprule
\textbf{Glycemic stratum} & \textbf{LSTM} & \textbf{GPFormer} & \textbf{Chronos} & \textbf{Moirai} & \textbf{Timer} & \textbf{TimesFM} & \textbf{Time-LLM} & \textbf{CALF} \\
\midrule
\multicolumn{9}{l}{\textbf{Horizon = 15 min}} \\
\quad Hyper ($>180$)        & 12.53 (4.74)  & 13.97 (5.08)  & 13.19 (4.84)  & 12.88 (5.84)  & 13.63 (5.50)  & \textbf{12.45 (5.72)} & 28.42 (9.03) & 16.80 (6.47) \\
\quad In-range ($[70,180]$) & \textbf{9.26 (3.76)}  & 11.11 (3.98)  & 9.56 (3.98)   & 9.78 (4.39)   & 10.35 (4.47)  & 9.39 (4.16)  & 17.33 (5.90) & 11.99 (4.49) \\
\quad Hypo ($<70$)          & 11.69 (6.22)  & \textbf{9.58 (6.25)}  & 11.13 (6.06)  & 9.92 (8.19)   & 11.73 (7.38)  & 9.90 (7.79)  & 30.40 (12.94) & 14.51 (9.23) \\
\addlinespace[0.2em]
\multicolumn{9}{l}{\textbf{Horizon = 30 min}} \\
\quad Hyper ($>180$)        & 26.03 (8.51)  & 30.38 (8.12)  & 26.75 (9.35)  & 26.57 (10.75) & 27.49 (9.73)  & \textbf{25.62 (10.45)} & 41.33 (12.07) & 29.40 (9.94) \\
\quad In-range ($[70,180]$) & \textbf{16.87 (4.86)} & 17.28 (4.42)  & 17.23 (5.56)  & 18.52 (6.66)  & 19.09 (6.53)  & 17.50 (6.19) & 22.84 (7.88) & 19.21 (6.05) \\
\quad Hypo ($<70$)          & 28.87 (10.30) & 32.07 (11.12) & 22.72 (11.68) & 22.84 (15.27) & 24.86 (13.82) & \textbf{21.48 (14.08)} & 41.02 (17.67) & 25.89 (13.82) \\
\addlinespace[0.2em]
\multicolumn{9}{l}{\textbf{Horizon = 60 min}} \\
\quad Hyper ($>180$)        & \textbf{47.04 (12.46)} & 52.00 (12.45) & 49.09 (14.86) & 50.59 (15.57) & 51.32 (14.76) & 48.54 (15.50) & 59.44 (14.91) & 51.38 (13.99) \\
\quad In-range ($[70,180]$) & \textbf{24.24 (6.70)}  & 30.63 (8.98)  & 27.38 (9.52)  & 29.50 (10.54) & 30.88 (10.64) & 28.32 (10.20) & 30.60 (12.14) & 28.02 (9.42)  \\
\quad Hypo ($<70$)          & 52.94 (16.79) & 61.74 (21.96) & 45.90 (22.65) & \textbf{44.92 (25.54)} & 49.38 (24.85) & 46.76 (25.45) & 61.40 (25.59) & 48.18 (22.87) \\
\addlinespace[0.2em]
\multicolumn{9}{l}{\textbf{Horizon = 90 min}} \\
\quad Hyper ($>180$)        & \textbf{60.03 (14.18)} & 70.18 (14.75) & 63.92 (17.05) & 66.48 (17.92) & 66.78 (16.74) & 63.43 (17.08) & 68.03 (15.76) & 65.60 (15.62) \\
\quad In-range ($[70,180]$) & \textbf{28.89 (8.56)}  & 32.75 (10.30) & 33.42 (13.34) & 36.40 (13.79) & 36.35 (12.81) & 34.43 (13.48) & 33.91 (13.66) & 32.64 (12.07) \\
\quad Hypo ($<70$)          & 68.39 (19.72) & 76.19 (23.80) & 63.16 (28.01) & \textbf{62.22 (32.46)} & 64.36 (30.02) & 66.20 (30.91) & 71.47 (29.61) & 63.40 (26.67) \\
\bottomrule
\end{tabular}
}
\end{subtable}

\vspace{0.6em}

\begin{subtable}{\textwidth}
\centering
\caption{Full-shot}
\resizebox{\textwidth}{!}{
\begin{tabular}{lcccccccc}
\toprule
\textbf{Glycemic stratum} & \textbf{LSTM} & \textbf{GPFormer} & \textbf{Chronos} & \textbf{Moirai} & \textbf{Timer} & \textbf{TimesFM} & \textbf{Time-LLM} & \textbf{CALF} \\
\midrule
\multicolumn{9}{l}{\textbf{Horizon = 15 min}} \\
\quad Hyper ($>180$)        & \textbf{12.21 (4.69)} & 12.81 (4.87)  & 12.84 (4.91)  & 12.59 (5.98)  & 12.86 (5.34)  & 12.33 (5.62) & 20.51 (6.74) & 15.51 (6.25) \\
\quad In-range ($[70,180]$) & \textbf{9.04 (3.79)}  & 9.38 (3.82)   & 9.48 (4.09)   & 9.60 (4.52)   & 9.72 (4.44)   & 9.49 (4.42)  & 13.88 (4.92) & 10.94 (4.44) \\
\quad Hypo ($<70$)          & 11.45 (6.09)  & 11.56 (6.15)  & 9.89 (5.84)   & 9.55 (8.06)   & 10.58 (6.67)  & \textbf{9.54 (7.49)} & 24.29 (10.77) & 12.50 (8.90) \\
\addlinespace[0.2em]
\multicolumn{9}{l}{\textbf{Horizon = 30 min}} \\
\quad Hyper ($>180$)        & \textbf{24.64 (8.44)} & 25.52 (8.73)  & 26.54 (8.78)  & 26.79 (10.95) & 25.44 (9.61)  & 25.58 (10.16) & 34.52 (10.37) & 27.94 (9.89) \\
\quad In-range ($[70,180]$) & \textbf{16.41 (4.57)} & 17.14 (4.94)  & 16.98 (5.47)  & 18.60 (6.84)  & 17.97 (6.14)  & 17.75 (6.27) & 20.14 (6.58) & 18.65 (5.81) \\
\quad Hypo ($<70$)          & 27.70 (10.51) & 27.81 (9.98)  & \textbf{21.46 (11.33)} & 22.89 (16.38) & 23.25 (13.35) & 21.83 (14.29) & 34.47 (15.09) & 24.06 (13.30) \\
\addlinespace[0.2em]
\multicolumn{9}{l}{\textbf{Horizon = 60 min}} \\
\quad Hyper ($>180$)        & \textbf{46.04 (12.33)} & 47.13 (12.60) & 50.42 (14.25) & 50.64 (16.09) & 47.90 (14.73) & 48.79 (15.23) & 54.29 (13.75) & 49.40 (14.16) \\
\quad In-range ($[70,180]$) & \textbf{24.64 (6.86)}  & 25.64 (7.94)  & 26.08 (8.95)  & 29.52 (10.54) & 29.23 (10.28) & 28.13 (9.80)  & 28.44 (10.41) & 28.02 (9.43) \\
\quad Hypo ($<70$)          & 53.35 (17.24) & 52.32 (19.61) & 45.46 (21.32) & \textbf{45.31 (27.70)} & 49.56 (24.66) & 47.42 (24.11) & 55.93 (23.46) & 47.25 (23.34) \\
\addlinespace[0.2em]
\multicolumn{9}{l}{\textbf{Horizon = 90 min}} \\
\quad Hyper ($>180$)        & \textbf{59.26 (13.90)} & 61.62 (13.96) & 65.30 (16.61) & 66.33 (17.38) & 63.53 (16.55) & 63.27 (17.27) & 65.90 (15.27) & 63.90 (16.12) \\
\quad In-range ($[70,180]$) & \textbf{29.29 (8.59)}  & 29.45 (8.91)  & 31.35 (12.28) & 36.14 (14.37) & 35.32 (12.58) & 34.49 (13.52) & 33.19 (13.06) & 32.94 (12.25) \\
\quad Hypo ($<70$)          & 69.11 (19.59) & 66.37 (22.45) & \textbf{61.84 (26.08)} & 64.45 (33.58) & 66.30 (29.75) & 65.72 (31.49) & 69.65 (28.53) & 62.51 (26.66) \\
\bottomrule
\end{tabular}
}
\end{subtable}

\end{table*}

\clearpage
\subsection{Time-of-day and Post-meal Stratified Analysis}
\label{sec:time-stratify}

\begin{table}[htbp]
\centering
\caption{30-minute forecasting error by time of day and post-meal state across 20 model--protocol combinations (12.6M forecasts). Relative change is reported as the mean with [min, max] across combinations.}
\label{tab:time_of_day}
\small
\begin{tabular}{lrrr}
\toprule
Window & Share of forecasts & RMSE (mg/dL) & $\Delta$ vs.\ daytime baseline \\
\midrule
Overnight (00:00--05:59)        & 24.8\% & 14.5 & $-30.2\%$ [$-39.0$, $-26.8$] \\
Daytime baseline$^{a}$          & 37.7\% & 20.6 & -- \\
Breakfast (07:00--09:59)        & 12.5\% & 19.4 & $-8.6\%$ [$-11.3$, $+7.0$] \\
Lunch (12:00--14:59)            & 12.5\% & 22.8 & $+6.8\%$ [$+4.1$, $+15.3$] \\
Dinner (18:00--20:59)           & 12.5\% & 23.2 & $+9.7\%$ [$+6.3$, $+10.8$] \\
\midrule
Post-meal, state-based$^{b}$    & 17.0\% & 26.6 & $+31.1\%$ [$+24.6$, $+54.7$] \\
\bottomrule
\end{tabular}
\par\smallskip
\raggedright\footnotesize
$^{a}$Hours between 06:00 and 23:59 outside the three meal windows.

$^{b}$Forecasts whose input context ends with a glucose rise of $\geq$15~mg/dL over the final 30 minutes; overlaps with the clock-based windows.
\end{table}
We analyzed 12.6 million 30-minute forecasts across 20 model–protocol combinations to examine temporal variations in forecasting performance. As shown in Table~\ref{tab:time_of_day}, overnight forecasting consistently achieved lower errors, with RMSE approximately 30\% below the daytime baseline. In contrast, periods characterized by a glucose increase of at least 15 mg/dL during the final 30 minutes of the input context exhibited substantially higher errors (+31.1\%). Clock-based meal windows showed smaller and less consistent differences, suggesting that glucose dynamics provide a more informative indicator of challenging forecasting periods than time of day alone.

\subsection{Glucose Variability Stratified Analysis}
\begin{table}[htbp]
\centering
\caption{30-minute forecasting error stratified by subject-level glycemic variability, measured by the coefficient of variation (CV). RMSE is reported as mean $\pm$ STD across subjects.}
\label{tab:cv_strata}
\small
\begin{tabular}{llrrc}
\toprule
Stratum & CV range\cite{Danne2017} & Subjects & Mean CV & RMSE (mg/dL) \\
\midrule
Low       & $<27\%$      & 317 & 20.7\% & $15.9 \pm 4.9$ \\
Moderate  & 27--36\%     & 376 & 31.5\% & $21.9 \pm 5.6$ \\
High      & 36--45\%     & 162 & 39.4\% & $25.1 \pm 6.3$ \\
Very high & $\geq 45\%$  & 29  & 50.3\% & $25.8 \pm 7.2$ \\
\bottomrule
\end{tabular}
\end{table}

We further investigated the relationship between participant-level glycemic variability, measured by the coefficient of variation (CV), and forecasting performance. As shown in Table~\ref{tab:cv_strata}, RMSE increased from 15.9 mg/dL in the low-variability group to 25.8 mg/dL in the very-high-variability group, indicating a positive association between glycemic variability and forecasting error. The increase was most pronounced across the low-, moderate-, and high-variability strata, with a smaller difference observed above 45\% CV. However, results for the very-high-variability group should be interpreted cautiously due to its limited sample size ($n=29$).

\section{Statistical Test between Model Performance}
\label{sec:statistical_test_model}

We conduct pairwise paired Wilcoxon tests on per-participant 30-minute RMSE with Benjamini--Hochberg false discovery rate correction. Adjusted q-values are reported in Tables~\ref{tab:rmse30-sig-zero}--\ref{tab:rmse30-sig-full}.

\begin{table}[t]
\centering
\small
\caption{Zero-shot pairwise paired Wilcoxon tests on per-patient RMSE at the 30-min horizon. Cells show BH-FDR adjusted $q$-values; significance: $^{\ast\ast\ast}q\!<\!10^{-3}$, $^{\ast\ast}q\!<\!10^{-2}$, $^{\ast}q\!<\!0.05$, \textsc{ns}\,$=$\,not significant.}
\label{tab:rmse30-sig-zero}
\begin{tabular}{lccc}
\toprule
 & Chronos2 & Timer & TimesFM \\
\midrule
Timer   & $1.5{\times}10^{-117}{}^{\ast\ast\ast}$ &                                          &                                         \\
TimesFM & $6.0{\times}10^{-12}{}^{\ast\ast\ast}$  & $1.1{\times}10^{-135}{}^{\ast\ast\ast}$  &                                         \\
Moirai  & $5.0{\times}10^{-4}{}^{\ast\ast\ast}$   & $8.0{\times}10^{-125}{}^{\ast\ast\ast}$  & $1.3{\times}10^{-78}{}^{\ast\ast\ast}$  \\
\bottomrule
\end{tabular}
\end{table}

\begin{table}[t]
\centering
\scriptsize
\setlength{\tabcolsep}{3pt}
\caption{Few-shot pairwise paired Wilcoxon tests on per-patient RMSE at the 30-min horizon. Cells show BH-FDR adjusted $q$-values; significance: $^{\ast\ast\ast}q\!<\!10^{-3}$, $^{\ast\ast}q\!<\!10^{-2}$, $^{\ast}q\!<\!0.05$, \textsc{ns}\,$=$\,not significant.}
\label{tab:rmse30-sig-few}
\begin{tabular}{lccccccc}
\toprule
 & Chronos2 & CALF & LSTM & GPFormer & TimeLLM & Timer & TimesFM \\
\midrule
CALF     & $3.4{\times}10^{-134}{}^{\ast\ast\ast}$ &                                          &                                          &                                          &                                          &                                          &                                         \\
LSTM     & $4.4{\times}10^{-16}{}^{\ast\ast\ast}$  & $2.4{\times}10^{-100}{}^{\ast\ast\ast}$  &                                          &                                          &                                          &                                          &                                         \\
GPFormer & $7.8{\times}10^{-116}{}^{\ast\ast\ast}$ & $0.044{}^{\ast}$                         & $6.0{\times}10^{-95}{}^{\ast\ast\ast}$   &                                          &                                          &                                          &                                         \\
TimeLLM  & $9.2{\times}10^{-143}{}^{\ast\ast\ast}$ & $9.2{\times}10^{-141}{}^{\ast\ast\ast}$  & $1.1{\times}10^{-138}{}^{\ast\ast\ast}$  & $1.9{\times}10^{-125}{}^{\ast\ast\ast}$  &                                          &                                          &                                         \\
Timer    & $1.0{\times}10^{-89}{}^{\ast\ast\ast}$  & $0.001{}^{\ast\ast}$                     & $1.9{\times}10^{-74}{}^{\ast\ast\ast}$   & $0.663{}^{\textsc{ns}}$                  & $3.9{\times}10^{-129}{}^{\ast\ast\ast}$  &                                          &                                         \\
TimesFM  & $5.2{\times}10^{-19}{}^{\ast\ast\ast}$  & $9.3{\times}10^{-111}{}^{\ast\ast\ast}$  & $0.235{}^{\textsc{ns}}$                  & $8.8{\times}10^{-71}{}^{\ast\ast\ast}$   & $3.8{\times}10^{-144}{}^{\ast\ast\ast}$  & $2.7{\times}10^{-96}{}^{\ast\ast\ast}$   &                                         \\
Moirai   & $4.2{\times}10^{-35}{}^{\ast\ast\ast}$  & $3.8{\times}10^{-21}{}^{\ast\ast\ast}$   & $1.1{\times}10^{-19}{}^{\ast\ast\ast}$   & $2.0{\times}10^{-11}{}^{\ast\ast\ast}$   & $2.5{\times}10^{-133}{}^{\ast\ast\ast}$  & $5.2{\times}10^{-19}{}^{\ast\ast\ast}$   & $3.6{\times}10^{-29}{}^{\ast\ast\ast}$ \\
\bottomrule
\end{tabular}
\end{table}

\begin{table}[t]
\centering
\scriptsize
\setlength{\tabcolsep}{3pt}
\caption{Full-shot pairwise paired Wilcoxon tests on per-patient RMSE at the 30-min horizon. Cells show BH-FDR adjusted $q$-values; significance: $^{\ast\ast\ast}q\!<\!10^{-3}$, $^{\ast\ast}q\!<\!10^{-2}$, $^{\ast}q\!<\!0.05$, \textsc{ns}\,$=$\,not significant.}
\label{tab:rmse30-sig-full}
\begin{tabular}{lccccccc}
\toprule
 & Chronos2 & CALF & LSTM & GPFormer & TimeLLM & Timer & TimesFM \\
\midrule
CALF     & $3.6{\times}10^{-121}{}^{\ast\ast\ast}$ &                                          &                                          &                                          &                                          &                                          &                                         \\
LSTM     & $1.4{\times}10^{-90}{}^{\ast\ast\ast}$  & $8.7{\times}10^{-136}{}^{\ast\ast\ast}$  &                                          &                                          &                                          &                                          &                                         \\
GPFormer & $9.8{\times}10^{-6}{}^{\ast\ast\ast}$   & $6.4{\times}10^{-90}{}^{\ast\ast\ast}$   & $1.5{\times}10^{-111}{}^{\ast\ast\ast}$  &                                          &                                          &                                          &                                         \\
TimeLLM  & $6.2{\times}10^{-143}{}^{\ast\ast\ast}$ & $2.2{\times}10^{-130}{}^{\ast\ast\ast}$  & $4.9{\times}10^{-143}{}^{\ast\ast\ast}$  & $2.3{\times}10^{-136}{}^{\ast\ast\ast}$  &                                          &                                          &                                         \\
Timer    & $2.8{\times}10^{-32}{}^{\ast\ast\ast}$  & $6.3{\times}10^{-36}{}^{\ast\ast\ast}$   & $6.5{\times}10^{-87}{}^{\ast\ast\ast}$   & $2.1{\times}10^{-15}{}^{\ast\ast\ast}$   & $1.9{\times}10^{-118}{}^{\ast\ast\ast}$  &                                          &                                         \\
TimesFM  & $8.3{\times}10^{-46}{}^{\ast\ast\ast}$  & $1.1{\times}10^{-42}{}^{\ast\ast\ast}$   & $2.0{\times}10^{-98}{}^{\ast\ast\ast}$   & $4.2{\times}10^{-20}{}^{\ast\ast\ast}$   & $5.9{\times}10^{-134}{}^{\ast\ast\ast}$  & $0.069{}^{\textsc{ns}}$                  &                                         \\
Moirai   & $3.2{\times}10^{-46}{}^{\ast\ast\ast}$  & $0.435{}^{\textsc{ns}}$                  & $2.8{\times}10^{-87}{}^{\ast\ast\ast}$   & $1.5{\times}10^{-32}{}^{\ast\ast\ast}$   & $6.0{\times}10^{-78}{}^{\ast\ast\ast}$   & $1.4{\times}10^{-25}{}^{\ast\ast\ast}$   & $3.5{\times}10^{-23}{}^{\ast\ast\ast}$ \\
\bottomrule
\end{tabular}
\end{table}

\clearpage
\section{Dataset License}

\begin{table}[h]
\centering
\caption{License information and access details for the open-access CGM datasets used in GlucoFM-Bench.}
\label{tab:dataset_licenses}
\small
\begin{tabularx}{\textwidth}{l l l X}
\toprule
\textbf{Sub-dataset} & \textbf{Access Type} & \textbf{License} & \textbf{Source / Notes} \\
\midrule
Hall\_2018 & Open Access & CC BY 4.0 & Journal open data \\
D1NAMO & Open Access & CC BY-SA 4.0 & Zenodo record \\
Colas\_2019 & Open Access & CC BY 4.0 & Published open access \\
BIG IDEAs & Open Access & ODC-By v1.0 & Publication linked \\
ShanghaiT1DM & Open Access & CC BY 4.0 & Open \\
ShanghaiT2DM & Open Access & CC BY 4.0 & Open \\
UCHTT1DM$^\dagger$ & Open Access  & CC BY-NC-ND 4.0 & Hosted on GitHub \\
HUPA-UCM & Open Access & CC BY 4.0 & Mendeley Data \\
CGMacros & Open Access & CC BY-NC-SA 4.0 & Newly released open \\
T1D-UOM & Open Access & CC BY 4.0 & Zenodo \\
BrisT1D-Open & Open Access & CC BY 4.0 & Open \\
AZT1D & Open Access & CC BY 4.0 & Open \\
\bottomrule
\end{tabularx}
\vspace{2pt}
\footnotesize{$^\dagger$ Excluded from redistribution due to no-derivatives restriction; readers can access the original source.}
\end{table}


\clearpage 
\newpage

\end{document}